\documentclass[twoside,11pt]{article}

\usepackage{blindtext}

\usepackage[preprint]{jmlr2e}
\usepackage{amsmath}
\usepackage{algorithmic}
\usepackage{algorithm}
\usepackage{multirow}
\usepackage{subfigure}

\newtheorem{assumption}{Assumption}

\usepackage{lastpage}
\jmlrheading{23}{2022}{1-\pageref{LastPage}}{1/21; Revised 5/22}{9/22}{21-0000}{Chang-Wei Shi, Shen-Yi Zhao, Yin-Peng Xie, Hao Gao and Wu-Jun Li}

\ShortHeadings{Global Momentum Compression for Sparse Communication in Distributed Learning}{Shi, Zhao, Xie, Gao and Li}
\firstpageno{1}

\begin{document}
\def\a{{\bf a}}
\def\b{{\bf b}}
\def\c{{\bf c}}
\def\d{{\bf d}}
\def\e{{\bf e}}
\def\f{{\bf f}}
\def\g{{\bf g}}
\def\h{{\bf h}}
\def\i{{\bf i}}
\def\j{{\bf j}}
\def\k{{\bf k}}
\def\l{{\bf l}}
\def\m{{\bf m}}
\def\n{{\bf n}}
\def\o{{\bf o}}
\def\p{{\bf p}}
\def\q{{\bf q}}
\def\r{{\bf r}}
\def\s{{\bf s}}
\def\t{{\bf t}}
\def\u{{\bf u}}
\def\v{{\bf v}}
\def\w{{\bf w}}
\def\x{{\bf x}}
\def\y{{\bf y}}
\def\z{{\bf z}}

\def\A{{\bf A}}
\def\B{{\bf B}}
\def\C{{\bf C}}
\def\D{{\bf D}}
\def\E{{\bf E}}
\def\F{{\bf F}}
\def\G{{\bf G}}
\def\H{{\bf H}}
\def\I{{\bf I}}
\def\J{{\bf J}}
\def\K{{\bf K}}
\def\L{{\bf L}}
\def\M{{\bf M}}
\def\N{{\bf N}}
\def\O{{\bf O}}
\def\P{{\bf P}}
\def\Q{{\bf Q}}
\def\R{{\bf R}}
\def\S{{\bf S}}
\def\T{{\bf T}}
\def\U{{\bf U}}
\def\V{{\bf V}}
\def\W{{\bf W}}
\def\X{{\bf X}}
\def\Y{{\bf Y}}
\def\Z{{\bf Z}}

\def\0{{\bf 0}}
\def\1{{\bf 1}}
\def\2{{\bf 2}}
\def\3{{\bf 3}}
\def\4{{\bf 4}}
\def\5{{\bf 5}}
\def\6{{\bf 6}}
\def\7{{\bf 7}}
\def\8{{\bf 8}}
\def\9{{\bf 9}}

\def\AM{{\mathcal A}}
\def\BM{{\mathcal B}}
\def\CM{{\mathcal C}}
\def\DM{{\mathcal D}}
\def\EM{{\mathcal E}}
\def\FM{{\mathcal F}}
\def\GM{{\mathcal G}}
\def\HM{{\mathcal H}}
\def\IM{{\mathcal I}}
\def\JM{{\mathcal J}}
\def\KM{{\mathcal K}}
\def\LM{{\mathcal L}}
\def\MM{{\mathcal M}}
\def\NM{{\mathcal N}}
\def\OM{{\mathcal O}}
\def\PM{{\mathcal P}}
\def\QM{{\mathcal Q}}
\def\RM{{\mathcal R}}
\def\SM{{\mathcal S}}
\def\TM{{\mathcal T}}
\def\UM{{\mathcal U}}
\def\VM{{\mathcal V}}
\def\WM{{\mathcal W}}
\def\XM{{\mathcal X}}
\def\YM{{\mathcal Y}}
\def\ZM{{\mathcal Z}}

\def\AB{{\mathbb A}}
\def\BB{{\mathbb B}}
\def\CB{{\mathbb C}}
\def\DB{{\mathbb D}}
\def\EB{{\mathbb E}}
\def\FB{{\mathbb F}}
\def\GB{{\mathbb G}}
\def\HB{{\mathbb H}}
\def\IB{{\mathbb I}}
\def\JB{{\mathbb J}}
\def\KB{{\mathbb K}}
\def\LB{{\mathbb L}}
\def\MB{{\mathbb M}}
\def\NB{{\mathbb N}}
\def\OB{{\mathbb O}}
\def\PB{{\mathbb P}}
\def\QB{{\mathbb Q}}
\def\RB{{\mathbb R}}
\def\SB{{\mathbb S}}
\def\TB{{\mathbb T}}
\def\UB{{\mathbb U}}
\def\VB{{\mathbb V}}
\def\WB{{\mathbb W}}
\def\XB{{\mathbb X}}
\def\YB{{\mathbb Y}}
\def\ZB{{\mathbb Z}}
\def\wS{{\widetilde{\S}}}
\def\wSS{{\widetilde{S}}}
\def\wb{{\widetilde{\b}}}
\def\wbb{{\widetilde{b}}}
\def\wQ{{\widetilde{\Q}}}
\def\ob{{\overline{\b}}}
\def\obb{{\overline{b}}}
\def\oQ{{\overline{\Q}}}
\def\op{{\overline{\p}}}
\def\opp{{\overline{p}}}
\def\cW{{\mathcal{W}}}
\def\cP{{\mathcal{P}}}
\def\balpha { {\bm \alpha}}
\def\bbeta { {\bm \beta}}
\def\bvartheta { {\bm \vartheta}}

\newcommand{\bigO}[1]{\ensuremath{\operatorname{\OM}(#1)}}

\title{Global Momentum Compression \\for Sparse Communication in Distributed Learning}
\author{\name Chang-Wei Shi\footnotemark[2] \email shicw@smail.nju.edu.cn \\
       \name Shen-Yi Zhao\footnotemark[2] \email zhaosy1009@gmail.com \\
       \name Yin-Peng Xie \email 2418111594@qq.com \\
       \name Hao Gao \email gh131220003@gmail.com \\
       \name Wu-Jun Li\footnotemark[1]\email liwujun@nju.edu.cn \\
       \addr National Key Laboratory for Novel Software Technology\\
       Department of Computer Science and Technology\\
       Nanjing University, Nanjing 210023, China}

% \contributions{Equal contribution.}
\renewcommand{\thefootnote}{\fnsymbol{footnote}}
\footnotetext[2]{Equal contribution.}
\footnotetext[1]{Corresponding author.}
\renewcommand{\thefootnote}{\arabic{footnote}}

\editor{My editor}

\maketitle

\begin{abstract}%   <- trailing '%' for backward compatibility of .sty file
  With the rapid growth of data, distributed momentum stochastic gradient descent~(DMSGD) has been widely used in distributed learning, especially for training large-scale deep models. 
  Due to the latency and limited bandwidth of the network, communication has become the bottleneck of distributed learning. 
  Communication compression with sparsified gradient, abbreviated as \emph{sparse communication}, has been widely employed to reduce communication cost. 
  All existing works about sparse communication in DMSGD employ local momentum, in which the momentum only accumulates stochastic gradients computed by each worker locally. 
  In this paper, we propose a novel method, called \emph{\underline{g}}lobal \emph{\underline{m}}omentum \emph{\underline{c}}ompression~(GMC), for sparse communication.
  Different from existing works that utilize local momentum, GMC utilizes global momentum. 
  Furthermore, to enhance the convergence performance when using more aggressive sparsification compressors (e.g., RBGS), we extend GMC to GMC+.
  We theoretically prove the convergence of GMC and GMC$+$. 
  To the best of our knowledge, this is the first work that introduces global momentum for sparse communication in distributed learning. 
  Empirical results demonstrate that, compared with the local momentum counterparts, our GMC and GMC+ can achieve higher test accuracy and exhibit faster convergence, especially under non-IID data distribution.
\end{abstract}


\section{Introduction}
Many machine learning models can be formulated as the following empirical risk minimization problem:
\begin{align}\label{equation: object}
	\min_{\w\in \RB^d} F(\w) = \frac{1}{|\DM|}\sum_{\xi \in \DM} f(\w;\xi),
\end{align}
where $\w$ denotes the model parameter of dimension $d$, $\DM$ denotes the training dataset, $f(\w;\xi)$ is the empirical loss on the instance $\xi$, $|\DM|$ is the number of training instances. 
For example, let $\xi = (\x,y)$, where $\x$ denotes the feature of the training data and $y$ denotes the label. Then in logistic regression $f(\w;\xi) = \log(1 + e^{-y\x^T\w}) + \frac{\lambda}{2}\|\w\|^2$, and in support vector machine~(SVM) $f(\w;\xi) = \max(0, 1-y\x^T\w) + \frac{\lambda}{2}\|\w\|^2$. Many deep learning models can also be formulated as~(\ref{equation: object}).

% One of the efficient ways to solve (\ref{equation: object}) is stochastic gradient descent~(SGD)~\citep{Robbins&Monro:1951, DBLP:journals/chinaf/ZhaoXL21, DBLP:journals/corr/abs-2007-13985}. 
Stochastic gradient descent~(SGD) and its variants~\citep{Robbins&Monro:1951, DBLP:conf/compstat/Bottou10, DBLP:conf/nips/Johnson013, NEURIPS2018_f4334c13, DBLP:journals/corr/abs-2007-13985, DBLP:journals/chinaf/ZhaoXL21} have been the dominating optimization methods for solving~(\ref{equation: object}).
In each iteration, SGD calculates a (mini-batch) stochastic gradient and uses it to update the model parameters. Inspired by momentum and Nesterov's accelerated gradient descent, momentum SGD~(MSGD)~\citep{article,DBLP:journals/siamjo/Tseng98,DBLP:journals/mp/Lan12,DBLP:journals/corr/KingmaB14} has been proposed and widely used in machine learning. In practice, MSGD often outperforms SGD~\citep{DBLP:conf/nips/KrizhevskySH12,DBLP:conf/icml/SutskeverMDH13}. Many machine learning platforms, such as TensorFlow, PyTorch and MXNet, include MSGD as one of their optimization methods.

With the rapid growth of data, distributed SGD~(DSGD) and its variant distributed MSGD~(DMSGD) have garnered much attention. They distribute the stochastic gradient computation across multiple workers to expedite the model training. 
These methods can be implemented on distributed frameworks like parameter server and all-reduce frameworks. 
Each worker computes stochastic gradients locally and communicates with the server or other workers to obtain the aggregated stochastic gradients for updating the model parameter. Recently, more and more large-scale deep learning models, such as large language models~\citep{Devlin2019BERTPO, brown2020language, DBLP:journals/corr/abs-2302-13971}, have been proposed in machine learning. In large-scale model training tasks, the communicated messages become high-dimensional vectors. Due to the latency and limited bandwidth of the network, communication cost has become the bottleneck of distributed learning, especially for cases with large-scale models. 
To address the communication bottleneck issue, there have been many works dedicated to compressing the communicated messages~\citep{DBLP:conf/nips/WenXYWWCL17,DBLP:conf/nips/AlistarhG0TV17,DBLP:conf/emnlp/AjiH17, DBLP:conf/nips/JiangA18,DBLP:conf/nips/AlistarhH0KKR18, DBLP:conf/nips/StichCJ18, DBLP:conf/iclr/LinHM0D18,DBLP:conf/nips/VogelsKJ19, DBLP:conf/icml/KarimireddyRSJ19,DBLP:conf/icml/TangYLZL19, DBLP:conf/nips/XieZKGLL20, DBLP:conf/icml/XuH22}. %and infrequent communication~\citep{DBLP:conf/iclr/LinSPJ20, DBLP:conf/iclr/Stich19, DBLP:conf/aaai/YuYZ19, DBLP:conf/icml/YuJY19, DBLP:conf/nips/0001DKD19, DBLP:conf/iclr/YouLRHKBSDKH20, DBLP:conf/iclr/LiuCCHY22, DBLP:journals/corr/abs-2007-13985}.

% In this work, we focus on the communication compression methods. 
Researchers have proposed two main categories of communication compression methods for reducing communication cost: quantization~\citep{DBLP:conf/nips/WenXYWWCL17,DBLP:conf/nips/AlistarhG0TV17,DBLP:conf/nips/JiangA18} and sparsification~\citep{DBLP:conf/emnlp/AjiH17,DBLP:conf/nips/AlistarhH0KKR18,DBLP:conf/nips/StichCJ18,DBLP:conf/icml/KarimireddyRSJ19,DBLP:conf/icml/TangYLZL19}. Quantization methods quantize the value~(gradient or parameter) representation from float~(typically 32 bits) to lower bit-width, such as eight bits or four bits. 
It's easy to find that the communication cost can be reduced at most by 31-fold in the ideal case. 
Sparsification methods, which are also called \emph{sparse communication} methods, select only a few components of the vector for communicating with the server or the other workers. The most widely used sparsification compressor adopted in sparse communication methods is top-$s$, where each worker selects $s$ components of the vector with the largest absolute values. 
Considering the high-dimensional nature of communicated vectors in large-scale model training, sparsification methods have the potential to achieve a much higher level of communication compression compared to the maximum 31-fold compression achieved by quantization methods.
% Due to the high-dimensional nature of gradients in the training process of large-scale deep learning models, sparsification methods can compress the gradients by a higher factor compared to quantization methods.

Due to the presence of compressed error, naively compressing the communicated vectors in DSGD or DMSGD will damage the convergence, especially when the compression ratio is high.
The most representative technique designed to tackle this issue is error feedback~\citep{DBLP:conf/nips/StichCJ18, DBLP:conf/icml/KarimireddyRSJ19}, %which is called error compensation~\citep{DBLP:conf/nips/StichCJ18, DBLP:conf/nips/0001DKD19} or local gradient accumulation~\citep{DBLP:conf/iclr/LinHM0D18}.
also called error compensation or memory vector.
The error feedback technique keeps the compressed error into the error residual on each worker and incorporates the error residual into the next update. 
Error feedback based sparse communication methods have been widely adopted by recent communication compression methods and achieved better performance than quantization methods and other sparse communication methods without error feedback.  
In existing error feedback based sparse communication methods, most are for vanilla DSGD~\citep{DBLP:conf/emnlp/AjiH17,DBLP:conf/nips/AlistarhH0KKR18,DBLP:conf/nips/StichCJ18,DBLP:conf/icml/KarimireddyRSJ19,DBLP:conf/icml/TangYLZL19}. 
There has appeared one error feedback based sparse communication method for DMSGD, called Deep Gradient Compression~(DGC)~\citep{DBLP:conf/iclr/LinHM0D18}, which has achieved better performance than vanilla DSGD with sparse communication in practice. 
However, the theory about the convergence of DGC is still lacking. Furthermore, although DGC combines momentum and error feedback, the momentum in DGC only accumulates stochastic gradients computed by each worker locally. Therefore, the momentum in DGC is a local momentum without global information.

Recently, \cite{DBLP:conf/nips/XieZKGLL20} proposes Communication-efficient SGD
with Error Reset~(CSER) that combines partial synchronization and error reset techniques. %Different from error feedback, error reset immediately applies the error residual into the parameter update.
Due to the extra communication and computation overhead of the top-$s$ compressor, 
some works~\citep{DBLP:conf/nips/VogelsKJ19, DBLP:conf/nips/XieZKGLL20, DBLP:conf/icml/XuH22} also consider a more aggressive sparsification compressor, called Random Blockwise Gradient Sparsification~(RBGS).
Since RBGS introduces a larger compressed error  compared with top-$s$ when selecting the same number of components of the original vector to communicate, vanilla error feedback methods usually fail to converge when using RBGS as the sparsification compressor. 
To address this convergence issue,
\cite{DBLP:conf/icml/XuH22} proposes DEF-A, which utilizes the detached error feedback technique.
Both CSER and DEF-A have their momentum variants~\citep{DBLP:conf/nips/XieZKGLL20, DBLP:conf/icml/XuH22}, but they use local momentum similar to DGC.

In this paper, we introduce global momentum and propose a novel method, called \emph{\underline{g}}lobal \emph{\underline{m}}omentum \emph{\underline{c}}ompression~(GMC), for sparse communication in distributed learning based on error feedback. 
The main contributions of this paper are outlined as follows:
\begin{itemize}
  \item GMC combines error feedback and momentum to achieve sparse communication in distributed learning. But different from existing sparse communication methods like DGC which adopt local momentum, GMC adopts global momentum.
  To the best of our knowledge, this is the first work to introduce global momentum into sparse communication methods.
  \item Furthermore, to enhance the convergence performance when using more aggressive sparsification compressors (e.g., RBGS), we extend GMC to GMC+ by introducing global momentum to the detached error feedback technique.
  \item We theoretically prove the convergence of both GMC and GMC$+$. 
  \item Empirical results demonstrate that, compared with the local momentum counterparts, GMC and GMC$+$ can achieve higher test accuracy and exhibit faster convergence, especially under non-IID data distribution.
\end{itemize}

% Manual newpage inserted to improve layout of sample file - not
% needed in general before appendices/bibliography.
\section{Preliminary}
\subsection{Notations}
In this paper, we use $\|\cdot\|$ to denote $L_2$ norm and use $\mathcal{C}(\cdot)$ to denote the sparsification compressor. For $n \in \mathbb{N}_{+}$, we use $[n]$ to denote $\{0,1,2,\ldots,n-1\}$.
$\nabla f(\w;\mathcal{I}) \triangleq \frac{1}{|\IM|}\sum_{\xi \in \IM} \nabla f(\w;\xi)$ denotes one stochastic gradient with respect to a mini-batch of training instances $\IM$.
For a vector $\a$, we use $a^{(j)}$ to denote its $j$-th component. $\|\a\|_0$ denotes the number of non-zero components in $\a$.

\subsection{Problem Formulation}

Assume we have $K$ workers. The training data are distributed or partitioned  across $K$ workers. Let $\mathcal{D}_k$ denote the training data stored on worker $k$, and $F_k(\w) = \frac{1}{|\mathcal{D}_k|}\sum_{\xi \in \mathcal{D}_k}f(\w;\xi)$ denote the local objective function on worker $k$. Assume $\mathcal{D}_k\cap \mathcal{D}_{k'}=\emptyset$ if $k\neq k'$, it's easy to verify that $F(\w) = \frac{1}{|\mathcal{D}|}\sum_{\xi \in \mathcal{D}}f(\w;\xi)=\sum_{k \in [K]} \frac{|\mathcal{D}_k|}{|\mathcal{D}|}F_k(\w)$, where $\DM = \cup_{k=1}^{K}\mathcal{D}_k$.
Without loss of generality, we assume that each worker has the same number of training data, that is $|\mathcal{D}_k|=|\mathcal{D}_{k'}|, \forall k,k' \in [K]$. 
Then the problem in~(\ref{equation: object}) can be rewritten as 
\begin{align}\label{equation:dist object}
  \min_{\w\in \RB^d}F(\w) = \frac{1}{K}\sum_{k \in [K]} F_k(\w). 
\end{align}

\subsection{Distributed Momentum SGD}
\label{sec:DMSGD}
The widely used momentum SGD~(MSGD)~\citep{article} can be written as
\begin{align*}
  &\m_{t+1} = \beta\m_{t} + \eta\nabla f(\w_t;\IM_t), \\
  &\w_{t+1} = \w_t - \m_{t+1},
\end{align*}
where $\beta\in [0,1)$. $\m_t$ is the Polyak's momentum. $\nabla f(\w_t;\IM_t)$ is one stochastic gradient with respect to a mini-batch of training instances $\IM_t$ sampled from $\DM$. Since $\m_t=-(\w_{t} - \w_{t-1})$, MSGD can also be written as
\begin{align}\label{eq:momentum SGD}
  \w_{t+1} = \w_t - \eta(\nabla f(\w_t;\IM_t)-\frac{\beta}{\eta}(\w_{t} - \w_{t-1})).
\end{align}
Please note that if $\beta = 0$, MSGD degenerates to SGD.

One way to implement distributed MSGD~(DMSGD) is using \emph{local momentum}:
\begin{align}
  & \m_{t+1,k} = \beta\m_{t,k} + \eta\nabla f(\w_t;\IM_{t,k}), |\IM_{t,k}|=b, k \in [K], \label{eq:localmo}\\
  & \w_{t+1} = \w_t - \frac{1}{K}\sum_{k \in [K]}\m_{t+1,k},
\end{align}
where $\m_{0,k}=\0, k \in [K]$ and $\nabla f(\w_t;\IM_{t,k})$ is the stochastic gradient with respect to a mini-batch $\IM_{t,k}$ of size $b$ sampled from $\DM_k$ on worker $k$. 
This update process can be rewritten as 
\begin{align}\label{eq:lm}
\w_{t+1} = \w_t - \frac{1}{K}\sum_{k \in [K]}\m_{t+1,k} = \w_t - \frac{\eta}{K}\sum_{k \in [K]}\nabla f(\w_t;\IM_{t,k}) - \frac{\beta}{K}\sum_{k \in [K]}\m_{t,k}.
\end{align}
Since $\frac{1}{K}\sum_{k \in [K]}\m_{t,k} = -(\w_{t} - \w_{t-1})$ and $\frac{1}{K}\sum_{k \in [K]}\nabla f(\w_t;\IM_{t,k}) = \nabla f(\w_t;\IM_{t})$, $\IM_{t} = \cup_{k \in [K]} \IM_{t,k}$, this implementation of DMSGD is equivalent to the serial MSGD in~(\ref{eq:momentum SGD}).
$\m_{t,k}, k \in [K]$ is called \emph{local momentum} since it only accumulates local gradient information from worker $k$. 

Another way to implement DMSGD is that each worker parallelly computes some stochastic gradients and then the stochastic gradients of all workers are aggregated to get $\nabla f(\w_t;\IM_t) - \beta(\w_{t} - \w_{t-1})/\eta$ in~(\ref{eq:momentum SGD}). The update process of $\w$ in this way is equivalent to the serial MSGD. We call $-(\w_{t} - \w_{t-1})/\eta$ the \emph{global momentum}, because it captures the global gradient information from all workers.

We can find that both local momentum and global momentum implementations of DMSGD are equivalent to the serial MSGD if no sparse communication is adopted. However, when it comes to adopting sparse communication, things become different. In the later sections, we will demonstrate that global momentum is better than local momentum when implementing sparse communication in DMSGD.

\section{Global Momentum Compression}
Our method \underline{g}lobal \underline{m}omentum \underline{c}ompression~(GMC) mainly performs the following operations iteratively:
\begin{itemize}
  \item Each worker computes $\nabla f(\w_t;\IM_{t,k}) = \frac{1}{b}\sum_{\xi \in \IM_{t,k}}\nabla f(\w_t;\xi)$, where $\IM_{t,k}$ is a mini-batch randomly sampled from $\DM_k$ and $|\IM_{t,k}| = b$;
  \item Each worker computes $\e_{t+\frac{1}{2},k} = \e_{t,k}+\nabla f(\w_t;\IM_{t,k})-\frac{\beta}{\eta} (\w_{t}-\w_{t-1})$, and then sends the sparsified vector $\mathcal{C}(\e_{t+\frac{1}{2},k})$ to the server or other workers;
  \item Each worker updates the error residual $\e_{t+1,k} = \e_{t+\frac{1}{2},k}-\mathcal{C}(\e_{t+\frac{1}{2},k})$;
  \item Update parameter $\w_{t+1} = \w_t -  \frac{\eta}{K}\sum_{k \in [K]}\mathcal{C}(\e_{t+\frac{1}{2},k})$.
\end{itemize}
\subsection{Framework of GMC}

GMC can be easily implemented on the all-reduce distributed framework in which each worker sends the sparsified vector $\mathcal{C}(\e_{t+\frac{1}{2},k})$ to all the other workers, then each worker updates $\w_{t+1}$ after receiving the sparsified vectors from all the other workers.

\begin{algorithm}[t]
  \caption{GMC}
  \label{alg:gmc}
  \begin{algorithmic}[1]
  \STATE \textbf{Input}: sparsification compressor $\mathcal{C}(\cdot)$, number of workers $K$, number of iterations $T$, model parameters $\w_0$, learning rate $\eta$, momentum coefficient $\beta \in [0,1)$, training dataset $\mathcal{D}_k, \forall k \in [K]$;
  \STATE Set $\w_{-1} = \w_{0},\e_{0,k}=\0, \forall k \in [K]$;
  \FOR {iteration $t \in [T]$}
  \STATE \underline{Workers:}
  \FOR {worker $k \in [K]$ parallelly}
  \STATE Randomly pick a mini-batch of training data $\IM_{t,k}\subseteq \DM_k$ with $|\IM_{t,k}| = b$ and compute $\nabla f(\w_t;\IM_{t,k}) = \frac{1}{b}\sum_{\xi \in \IM_{t,k}}\nabla f(\w_t;\xi)$;
  \STATE $\e_{t+\frac{1}{2},k} = \e_{t,k}+\nabla f(\w_t;\IM_{t,k})-\frac{\beta}{\eta} (\w_{t}-\w_{t-1})$;
  \STATE Generate a sparse vector $\mathcal{C}(\e_{t+\frac{1}{2},k})$ and send $\mathcal{C}(\e_{t+\frac{1}{2},k})$ to the server;
  \STATE Update the error residual $\e_{t+1,k} = \e_{t+\frac{1}{2},k}-\mathcal{C}(\e_{t+\frac{1}{2},k})$;
  \STATE Receive $\w_{t+1} - \w_{t}$ from server;
  \STATE Get $\w_{t+1}$ by $\w_{t+1} = \w_t+(\w_{t+1}-\w_t)$;       
  \ENDFOR
  \STATE \underline{Server:}
  \STATE Receive $\mathcal{C}(\e_{t+\frac{1}{2},k})$ from all the workers;
  \STATE $\w_{t+1} = \w_{t} - \eta \frac{1}{K}\sum_{k \in [K]}\mathcal{C}(\e_{t+\frac{1}{2},k})$;
  \STATE Send $\w_{t+1} - \w_t$ to workers;  
  \ENDFOR
  % \STATE \textbf{Output:} $\w_{T}$ 
  \end{algorithmic}
  \end{algorithm}
  
Recently, parameter server~\citep{DBLP:conf/osdi/LiAPSAJLSS14} has been one of the most popular distributed frameworks in machine learning. GMC can also be implemented on the parameter server framework. 
In this paper, we adopt the parameter server framework for illustration. The theories in this paper can also be adapted for the all-reduce framework. 
The details of GMC implemented on the parameter server framework are shown in Algorithm~\ref{alg:gmc}. 
After updating $\w_{t+1}$, the server in GMC will send $\w_{t+1} - \w_t$, rather than $\w_{t+1}$, to workers. 
Since $\mathcal{C}(\e_{t+\frac{1}{2},k})$ is sparse, $\w_{t+1} - \w_t$ is sparse as well. Hence, sending $\w_{t+1} - \w_t$ can reduce the communication cost compared with sending $\w_t$. 
Workers can get $\w_{t+1}$ by $\w_{t+1} = \w_t + (\w_{t+1} - \w_t)$.
\begin{remark}
  There are some other ways to combine momentum and error feedback. For example, we can put the momentum term on the server. However, these ways lead to worse performance than the way adopted in this paper. More discussions can be found in Appendix~\ref{appendix:global momentum}.
\end{remark}
We can find that DGC~\citep{DBLP:conf/iclr/LinHM0D18} is mainly based on the local momentum while GMC is based on the global momentum. Hence, each worker in DGC cannot capture the global information from its local momentum, while that in GMC can capture the global information from the global momentum even if sparse communication is adopted. 
\begin{remark}
  We find that due to the \underline{m}omentum \underline{f}actor \underline{m}asking~(mfm) in DGC~\citep{DBLP:conf/iclr/LinHM0D18}, DGC~(w/ mfm) will degenerate to DSGD rather than DMSGD if sparse communication is not adopted, while GMC will degenerate to DMSGD if sparse communication is not adopted. 
  To make a comprehensive comparison of these methods,  we will compare GMC with two implementations of DGC: DGC~(w/ mfm) and DGC~(w/o mfm). Different from DGC~(w/ mfm), DGC~(w/o mfm) will degenerate to DMSGD if sparse communication is not adopted. 
\end{remark}

\subsection{Benefit of Global Momentum}

To illustrate the benefit of global momentum for sparse communication, we investigate the convergence behavior of these methods when minimizing a simple quadratic objective function $F(\w)$.
Let $n =K= 2$,  $\w=(w^{(0)}, w^{(1)}, \cdots, w^{(d-1)})^T \in \mathbb{R}^d$, $\DM = \{\xi_1, \xi_2\}$, $\DM_1=\{\xi_1\}, \DM_2=\{\xi_2\}$, we define the objective functions as follows:
\begin{align*}
  F_0(\w)&=f(\w;\xi_1) = \sum_{i \in [d]}(d-i)*[w^{(i)}-(i+1)]^2, \\
  F_1(\w)&=f(\w;\xi_2) = \sum_{i \in [d]}(d-i)*[w^{(i)}+(i+1)]^2, \\
  F(\w) &= \frac{1}{2}(F_0(\w)+F_1(\w)) = \sum_{i \in [d]}(d-i)*[(w^{(i)})^2+(i+1)^2].
\end{align*}
It's easy to verify that the global optimal point to minimize $F(\w)$ is $\w^{*}=(0, 0, \cdots, 0)^T \in \RB^d$. The optimal point to minimize local objective function $F_0(\w)$ is $(1, 2, \cdots, d)^T$, while it's $(-1, -2, \cdots, -d)^T$ for $F_1(\w)$.
We run DMSGD, DGC~(w/ mfm), DGC~(w/o mfm) and GMC respectively to solve the optimization problem: $\min_{\w\in \RB^d} F(\w)$. The momentum coefficient $\beta$ is set as $0.9$ and the learning rate is set as 0.005. We use top-$s$ as the sparsification compressor.

\begin{figure*}[!t]
  \centering
  \subfigure[DMSGD]{
    \label{fig:DMSGD_tra}
    \includegraphics[width=3.5cm]{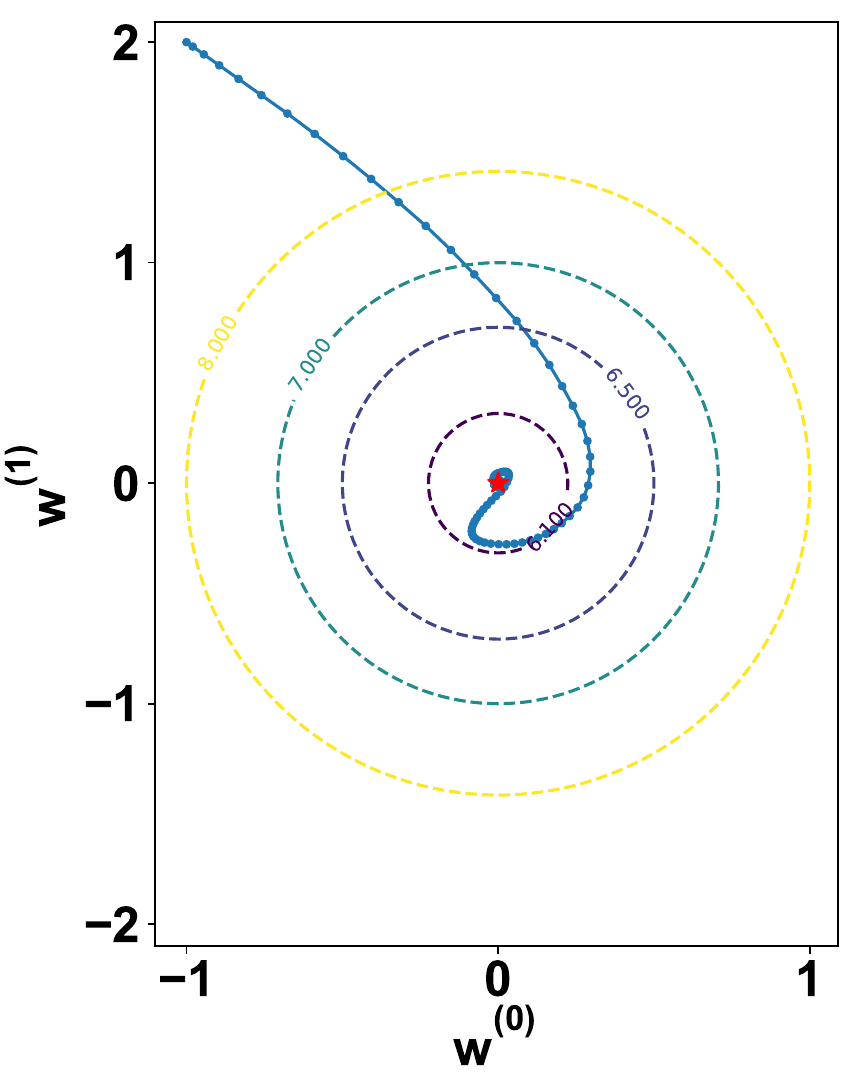}}
  \subfigure[DGC~(w/ mfm)]{
    \label{fig:DGC_topk_tra}
    \includegraphics[width = 3.5cm]{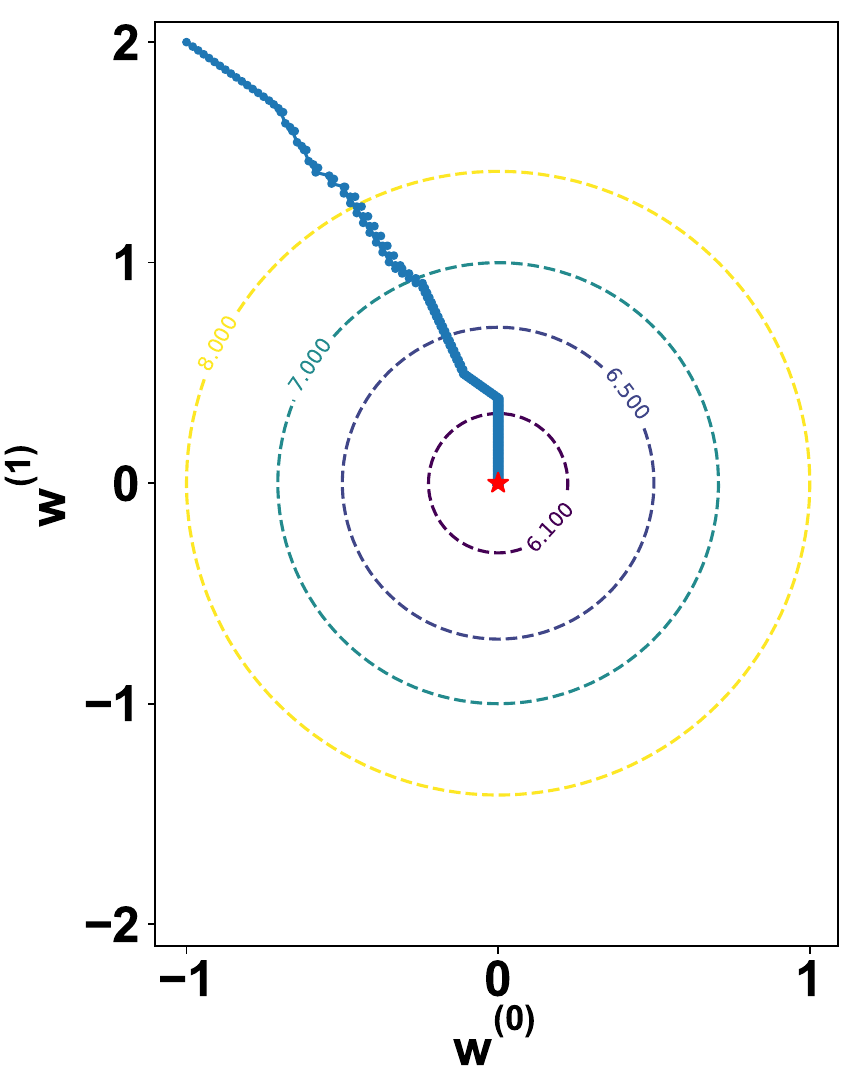}}
  \subfigure[DGC~(w/o mfm)]{
    \label{fig:LMC_topk_tra}
    \includegraphics[width = 3.5cm]{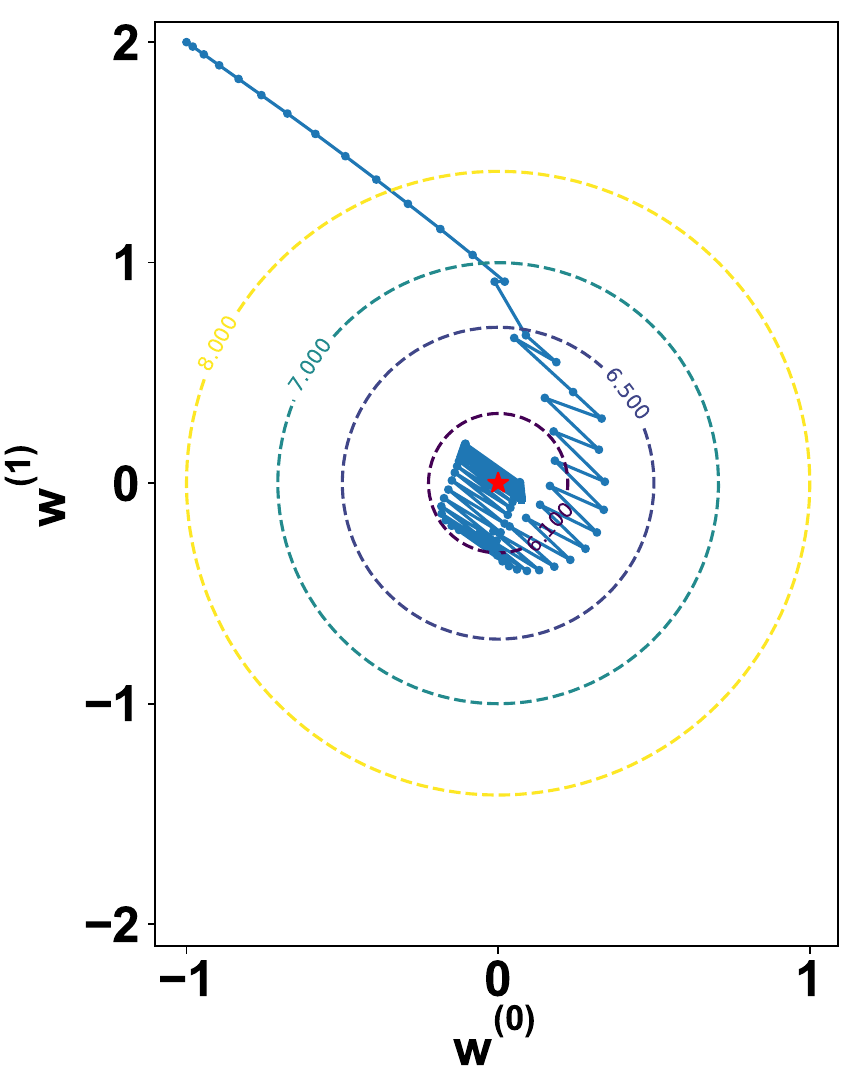}}
  \subfigure[GMC]{
    \label{fig:GMC_topk_tra}
    \includegraphics[width = 3.5cm]{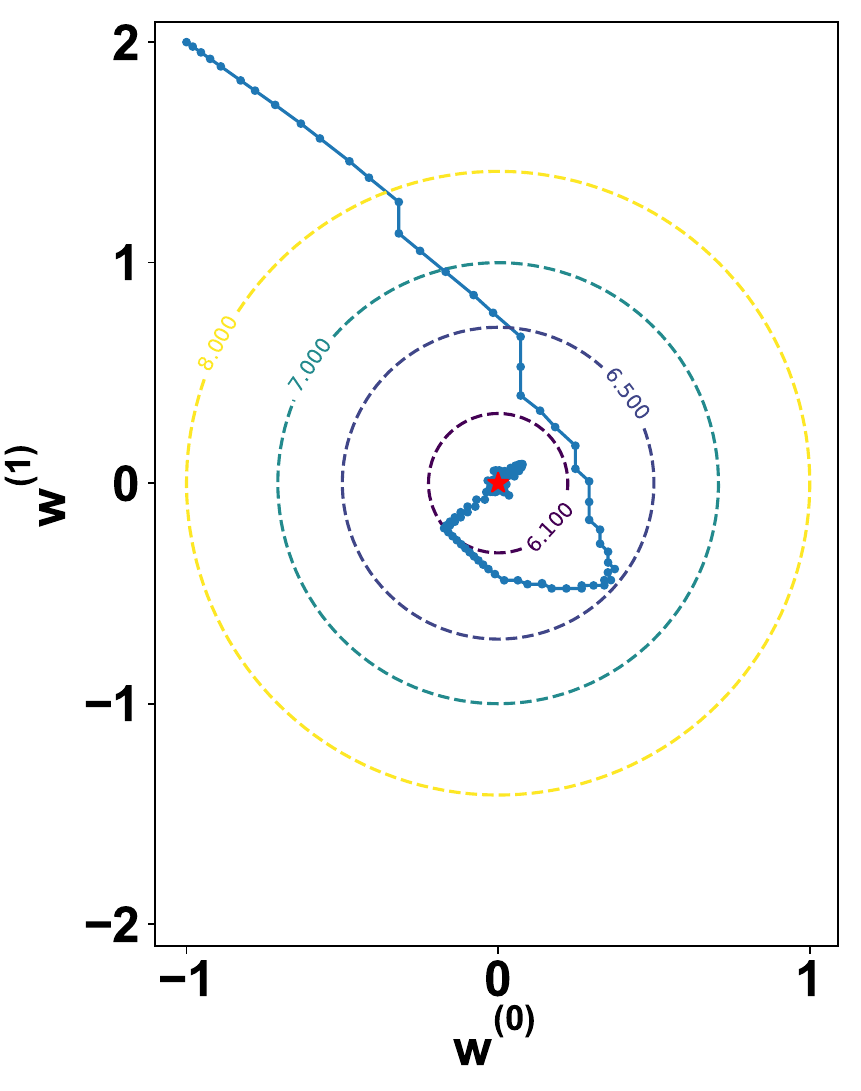}}

\caption{Comparison of optimization trajectories of different methods.}\label{fig:GMC_LMC_DMSGD_top1}
\end{figure*}  

\begin{figure*}[!t]
  \centering
  \subfigure[d=2, s=0.5d]{
    \label{fig:dis1}
    \includegraphics[width=3.5cm]{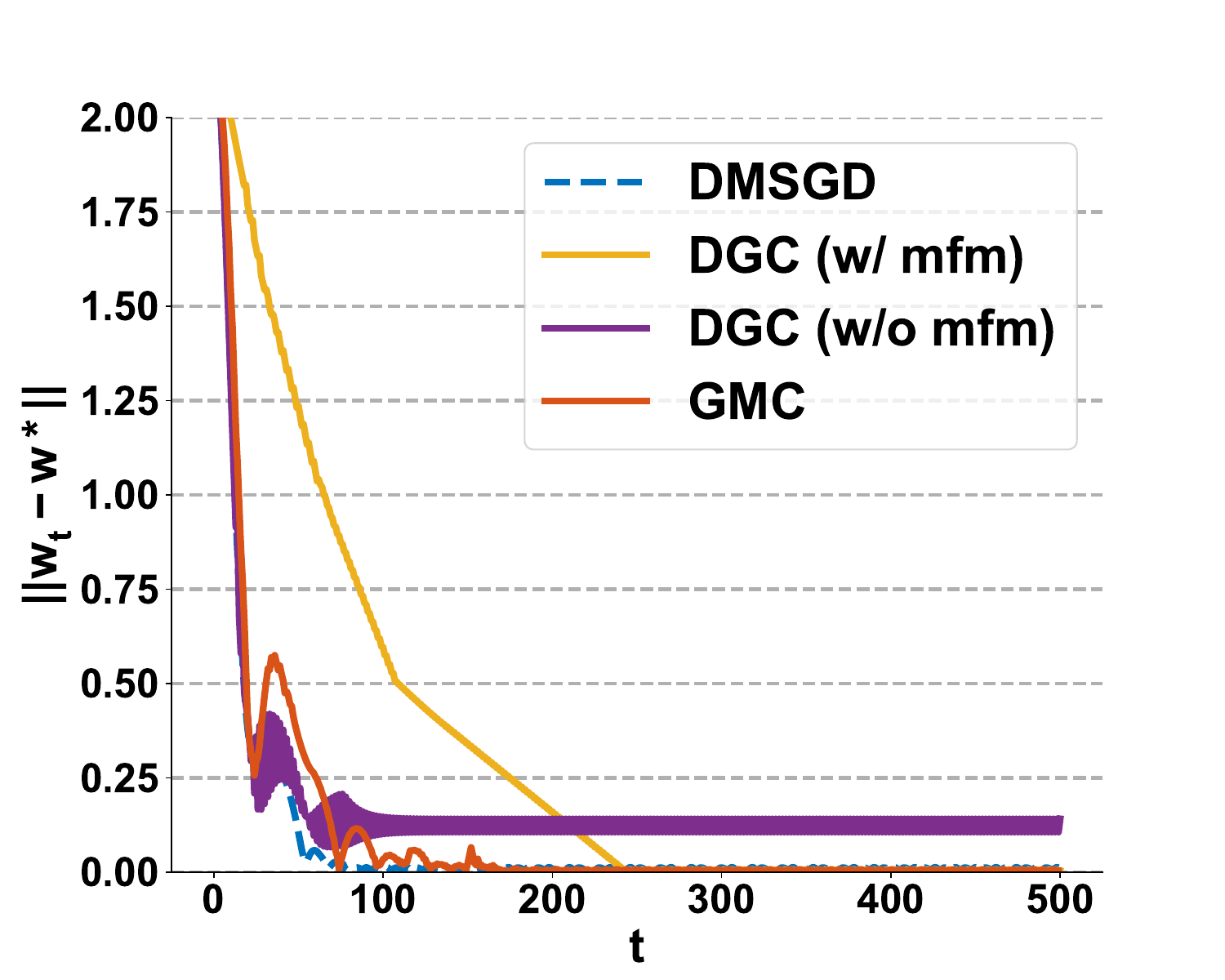}}
  \subfigure[d=20, s=0.1d]{
    \label{fig:dis2}
    \includegraphics[width = 3.5cm]{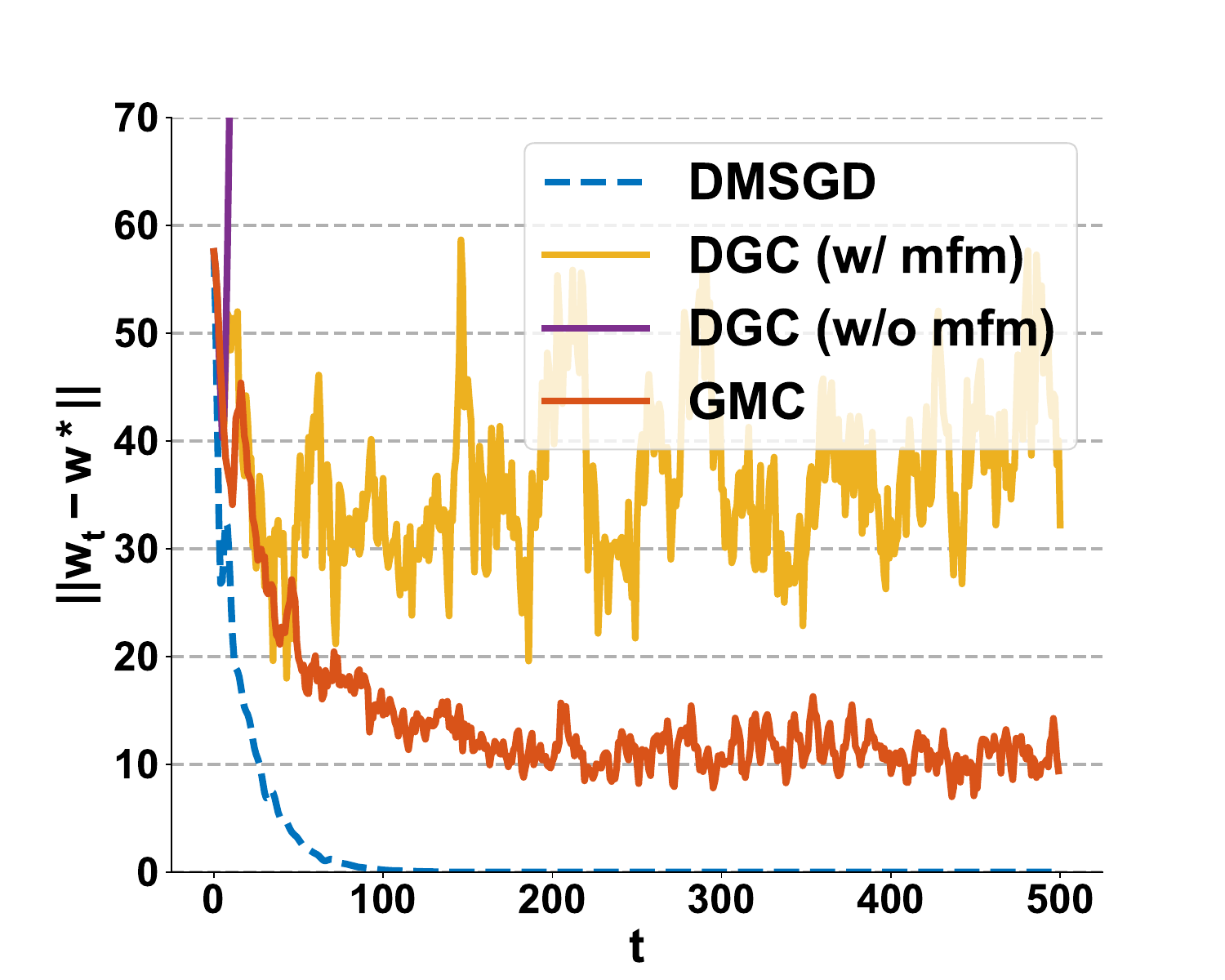}}
  \subfigure[d=20, s=0.5d]{
    \label{fig:dis3}
    \includegraphics[width = 3.5cm]{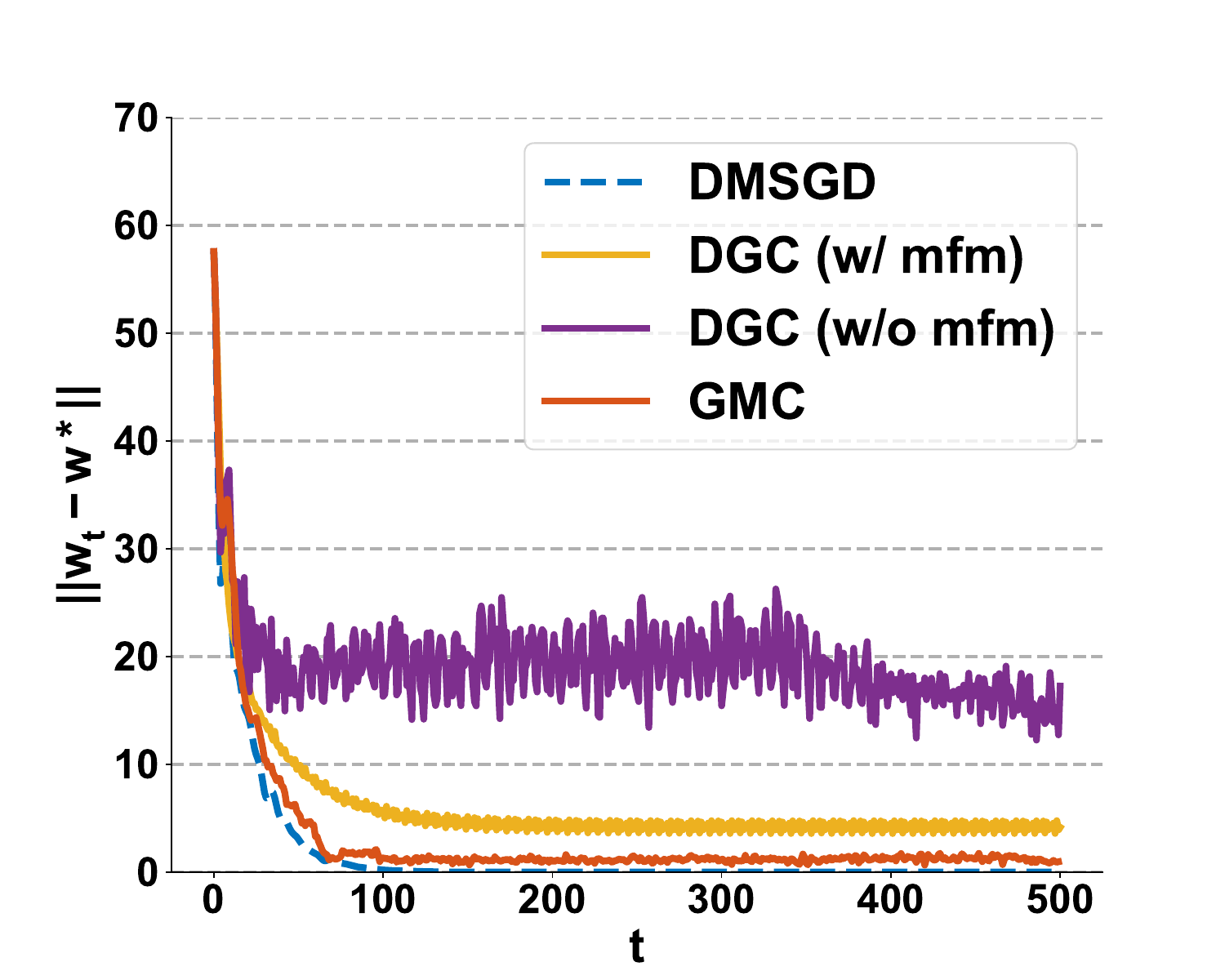}}
  \subfigure[d=20, s=0.8d]{
    \label{fig:dis4}
    \includegraphics[width = 3.5cm]{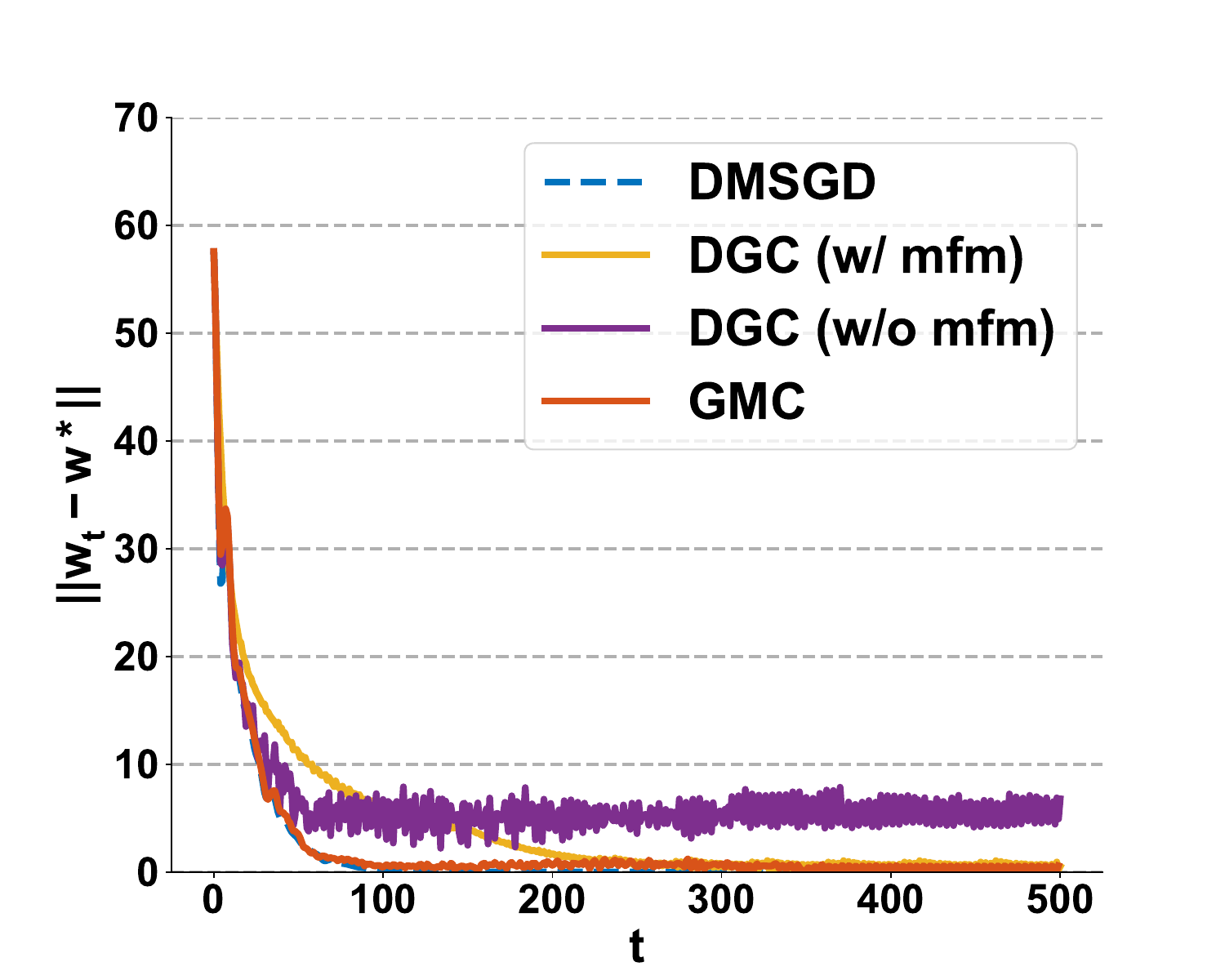}}

\caption{Comparision of the distances to the global optimal point of different methods. }\label{fig:dis}
\end{figure*}    

Firstly, we set $d=2$ and the optimization trajectories of different methods starting from the point $(-1, 2)$ are shown in Figure~\ref{fig:GMC_LMC_DMSGD_top1}. 
We can find that after a sufficient number of iterations, the parameter in DGC~(w/o mfm) can only oscillate within a relatively large neighborhood of the optimal point. Compared with DGC~(w/o mfm), the parameter in GMC converges closer to the optimal point and then remains stable.
Figure~\ref{fig:dis1} shows the distances to the global optimal point during the optimization process. We can find that although the momentum factor masking trick can make the convergence trajectory appear more stable, it also slows down the convergence.
Figure~\ref{fig:dis2}, \ref{fig:dis3} and \ref{fig:dis4} show the distances to the global optimal point when using different $s$ for the case when $d=20$. We can find that, compared with the local momentum methods, the global momentum method GMC converges faster and more stably.

To gain a better intuition for the advantage of global momentum over local momentum, we rethink this simple quadratic function optimization problem by combining the update rule of the momentum term in Section~\ref{sec:DMSGD}.
In local momentum methods, $\m_{t,k}$ only accumulates $\nabla F_k(\w_t)$, which is the gradient of $F_k(\w)$ on worker $k$. 
The gradients of $F_k(\w)$ tend to help the parameter to converge to the optimal point of $F_k(\w)$ instead of the optimal point of $F(\w)$, i.e., local momentum accumulates biased gradients during the optimization
process. As for global momentum, the momentum term $-(\w_{t} - \w_{t-1})/\eta$ contains global information from all the workers. Since we are optimizing the objective function $F(\w)$,  $\w_{t} - \w_{t-1}$ denotes the descent direction of $F(\w)$ with high probability in the next iteration, which will help the parameter to converge to the global optimal point.  
If no sparse communication is adopted, DGC~(w/o mfm) will degenerate to DMSGD, and DGC~(w/ mfm) will degenerate to DSGD.
But if we adopt sparse communication, only a few components of the vectors will be communicated. The biases of the local momentums will disrupt the convergence because of the existence of compressed errors. 

We have only considered a simple quadratic function optimization problem here. When it comes to deep model training, the objective functions are typically high-dimensional, non-convex, and characterized by numerous local minima and saddle points, which are much more complex than the above example.
Furthermore, when we distribute the training across multiple workers, the local objective functions may differ from each other due to the heterogeneous training data distribution. In Section~\ref{experiments}, we will demonstrate that the global momentum method outperforms its local momentum counterparts in distributed deep model training. 

\subsection{Convergence of GMC}

In this section, we prove the convergence of GMC for non-convex problems. % The detailed proofs are put in Appendix~\ref{appendix:efgmc proof}. 
Here, we only present the main Lemmas and Theorems, and the proof details can be found in Appendix~\ref{CAGMC}.
Firstly, we make the following assumptions, which are widely used in existing communication compression methods~\citep{DBLP:conf/nips/0001DKD19, DBLP:conf/nips/XieZKGLL20, koloskova2019decentralized, DBLP:conf/icml/XuH22}:

\begin{assumption}\label{assu:compressor}
  For the compressor $\mathcal{C}: \mathbb{R}^d \rightarrow \mathbb{R}^d$, we assume that it's a $\delta$-approximate compressor, i.e.,  there exists a constant $\delta \in (0,1]$, such that $$\mathbb{E}_{\mathcal{C}}\|\mathcal{C}(\w)-\w\|^2 \leq (1-\delta)\|\w\|^2, \forall \w \in \mathbb{R}^d.$$% and $\mathbb{E}_{\mathcal{C}}\|\mathcal{C}(\v)\|^2 \leq \|\v\|^2$. 
\end{assumption}
\begin{assumption}\label{assu:unbiased gradient}
  For any stochastic gradient $\nabla f(\w; \xi)$, where $\xi$ is randomly sampled from local dataset $\DM_k$ on worker $k$, $\forall 
  k \in [K]$, we assume it's unbiased and variance bounded, i.e.,
   $$\EB_{\xi \thicksim \DM_k}[\nabla f(\w; \xi)]=\nabla F_k(\w), \forall \w \in \RB^d, \forall k \in [K]; $$
   $$\EB_{\xi \thicksim \DM_k}\|\nabla f(\w; \xi)-\nabla F_k(\w)\|^2 \leq \sigma^2, \forall \w \in \RB^d, \forall k \in [K].$$
  Furthermore, the second moment of the full gradient is also bounded, i.e.,
  $$\|\nabla F_k(\w)\|^2 \leq M^2, \forall \w \in \RB^d, \forall k \in [K].$$
  \end{assumption}
It implies that the stochastic gradients are bounded: $\mathbb{E}_{\xi \thicksim \mathcal{D}_k}\|\nabla f(\w;\xi)\|^2 \leq G^2 \triangleq \sigma^2 + M^2, \forall \w \in \RB^d, \forall k \in [K]$.

\begin{assumption}\label{assu:smooth function}
	For any $k \in [K]$, $F_k(\w)$ is $L$-smooth~($L>0$):
  $$F_k(\w) - F_k(\w') - \nabla F_k(\w')^T(\w - \w') \leq \frac{L}{2}\|\w - \w'\|^2, \forall \w,\w'\in \RB^d, \forall k \in [K].$$
\end{assumption}

\begin{assumption}\label{assu:lower bounded function}
  The objective function $F(\w)$ is lower bounded by $F^*$: $F(\w)\geq F^*, \forall \w \in \mathbb{R}^d$.
\end{assumption}

We define $\nabla f(\w_t;\IM_{t}) = \frac{1}{K}\sum_{k \in [K]}\nabla f(\w_t;\IM_{t,k})$, $\bar{\e}_t = \frac{1}{K}\sum_{k \in [K]}\e_{t,k}$. By eliminating $\mathcal{C}(\cdot)$, the update rule in Algorithm~\ref{alg:gmc} can be rewritten as:
\begin{align} \label{eq:update rule}
  \w_{t+1} = \w_{t}- \eta \nabla f(\w_t;\IM_{t})-\eta \bar{\e}_{t}+\eta \bar{\e}_{t+1} + \beta (\w_{t}-\w_{t-1}).
  \end{align}
We can find that if we do not use the sparse communication technique, then $\bar{\e}_{t} = \0$ and (\ref{eq:update rule}) is the same as MSGD in~(\ref{eq:momentum SGD}).

We introduce an auxiliary variable $\z_t$ and have the following lemma:
\begin{lemma} \label{lemma:w to z}
  Let $\z_t \triangleq \w_{t}+\frac{\beta}{1-\beta}(\w_{t} - \w_{t-1}) - \frac{\eta}{1-\beta} \bar{\e}_{t}$, then we have:
  \begin{align}\label{eq:update of z}
    \z_{t+1} = \z_{t}-\frac{\eta}{1-\beta}\nabla f(\w_t;\IM_{t}).  
  \end{align}
\end{lemma}

Equation~(\ref{eq:update of z}) is similar to the update equation in SGD. Inspired by this, we only need to prove the convergence of $\z_t$ and bound the gap $\|\z_t - \w_t\|$. 

The error residual $\e_{t,k}$ on each worker has the following property:
\begin{lemma} \label{lemma:bounded error}
  With Assumption~\ref{assu:compressor}, \ref{assu:unbiased gradient}, if $\beta \leq \frac{\delta}{4\sqrt{2+\delta}}$, the error residual can be bounded:       
    $$\frac{1}{K}\sum_{k \in [K]}\mathbb{E}\|\e_{t, k}\|^2 \leq E^2,$$
  where $E^2 = \frac{(1-\delta)(1+\frac{4}{\delta})G^2}{1-[(1-\frac{\delta}{4})(1-\frac{\beta}{K})^2+(1+\frac{\delta}{4})(1+\frac{2}{\delta})\beta^2]}.$
  \end{lemma}

  For the auxiliary variable $\z_t$ in Lemma \ref{lemma:w to z}, the gap between $\z_t$ and $\w_t$ has the following property:
\begin{lemma}\label{lemma:bounded gap}
  The gap between $\z_t$ and $\w_t$ can be bounded: $\mathbb{E}\|\z_t - \w_t\|^2 \leq C_1\eta^2$, where $C_1=\frac{6 (2E^2+G^2)\beta^2}{(1-\beta)^4} +\frac{2E^2}{(1-\beta)^2}$.
\end{lemma}

Then we have the following convergence result:
\begin{theorem} \label{theorem:gmc}
  With Assumptions~\ref{assu:compressor}, \ref{assu:unbiased gradient}, \ref{assu:smooth function} and \ref{assu:lower bounded function}, if $\beta \leq \frac{\delta}{4\sqrt{2+\delta}}$ and $\eta \leq \frac{1-\beta}{2L}$, Algorithm~\ref{alg:gmc} has the following convergence rate:
  $$ \frac{1}{T}\sum_{t \in [T]}\mathbb{E}\|\nabla F(\w_{t})\|^2\leq  \frac{4(1-\beta)(F(\z_{0})-F^*)}{T\eta} + \frac{2L\sigma^2}{(1-\beta)b}\frac{\eta}{K}+ 2C_1 L^2\eta^2, $$ where $C_1 = \frac{6 (2E^2+G^2)\beta^2}{(1-\beta)^4} +\frac{2E^2}{(1-\beta)^2}$.
\end{theorem}
\begin{remark}
  By taking $\eta = \mathcal{O}(\sqrt{\frac{K}{T}})$ and assuming that $T$ is large enough~(i.e., $T=\Omega(K^3)$), we can get the convergence rate of GMC to a critical point: $\frac{1}{T}\sum_{t \in [T]}\mathbb{E}\|\nabla F(\w_{t})\|^2\leq  \mathcal{O}(\frac{1}{\sqrt{KT}})$, which is the same as that of vanilla MSGD. 
  It also indicates that the convergence rate of GMC has a linear speedup guarantee with respect to the number of workers $K$. 
\end{remark}
\begin{remark}
  Note that we impose a constraint on the momentum coefficient $\beta$ during the theoretical proof. But in practice, even when the constraint is relaxed, e.g., $\beta=0.9$,
  GMC still converges well. More details about the convergence performance of GMC are provided in Section~\ref{experiments}.
\end{remark}
\section{GMC+}
The most widely used sparsification compressor adopted in sparse communication methods is top-$s$, which selects $s$ largest components according to their absolute values.
However, the top-$s$ compressor requires extra computation overhead to find the largest components and extra communication overhead to communicate the indices of the components. Some works~\citep{DBLP:conf/nips/VogelsKJ19, DBLP:conf/nips/XieZKGLL20, DBLP:conf/icml/XuH22} consider Random Blockwise Gradient Sparsification~(RBGS) compressor, which randomly selects a block which contains $s$ components using the same random seed among the workers.

\begin{algorithm}[t]
  \caption{GMC+}
  \label{alg:defgmc}
  \begin{algorithmic}[1]
    \STATE \textbf{Input}: sparsification compressor $\mathcal{C}(\cdot)$, number of workers $K$, number of iterations $T$, model parameters $\w_0$, learning rate $\eta$, detached coefficient $\lambda \in [0,1]$, momentum coefficient $\beta \in [0,1)$, training dataset $\mathcal{D}_k, \forall k \in [K]$;
  \STATE Set $\w_{-1} = \w_{0}$, $\w_{0, k}=\w_0$, $\e_{0, k}=\0, \forall k \in [K]$;
  \FOR {iteration $t \in [T]$}
  \STATE \underline{Workers:}
  \FOR {worker $k \in [K]$ parallelly}
  \STATE Randomly pick a mini-batch of training data $\IM_{t,k}\subseteq \DM_k$ with $|\IM_{t,k}| = b$ and compute $\nabla f(\w_{t,k};\IM_{t,k}) = \frac{1}{b}\sum_{\xi \in \IM_{t,k}}\nabla f(\w_{t,k};\xi)$;
  \STATE $\e_{t+\frac{1}{2},k} = \e_{t,k}+\nabla f(\w_{t,k};\IM_{t,k})-\frac{\beta}{\eta} (\w_{t}-\w_{t-1})$;
  \STATE Generate a sparse vector $\mathcal{C}(\e_{t+\frac{1}{2},k})$ and send $\mathcal{C}(\e_{t+\frac{1}{2},k})$ to the server;
  \STATE Update the error residual $\e_{t+1,k} = \e_{t+\frac{1}{2},k}-\mathcal{C}(\e_{t+\frac{1}{2},k})$;
  \STATE Receive $\w_{t+1} - \w_{t}$ from server;
  \STATE Get $\w_{t+1}$ by $\w_{t+1} = \w_t+(\w_{t+1}-\w_t)$;      
  \STATE $\w_{t+1,k} = \w_{t+1}-\lambda \eta \e_{t+1,k}$              \hfill{//detach}
  \ENDFOR
  \STATE \underline{Server:}
  \STATE Receive $\mathcal{C}(\e_{t+\frac{1}{2},k})$ from all the workers;
  \STATE $\w_{t+1} = \w_{t} - \eta \frac{1}{K}\sum_{k \in [K]}\mathcal{C}(\e_{t+\frac{1}{2},k})$;
  \STATE Send $\w_{t+1} - \w_t$ to workers;  
  \ENDFOR
  % \STATE \textbf{Output:} $\w_{T}$ 
  \end{algorithmic}
  \end{algorithm}

Due to the larger compressed error introduced by RBGS compared with top-$s$ when selecting the same number of components of the original vector to communicate, vanilla error feedback methods usually fail to converge. \cite{DBLP:conf/icml/XuH22} propose DEF-A to solve the convergence problem by using detached error feedback~(DEF) technique~\footnote{\cite{DBLP:conf/icml/XuH22} proposes two algorithms: DEF and DEF-A. Since DEF-A enhances the generalization performance of DEF, we only consider DEF-A in this paper.}.
The momentum variant of DEF-A in~\citep{DBLP:conf/icml/XuH22} uses local momentum. 
We improve DEF-A by changing its local momentum to global momentum, getting a new method called GMC+. The detail of GMC+ is shown in Algorithm~\ref{alg:defgmc}. 
We also adopt parameter server architecture for illustration. GMC+ can also be easily implemented on all-reduce frameworks. 

% 讨论GMC+与GMC的主要区别
Different from GMC, each worker in GMC+ evaluates the gradient at $\w_{t,k}$ rather than the current parameter $\w_t$. $\w_{t,k}$ is a point detached from the current parameter $\w_{t}$ using the error residual $\e_{t,k}$ with a tuned hyperparameter $\lambda$. When the detached coefficient $\lambda$ is set as 0, GMC+ will degenerate to GMC.

  We also provide the convergence analysis for GMC+. Here, we only present the main Lemmas and Theorems, and the detailed proof can be found in Appendix~\ref{CAGMC+}.
  We define $\bar{\w}_t = \frac{1}{K}\sum_{k \in [K]}\w_{t,k}$, $\bar{\e}_t = \frac{1}{K}\sum_{k \in [K]}\e_{t,k}$. The update rule of $\bar{\w}_t$ in Algorithm~\ref{alg:defgmc} can be rewritten as
  \begin{align*}
  \bar{\w}_{t+1}= \bar{\w}_{t}+\beta (\bar{\w}_{t}-\bar{\w}_{t-1})+ \eta (1-\lambda)(\bar{\e}_{t+1}-\bar{\e}_{t}) +\beta\lambda \eta (\bar{\e}_{t}-\bar{\e}_{t-1})-\eta\frac{1}{K}\sum_{k \in [K]}\nabla f(\w_{t,k};\IM_{t,k}).
  \end{align*} 
  The error residual $\e_{t,k}$ in Algorithm~\ref{alg:defgmc} has the same property as that in Algorithm~\ref{alg:gmc}:
\begin{lemma} \label{lemma:bounded error of def}
  With Assumption~\ref{assu:compressor}, \ref{assu:unbiased gradient}, if $\beta \leq \frac{\delta}{4\sqrt{2+\delta}}$, the error residual can be bounded:       
    $$\frac{1}{K}\sum_{k \in [K]}\mathbb{E}\|\e_{t, k}\|^2 \leq E^2,$$
  where $E^2 = \frac{(1-\delta)(1+\frac{4}{\delta})G^2}{1-[(1-\frac{\delta}{4})(1-\frac{\beta}{K})^2+(1+\frac{\delta}{4})(1+\frac{2}{\delta})\beta^2]}.$
  \end{lemma}
  \begin{lemma} \label{lemma:w to barz}
    Let $\bar{\z}_t \triangleq \bar{\w}_{t}+\frac{\beta}{1-\beta}(\bar{\w}_{t} - \bar{\w}_{t-1}) - \frac{1}{1-\beta}[(1-\lambda) \eta \bar{\e}_{t}+ \beta\lambda \eta \bar{\e}_{t-1}]$, then we have:
    \begin{align*}
      \bar{\z}_{t+1} = \bar{\z}_{t}-\frac{\eta}{1-\beta}\frac{1}{K}\sum_{k \in [K]}\nabla f(\w_{t,k};\IM_{t,k}).  
    \end{align*}
  \end{lemma}
  \begin{lemma} \label{lemma:bounded bargap}
    The gap between $\bar{\z}_t$ and $\bar{\w}_t$ can be bounded: $\mathbb{E}\|\bar{\z}_t - \bar{\w}_t\|^2 \leq C_2\eta^2,$ where $C_2=\frac{20\beta^2((1-\lambda)^2+\beta^2\lambda^2)E^2+10\beta^2G^2}{(1-\beta)^4} +\frac{4E^2((1-\lambda)^2+\beta^2\lambda^2)}{(1-\beta)^2}$.
  \end{lemma}
\begin{theorem} \label{theorem:defgmc}
  With Assumptions~\ref{assu:compressor}, \ref{assu:unbiased gradient}, \ref{assu:smooth function} and \ref{assu:lower bounded function}, if $\beta \leq \frac{\delta}{4\sqrt{2+\delta}}$and $\eta \leq \frac{3(1-\beta)}{4L}$, Algorithm~\ref{alg:defgmc} has the following convergence rate:
  $$     \frac{1}{T}\sum_{t \in [T]}\mathbb{E}\|\nabla F(\w_t)\|^2 \leq  \frac{2(1-\beta) (F(\bar{\z}_{0})-F^{*})}{T\eta}+\frac{L\sigma^2}{(1-\beta)b}\frac{\eta}{K} +(8C_2 +10\lambda^2E^2)L^2\eta^2, $$ where $C_2=\frac{20\beta^2((1-\lambda)^2+\beta^2\lambda^2)E^2+10\beta^2G^2}{(1-\beta)^4} +\frac{4E^2((1-\lambda)^2+\beta^2\lambda^2)}{(1-\beta)^2}$.
\end{theorem}

\begin{remark}
  By taking $\eta = \mathcal{O}(\sqrt{\frac{K}{T}})$ and assuming that $T$ is large enough~(i.e., $T=\Omega(K^3)$), we can get the convergence rate of GMC+ to a critical point: $\frac{1}{T}\sum_{t \in [T]}\mathbb{E}\|\nabla F(\w_{t})\|^2\leq  \mathcal{O}(\frac{1}{\sqrt{KT}})$, which is the same as that of GMC.
\end{remark}

\begin{remark}
  Note that the convergence guarantee of DEF-A and its momentum variant for non-convex problems is lacking in~\citep{DBLP:conf/icml/XuH22}. We provide the convergence analysis for GMC+, which can be seen as a global momentum variant of DEF-A. We eliminate the assumption of ring-allreduce compatibility from~\citep{DBLP:conf/icml/XuH22} and only assume that the compressor has the $\delta$-approximate property. This makes our convergence analysis for GMC+ applicable to a broader range of compressors, such as top-$s$, which is not ring-allreduce compatible.
\end{remark}

\section{Experiments} \label{experiments}

In this section, we evaluate the performance of GMC and other baselines in image classification tasks. The experiments
are conducted on a distributed platform with dockers. Each docker has access to one V100 GPU. All the experiments
are implemented based on the parameter server framework.

Since the server is typically the busiest node in parameter server architecture, we consider the communication cost on the server in our experiments. 
For DMSGD which doesn't use any communication compression techniques, the communication cost on the server includes receiving vectors from the $K$ workers and sending one vector to the $K$ workers for every iteration. So the communication cost of DMSGD is $2dKT$, where $d$ is the dimension of the model parameter,  $T$ is the number of iterations and $K$ is the number of workers. 
In sparse communication methods, since only a few components of the original vector will be communicated, workers typically communicate the sparsified vectors with the server by using the structure of $(index, value)$ to denote each component. The communication cost for communicating  each $(index, value)$ will be twice that for only communicating $value$ in DMSGD. 
For a sparse communication algorithm, we define the \textit{relative communication cost}~(RCC) as the ratio of its communication cost on the server to that of the DMSGD algorithm. 
We take GMC as an example to further illustrate how RCC is calculated. 
In GMC, the communication cost of receiving sparsified vectors from workers is $2\sum_{t \in [T]}\sum_{k \in [K]}\|\mathcal{C}(\e_{t+\frac{1}{2},k})\|_0$ and the communication cost of sending aggregated sparsified vectors to workers is $2K\sum_{t \in [T]}\|\w_{t+1} - \w_t\|_0$.
Hence, the total communication cost of GMC during the whole training process is $2(\sum_{t \in [T]}(\sum_{k \in [K]}\|\mathcal{C}(\e_{t+\frac{1}{2},k})\|_0+K\|\w_{t+1} - \w_t\|_0))$. The RCC of GMC is:
\begin{align} 
	\mbox{RCC}=\frac{1}{dKT}\sum_{t \in [T]}(K\|\w_{t+1} - \w_t\|_0 + \sum_{k \in [K]}\|\mathcal{C}(\e_{t+\frac{1}{2},k})\|_0).
\end{align}
The RCC of DMSGD is $100\%$~(no compression). Here, all numbers have the same unit~(float value).

We use the CIFAR10 and CIFAR100 datasets under both IID and non-IID data distribution. For the IID scenario, the training data is randomly assigned to each worker. For the non-IID scenario, we use Dirichlet distribution with parameter 0.1 to partition the training data as in~\citep{DBLP:journals/corr/abs-1909-06335, DBLP:conf/icml/0004KSJ21}.
We adopt two popular deep models: ResNet20~\citep{DBLP:conf/cvpr/HeZRS16} and Vision Transformer~(ViT)~\citep{DBLP:journals/corr/abs-2112-13492} with four Transformer blocks. Although Batch Normalization~(BN) in ResNet20 is effective in practice, it is known to be problematic in the non-IID setting due to its dependence on the estimated mean and variance~\citep{DBLP:conf/icml/HsiehPMG20}. We replace BN in ResNet20 with Group Normalization~(GN)~\citep{wu2018group} under non-IID data distribution as suggested in~\citep{DBLP:conf/icml/HsiehPMG20, DBLP:conf/icml/0004KSJ21}. We train the models with 200 epochs.

\subsection{Results of GMC}
We compare GMC with DGC~(w/ and w/o mfm)~\citep{DBLP:conf/iclr/LinHM0D18}, CSER~\citep{DBLP:conf/nips/XieZKGLL20} and DEF-A~\citep{DBLP:conf/icml/XuH22}. 
In the experiments of~\citep{DBLP:conf/iclr/LinHM0D18}, DGC gets far better performance on both accuracy and communication cost than quantization methods. Hence, we do not compare with quantization methods in this paper. 
We don't use the warm-up strategy in the experiments. The  momentum coefficient $\beta$ is set as $0.9$. The weight decay is set as 0.0001. We use 8 workers with a total batch size of 128.
We adopt the cosine annealing learning rate decay strategy~\citep{DBLP:conf/iclr/LoshchilovH17}~(without restarts). In the $m$-th epoch, the learning rate is $\eta_m = \eta*0.5*(1+cos(m\pi/200))$.
For all the sparse communication algorithms, the top-$s$ compressor is adopted to sparsify the vectors.  In DGC, DEF-A and GMC, we set $s=\frac{d}{1024}$.  
In CSER, there are two compressors $\mathcal{C}_1$, $\mathcal{C}_2$ and one synchronization interval $H$ that determine RCC together. Following the hyperparameter settings in~\citep{DBLP:conf/nips/XieZKGLL20}, we set $s=\frac{d}{32}$ for $\mathcal{C}_1$, $s=\frac{d}{2048}$ for $\mathcal{C}_2$ and $H=64$.
In DEF-A, the hyperparameter $\lambda$ is set as 0.3 as suggested in~\citep{DBLP:conf/icml/XuH22}.
Considering the computation overhead of the top-$s$ selection, we use an approximate way to implement it~\citep{DBLP:conf/iclr/LinHM0D18}: given a vector $\a \in \RB^d$, we first randomly choose a subset $S\subset [d]$ such that $|S| = \frac{d}{100}$. We get the threshold $\theta$ such that $|\{j||a^{(j)}|\geq \theta, j\in S\}| = \frac{|S|}{1024}$. Then we set $\mathcal{C}(\a)$ by choosing the indices in $\{j||a^{(j)}|\geq \theta, j \in [d]\}$. It implies that $\|\mathcal{C}(\a)\|_0$ is approximately $\frac{d}{1024}$.

\begin{table*}[!t] \small
  \centering
  \caption{Empirical results of different methods under IID data distribution.}\label{tab:GMC-IID}  
  \setlength{\tabcolsep}{1.6mm}{  
  \begin{tabular}{c|c|cccccc}
    \hline    
    \multirow{2}*{Dataset} &\multirow{2}*{Model} & \multirow{2}*{Method}  & DGC & DGC & \multirow{2}*{CSER} &\multirow{2}*{DEF-A} & \multirow{2}*{GMC}   \\
    & &      & (w/ mfm) & (w/o mfm) & ~ & ~& ~   \\ \hline
    \multirow{4}*{CIFAR10}& \multirow{2}*{ResNet20~(BN)} & Accuracy &  \textbf{92.44\%} & 92.30\%   & 91.70\% & 92.29\% & 92.30\%  \\
    ~& ~     & RCC &  0.66\% & 0.66\%   & 0.62\%& 0.65\% & 0.64\%  \\ \cline{2-8}
    ~& \multirow{2}*{ViT} & Accuracy  & \textbf{83.67\%} & 83.35\%   & 76.81\%& 82.67\% & 83.46\%   \\
    ~& ~     & RCC &  0.82\% & 0.81\%   & 0.62\%& 0.81\% & 0.80\%  \\ \hline
    \multirow{4}*{CIFAR100}& \multirow{2}*{ResNet20~(BN)}   & Accuracy  & 68.05\% & 68.66\%   & 66.62\%& 68.72\% & \textbf{68.89\%}  \\
    ~& ~     & RCC & 0.64\% & 0.64\%   & 0.65\%& 0.64\% & 0.64\% \\ \cline{2-8}
    ~& \multirow{2}*{ViT}   & Accuracy &  \textbf{59.28\%} & 59.15\%   & 50.35\%& 56.56\% & \textbf{59.28\%}  \\
    ~& ~     & RCC & 0.80\% & 0.79\%   & 0.65\%& 0.79\% & 0.78\% \\ \hline
  \end{tabular}}
  \end{table*}
  
\begin{figure*}[!t]
  \centering
  \subfigure[ResNet20, CIFAR10]{
    \label{fig:IID1}
    \begin{minipage}[b]{0.23\textwidth}
      \includegraphics[width=1\linewidth]{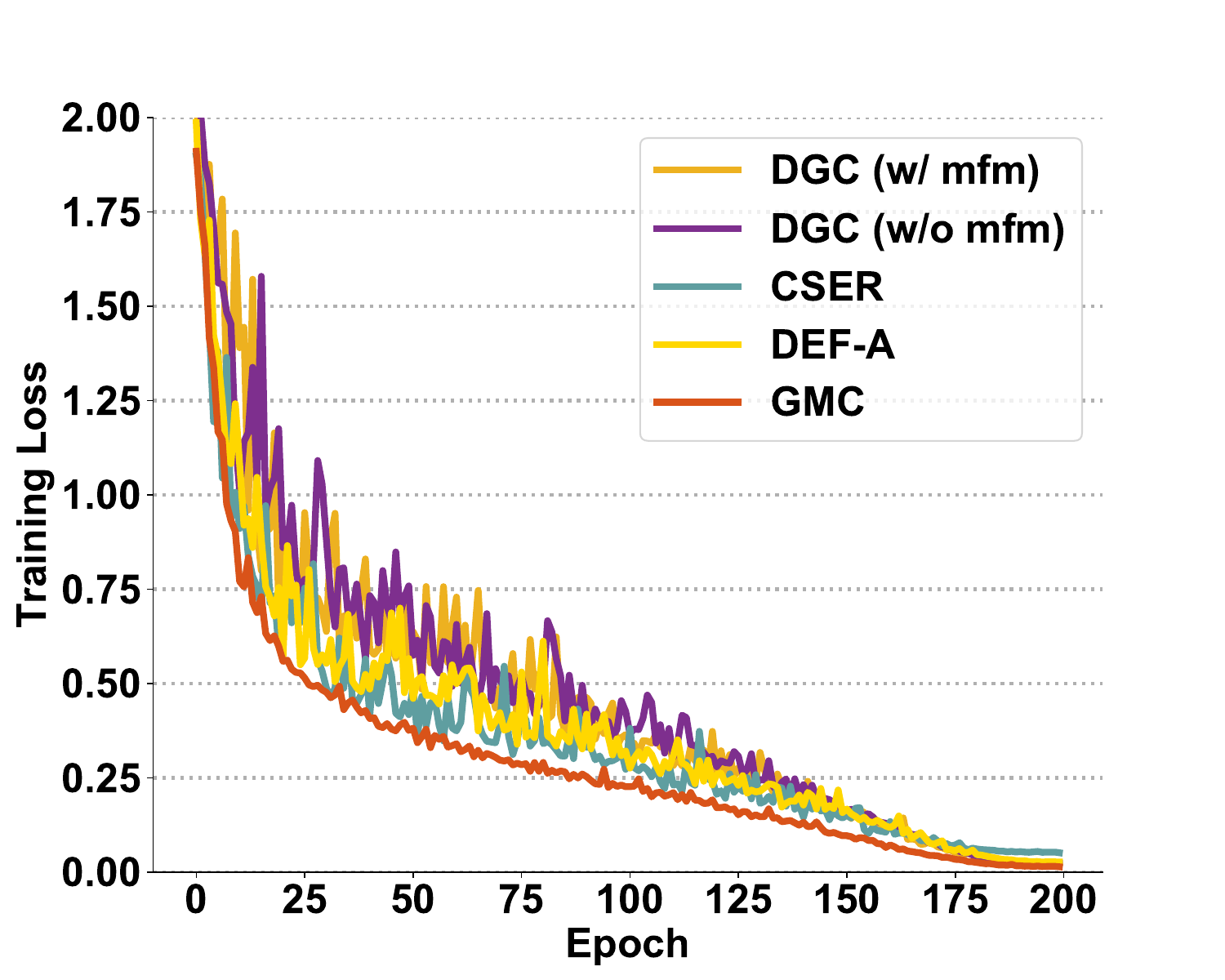}\vspace{-2pt}
      \includegraphics[width=1\linewidth]{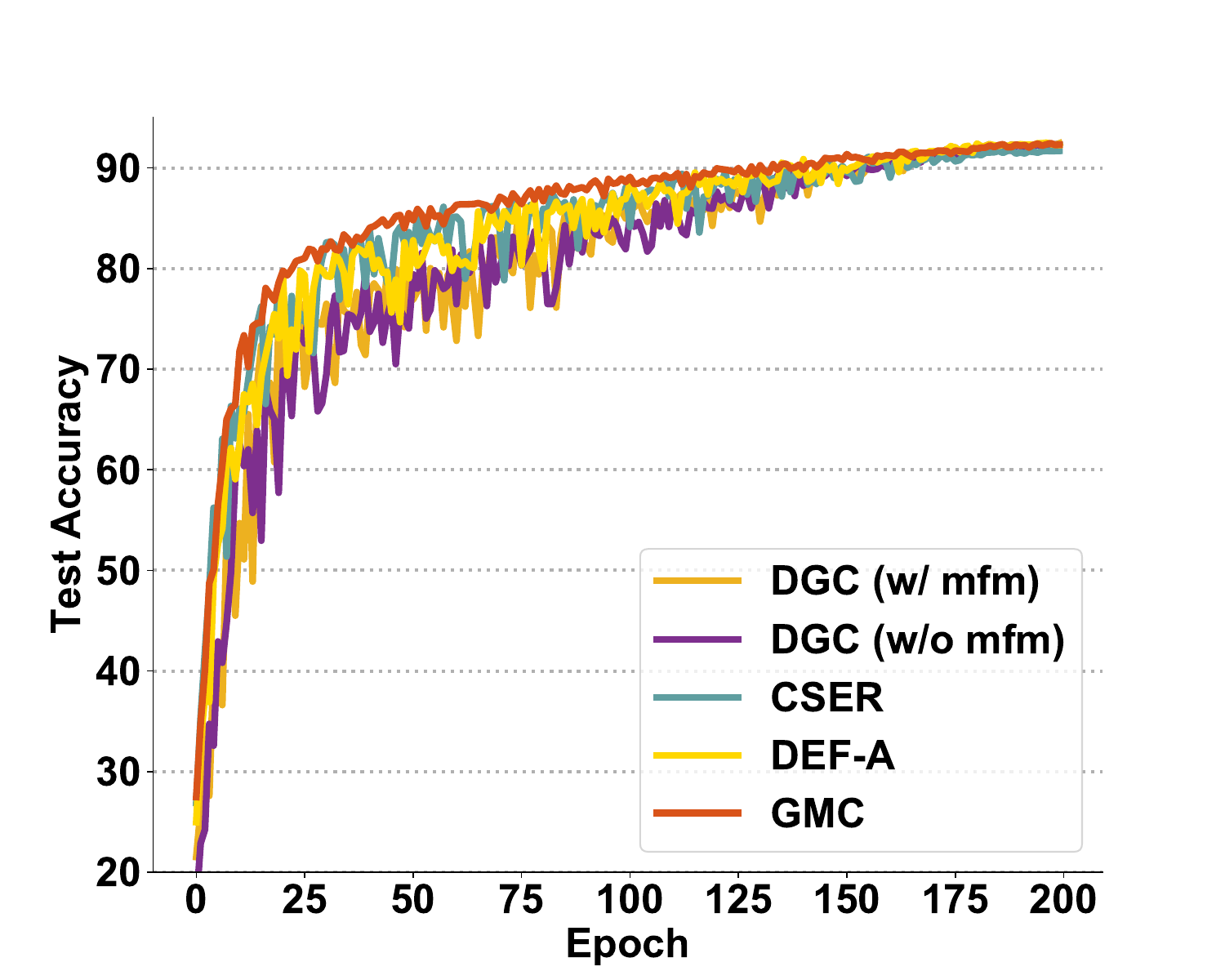}
      \end{minipage}}
  \subfigure[ViT, CIFAR10]{
    \label{fig:IID2}
    \begin{minipage}[b]{0.23\textwidth}
      \includegraphics[width=1\linewidth]{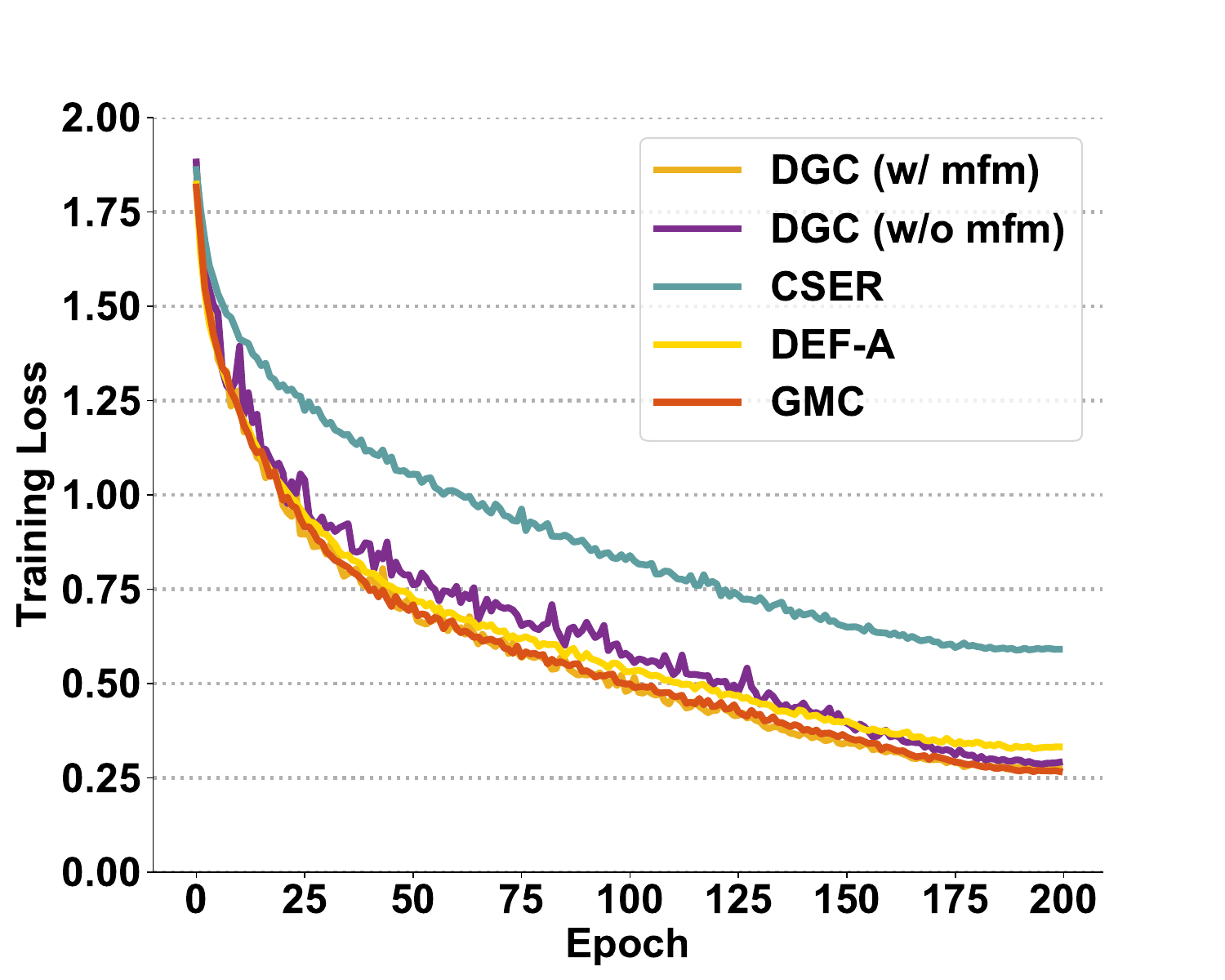}\vspace{-2pt}
      \includegraphics[width=1\linewidth]{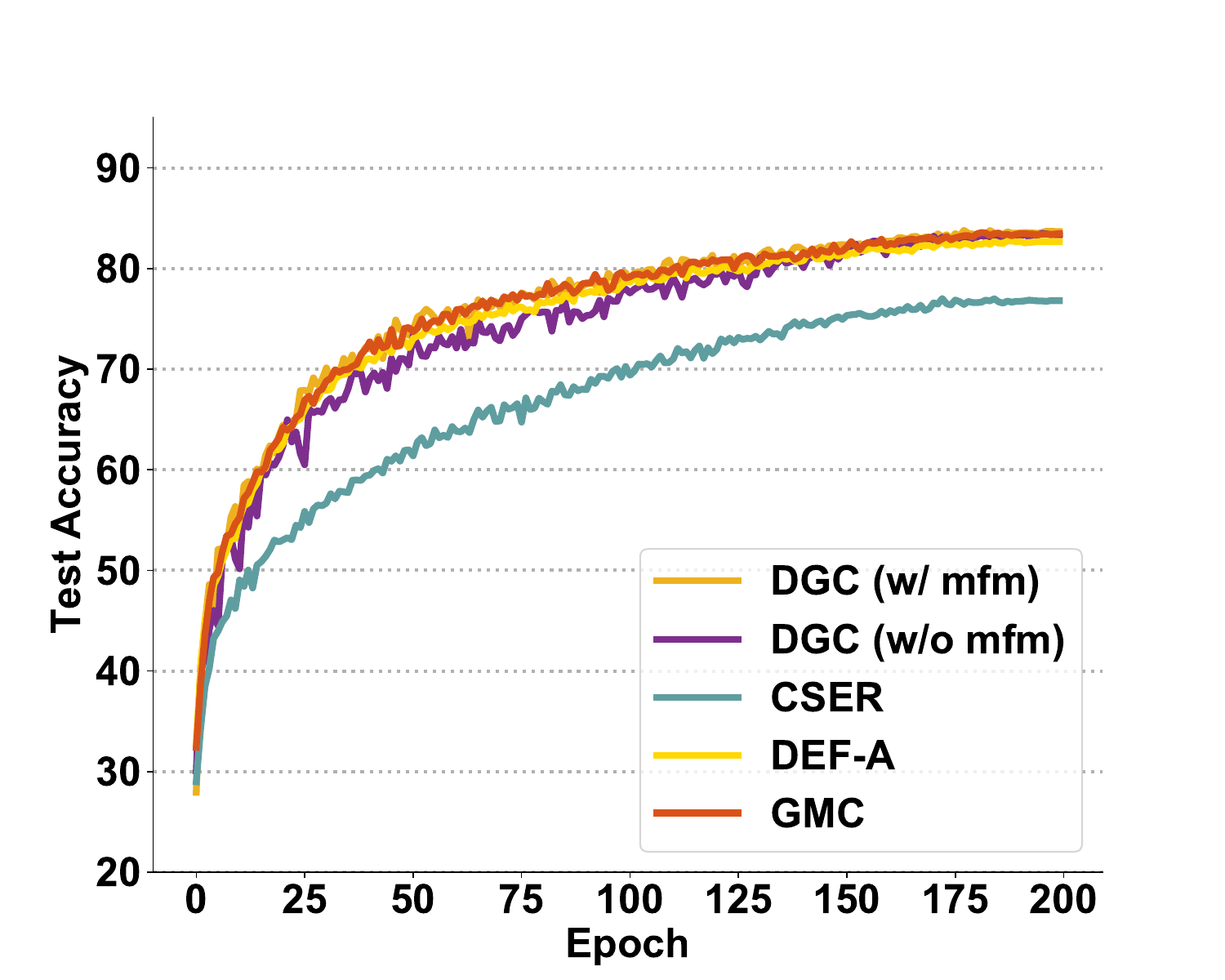}
      \end{minipage}}
  \subfigure[ResNet20, CIFAR100]{
    \label{fig:IID3}
    \begin{minipage}[b]{0.23\textwidth}
      \includegraphics[width=1\linewidth]{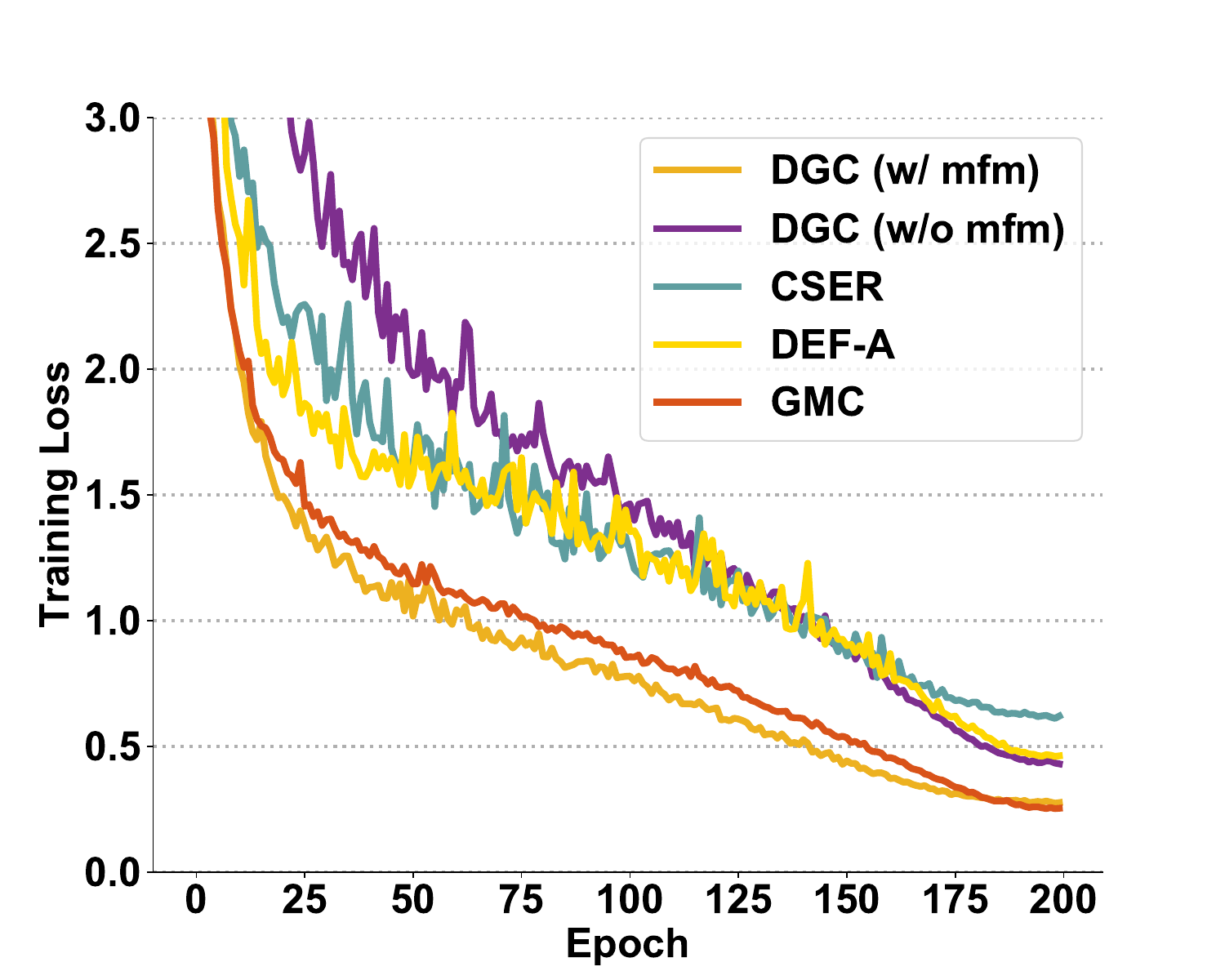}\vspace{-2pt}
      \includegraphics[width=1\linewidth]{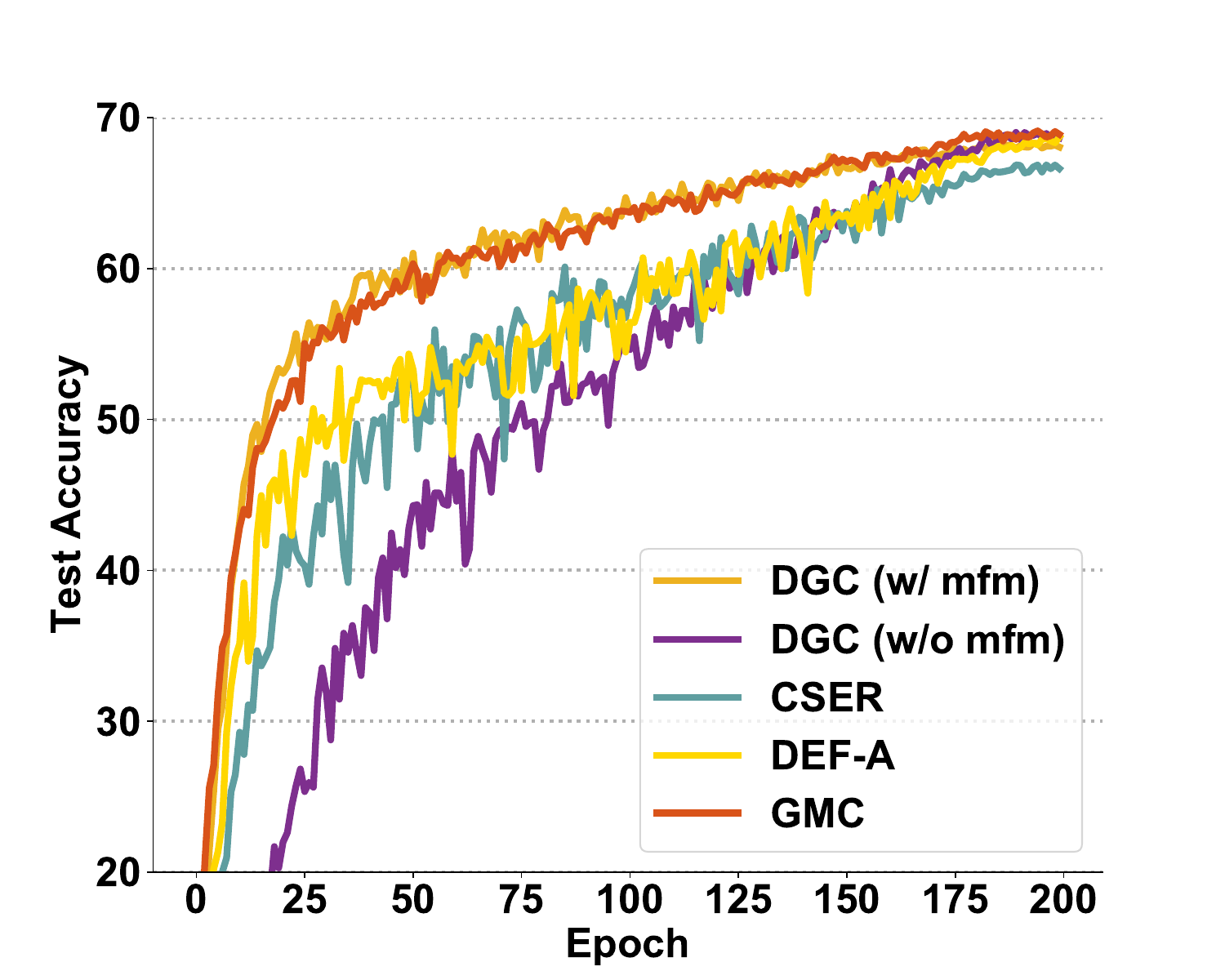}
      \end{minipage}}
  \subfigure[ViT, CIFAR100]{
    \label{fig:IID4}
    \begin{minipage}[b]{0.23\textwidth}
      \includegraphics[width=1\linewidth]{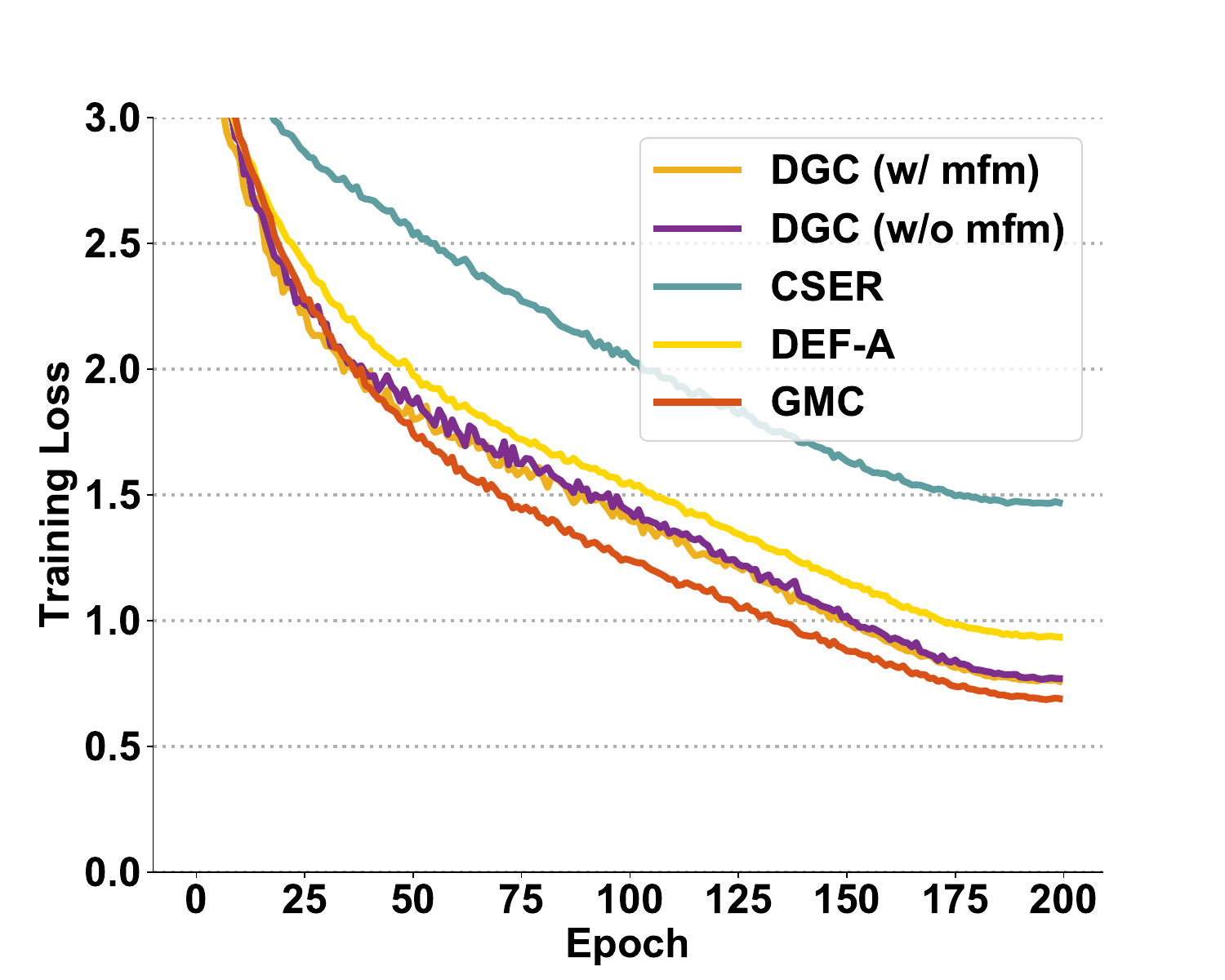}\vspace{-2pt}
      \includegraphics[width=1\linewidth]{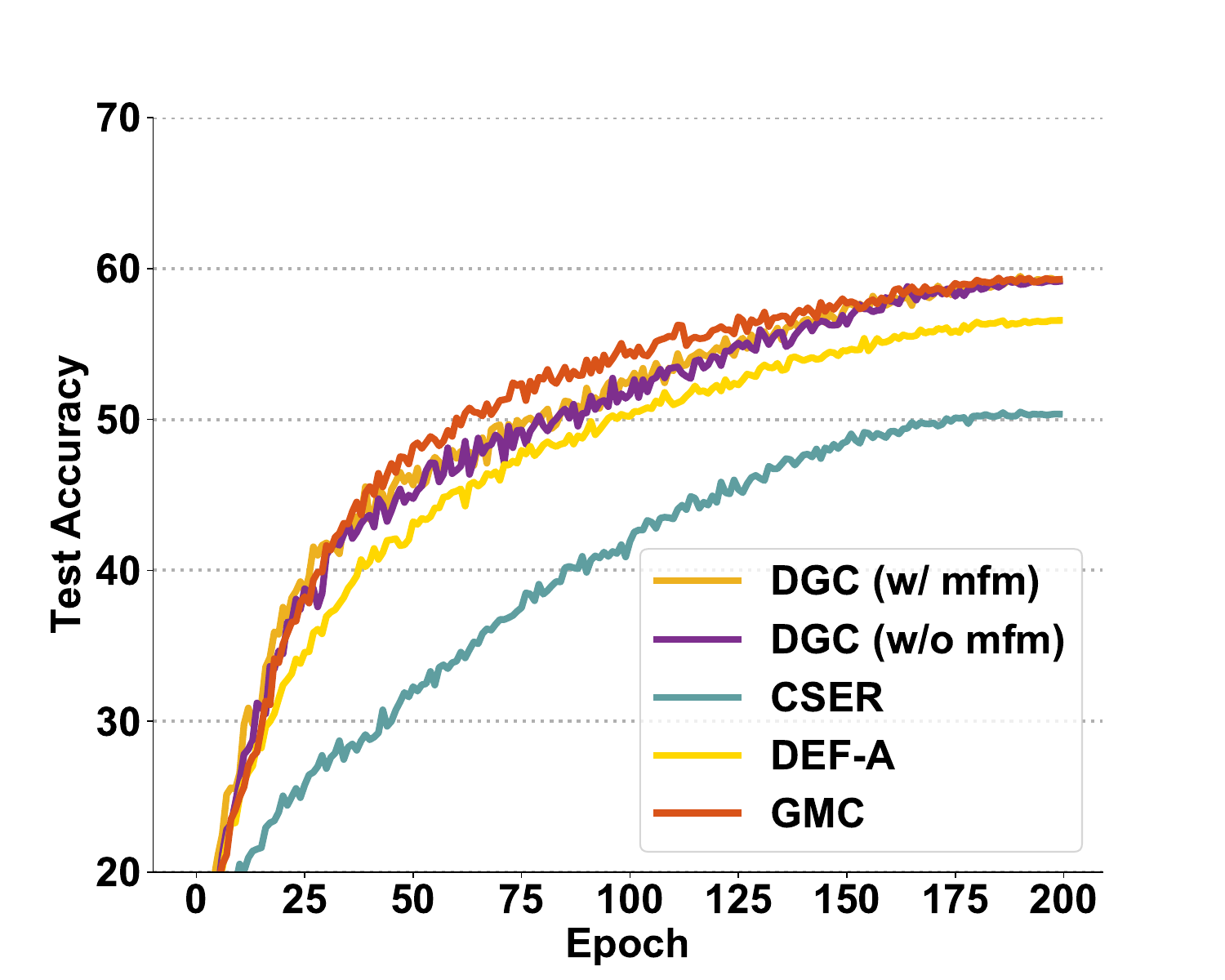}
      \end{minipage}}

\caption{Training curves of different methods under IID data distribution.}\label{fig:IID}
\end{figure*}
\begin{figure*}[!t]
      \centering
      \subfigure[ResNet20, CIFAR10]{
        \label{fig:NONIID1}
        \begin{minipage}[b]{0.23\textwidth}
          \includegraphics[width=1\linewidth]{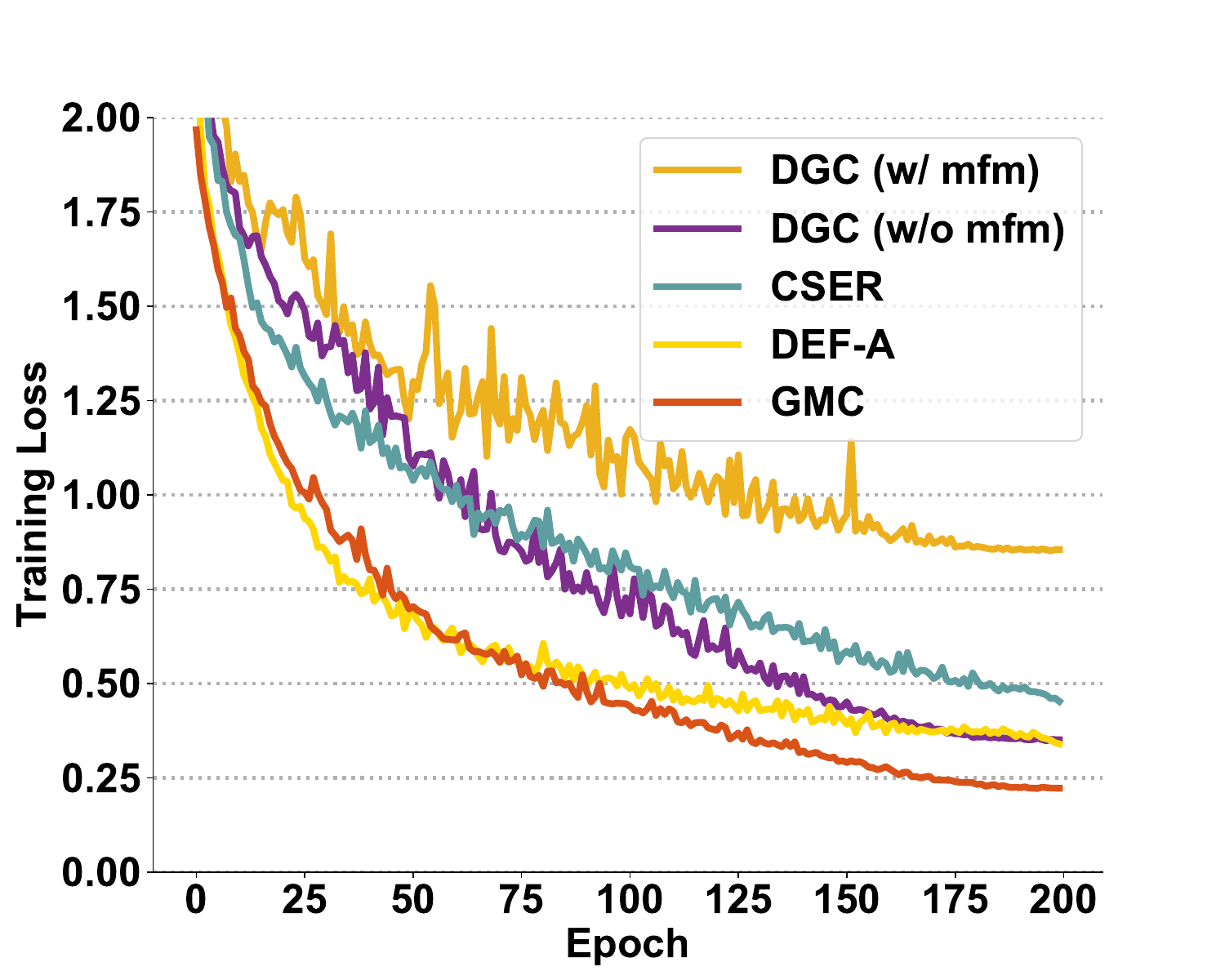}\vspace{-2pt}
          \includegraphics[width=1\linewidth]{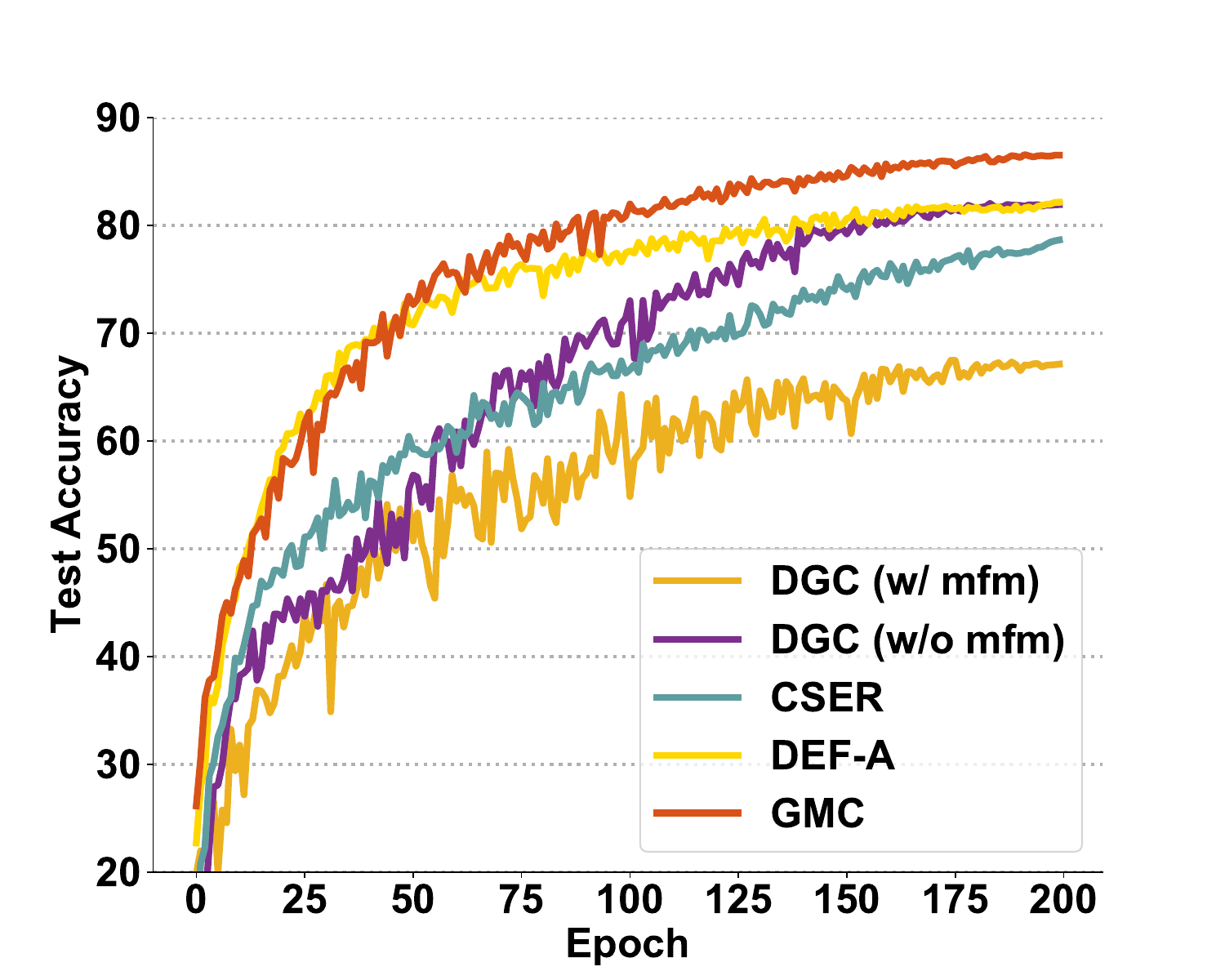}
          \end{minipage}}
      \subfigure[ViT, CIFAR10]{
        \label{fig:NONIID2}
        \begin{minipage}[b]{0.23\textwidth}
          \includegraphics[width=1\linewidth]{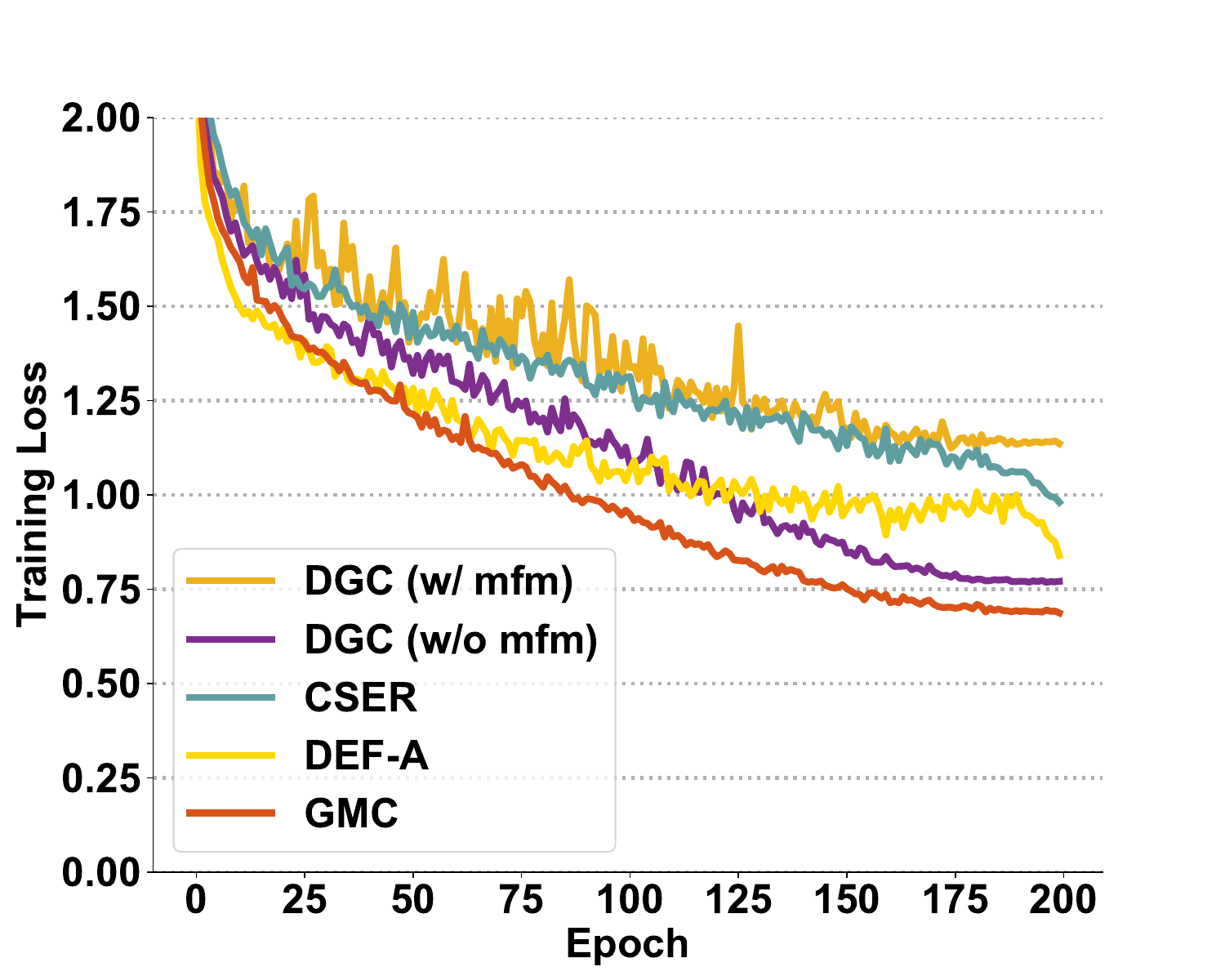}\vspace{-2pt}
          \includegraphics[width=1\linewidth]{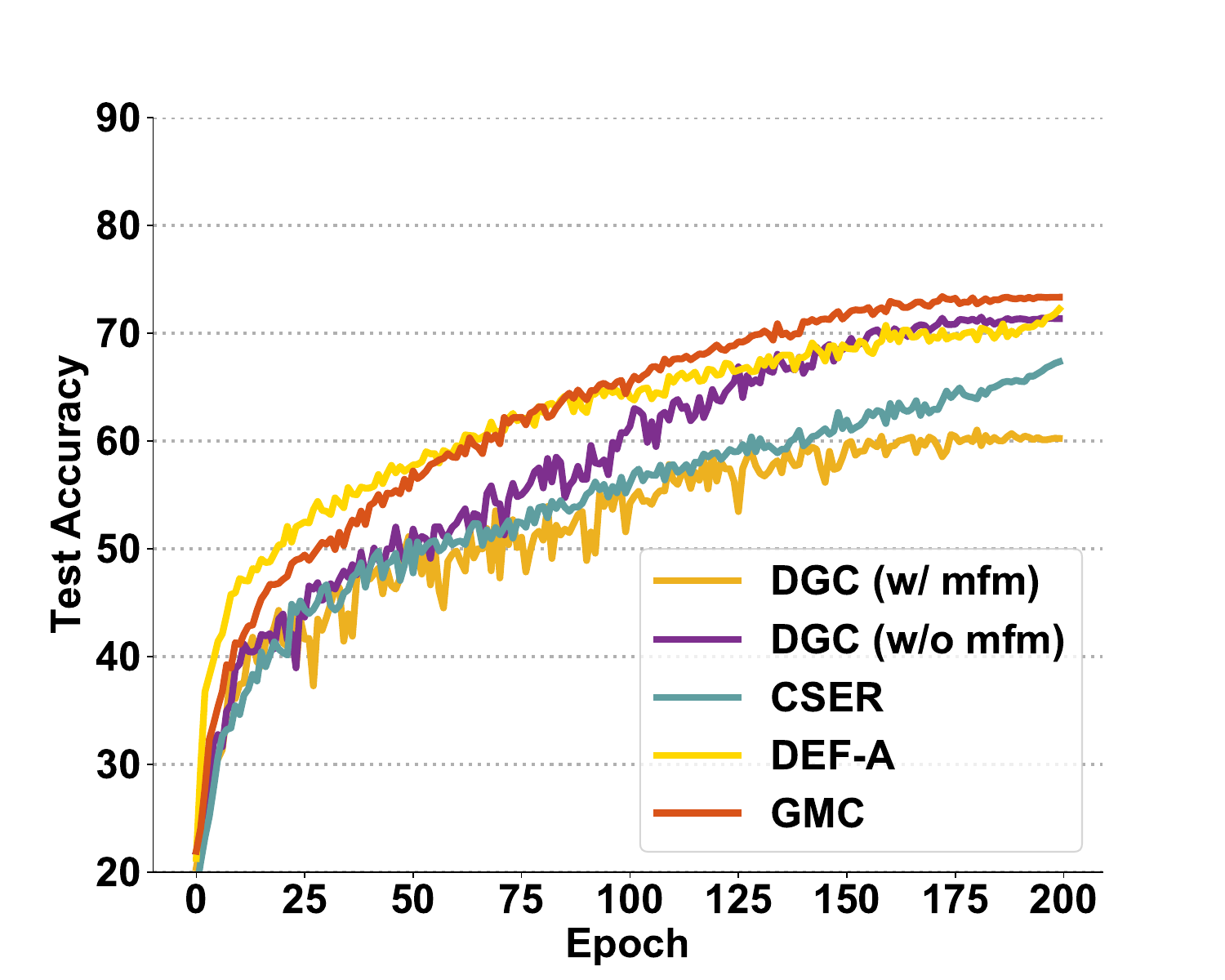}
          \end{minipage}}
      \subfigure[ResNet20, CIFAR100]{
        \label{fig:NONIID3}
        \begin{minipage}[b]{0.23\textwidth}
          \includegraphics[width=1\linewidth]{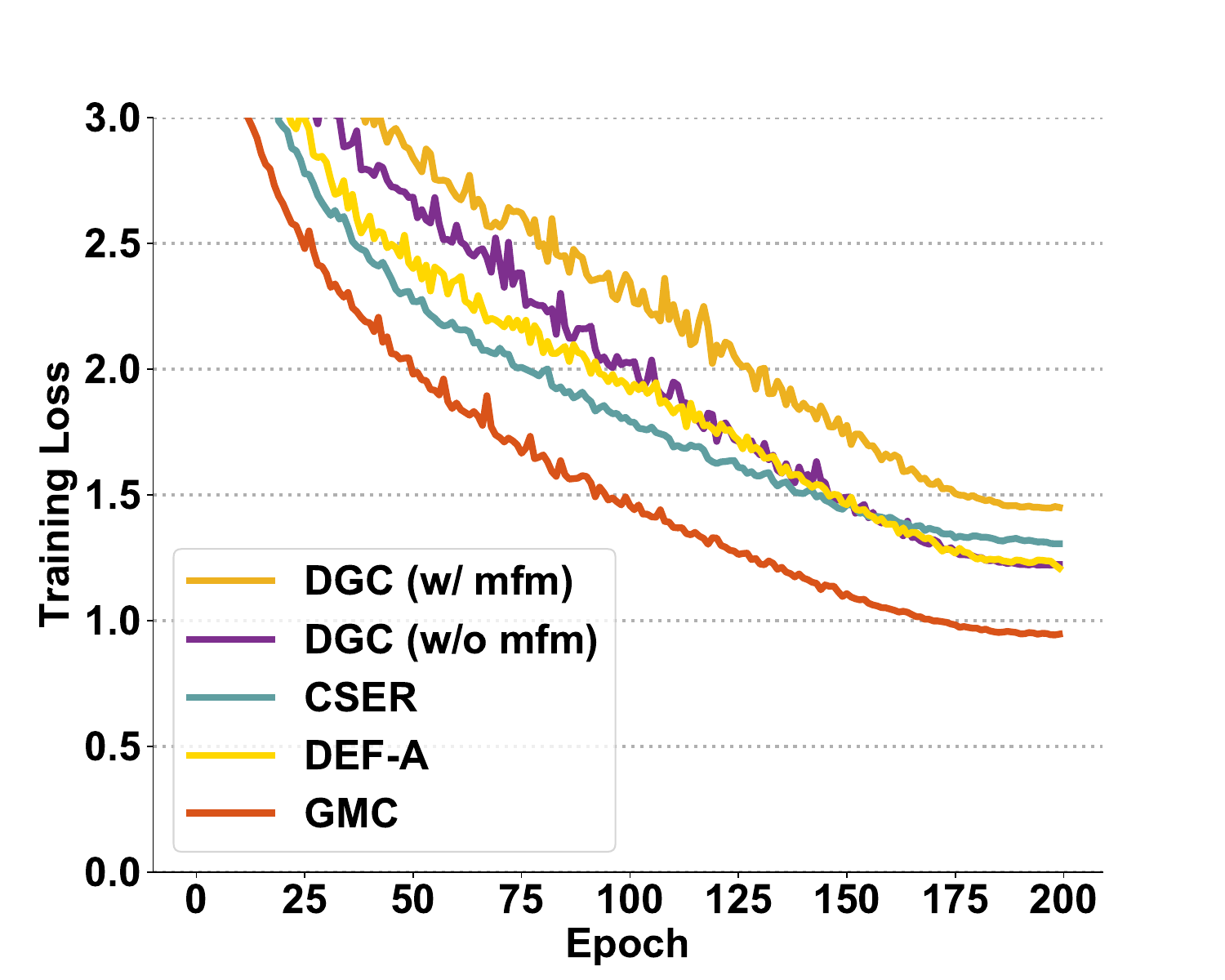}\vspace{-2pt}
          \includegraphics[width=1\linewidth]{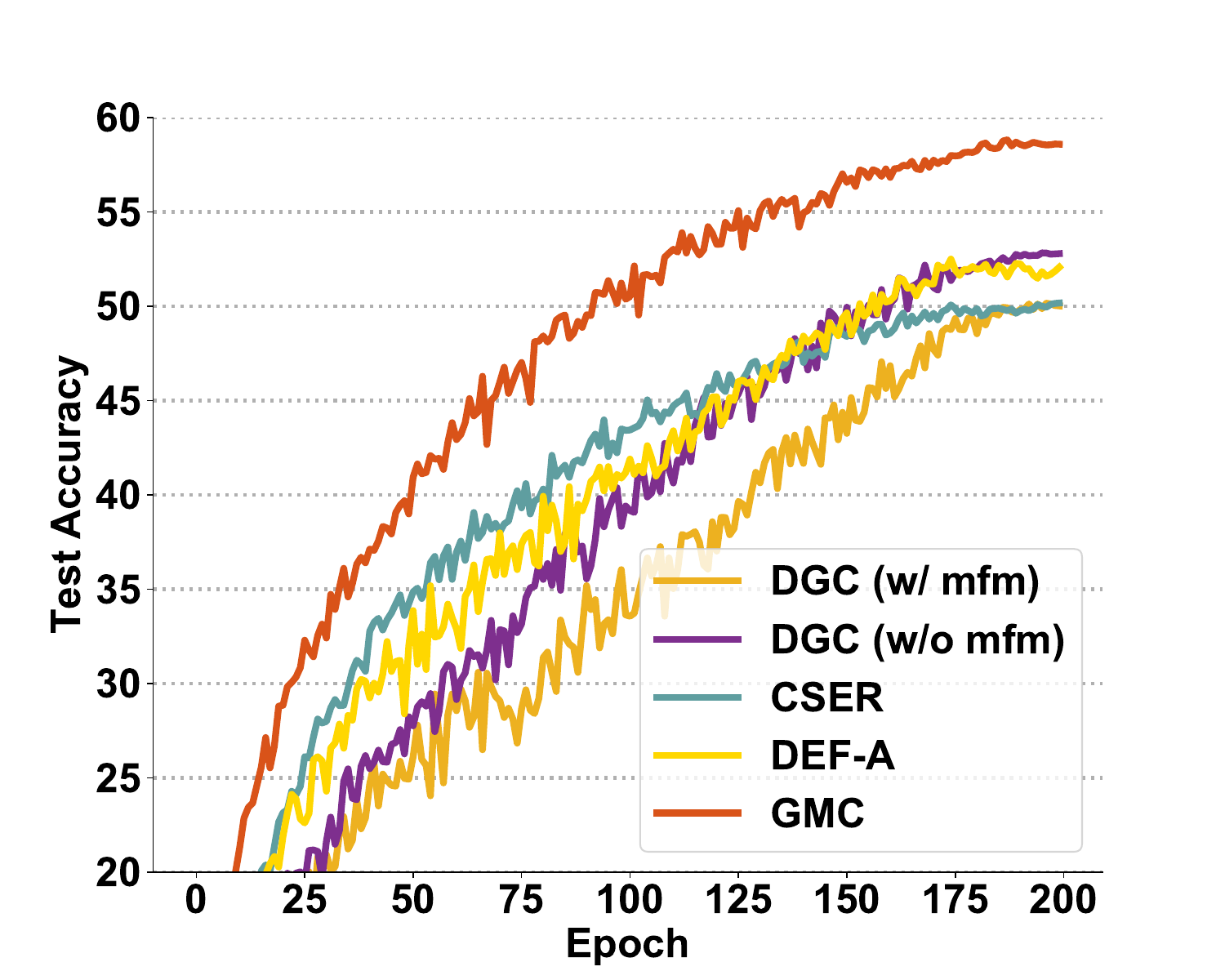}
          \end{minipage}}
      \subfigure[ViT, CIFAR100]{
        \label{fig:NONIID4}
        \begin{minipage}[b]{0.23\textwidth}
          \includegraphics[width=1\linewidth]{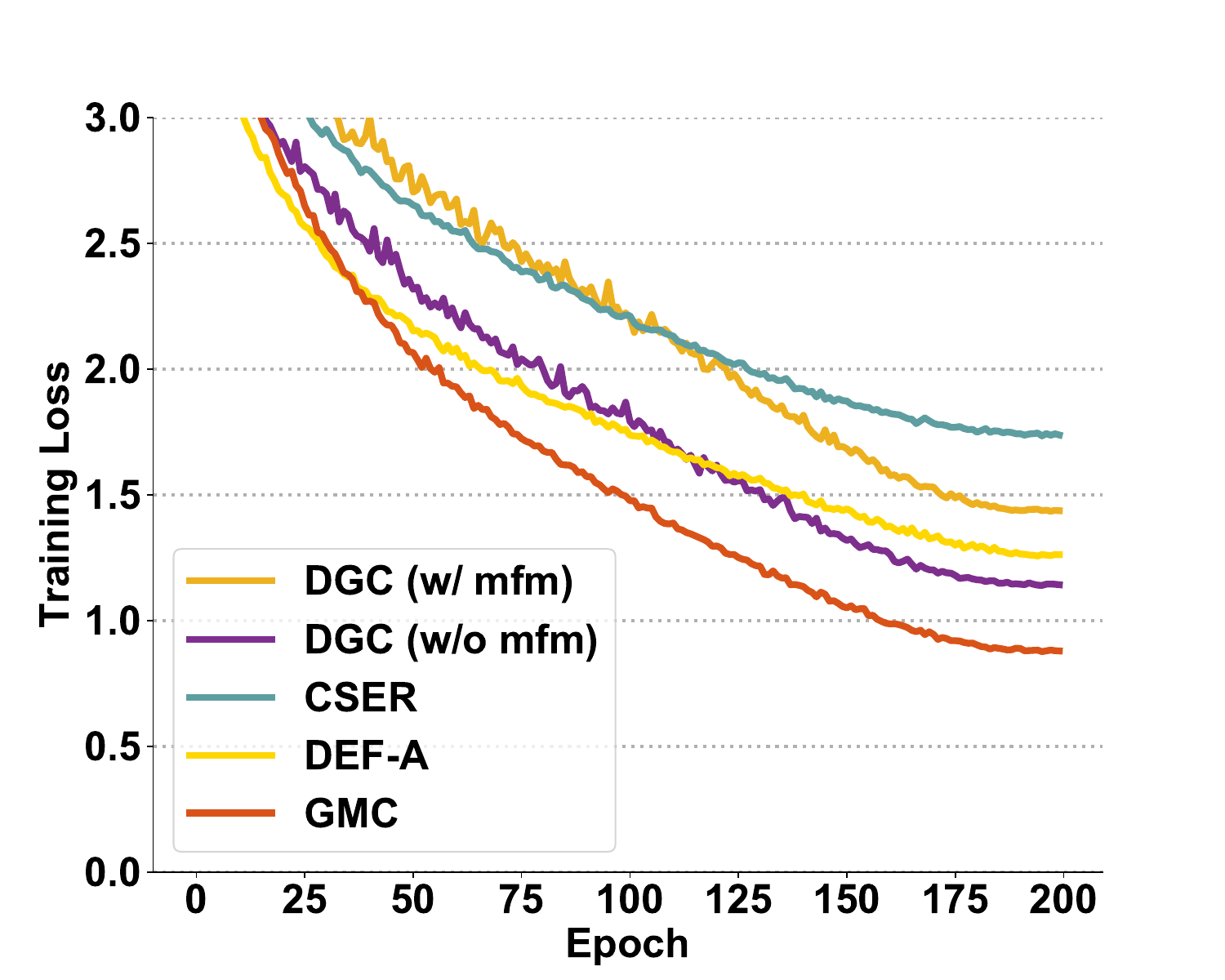}\vspace{-2pt}
          \includegraphics[width=1\linewidth]{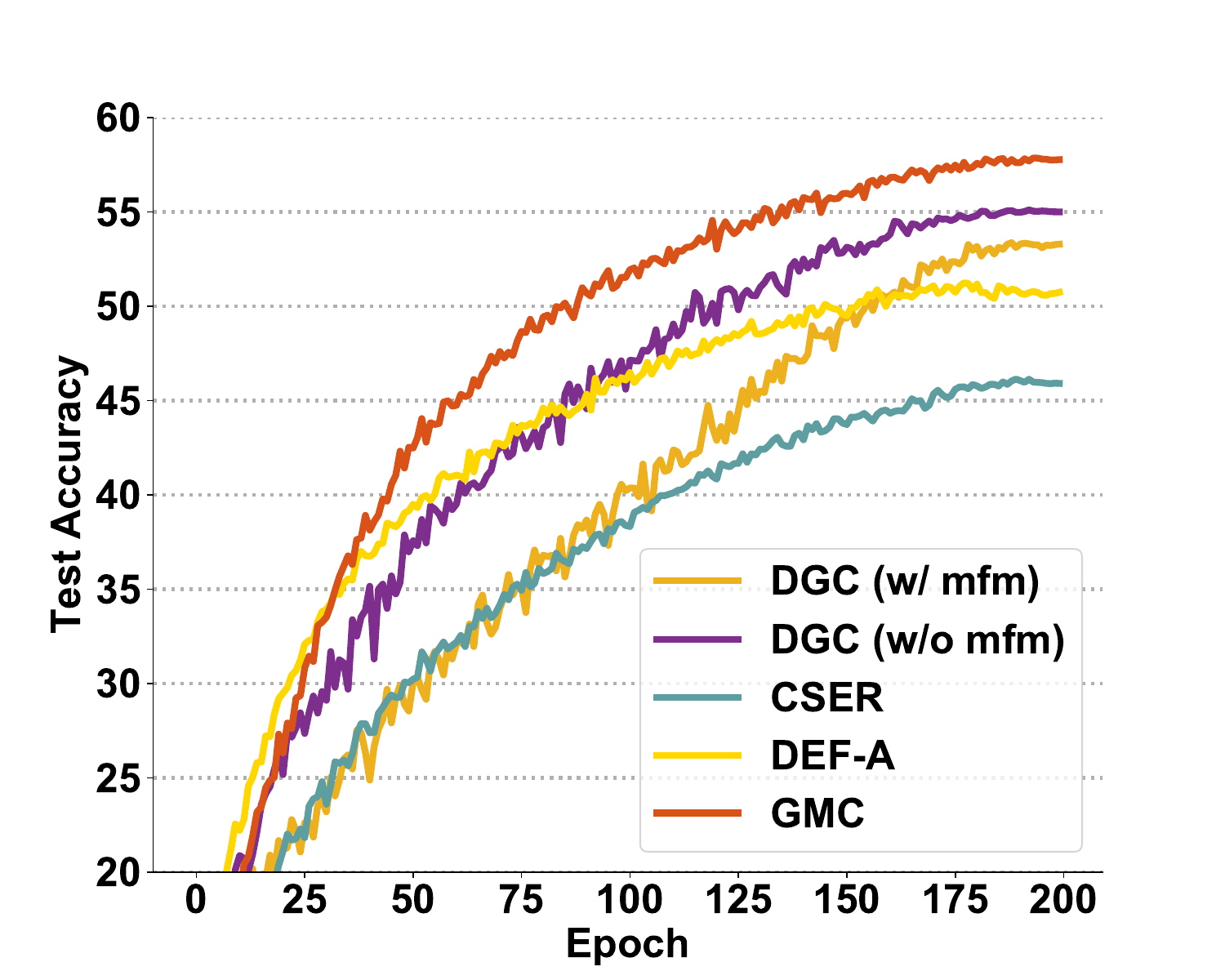}
          \end{minipage}}
    \caption{Training curves of different methods under non-IID data distribution.}\label{fig:NONIID}
    \end{figure*}

   
Table~\ref{tab:GMC-IID} shows the empirical results of different methods under IID data distribution. Figure~\ref{fig:IID} shows the training curves under IID data distribution. We can observe that each method achieves comparable RCC. As for test accuracy, GMC and DGC~(w/ mfm) exhibit comparable performance and outperform the other three methods.
Table~\ref{tab:GMC-NONIID} and Figure~\ref{fig:NONIID} show the performance under non-IID data distribution. We can find that GMC can  achieve much better test accuracy and faster convergence speed compared to other methods. Furthermore, we can find that the momentum factor masking trick will severely impair the performance of DGC under non-IID data distribution.

\begin{table*}[!t]\small
    \centering
    \caption{Empirical results of different methods under non-IID data distribution.}\label{tab:GMC-NONIID}
    \setlength{\tabcolsep}{1.6mm}{    
      \begin{tabular}{c|c|ccccccc}
      \hline    
      \multirow{2}*{Dataset} &\multirow{2}*{Model} & \multirow{2}*{Method} & DGC & DGC & \multirow{2}*{CSER}& \multirow{2}*{DEF-A} & \multirow{2}*{GMC}   \\
      & &  &  (w/ mfm) & (w/o mfm) & ~ & ~ & ~   \\ \hline
      \multirow{4}*{CIFAR10}& \multirow{2}*{ResNet20~(GN)} & Accuracy & 67.13\% & 81.93\%   & 78.65\% & 82.15\%  & \textbf{86.51\%}  \\
      ~& ~     & RCC & 0.65\% & 0.65\%   & 0.63\% & 0.65\%& 0.64\% \\\cline{2-8}
      ~& \multirow{2}*{ViT} & Accuracy & 60.24\% & 71.35\%   & 67.35\%& 72.25\% & \textbf{73.34\%}  \\
      ~& ~     & RCC & 0.81\% & 0.81\%   & 0.67\% & 0.81\% & 0.81\% \\ \hline
      \multirow{4}*{CIFAR100} & \multirow{2}*{ResNet20~(GN)}  & Accuracy & 50.02\% & 52.79\%   & 50.17\% & 52.07\% & \textbf{58.59\%}  \\
      ~& ~     & RCC & 0.64\% & 0.64\%   & 0.65\% & 0.64\%& 0.63\% \\    \cline{2-8}  
      ~& \multirow{2}*{ViT}   & Accuracy & 53.29\% & 55.00\%   & 45.91\%& 50.74\% & \textbf{57.77\%}  \\
      ~& ~     & RCC & 0.79\% & 0.80\%   & 0.70\% & 0.79\%& 0.79\% \\ \hline
    \end{tabular}}
    \end{table*}

\subsection{Results of GMC+}

To further verify the superiority of global momentum, we also evaluate DEF-A and GMC$+$ when using the RBGS compressor. In RBGS, we randomly select a block that contains $s$ components using the same random seed among the workers, where $\frac{s}{d}=\frac{1}{1024}$. Since the components being communicated among the workers have the same indices, there is no need to send the indices during communication. 
Hence, the communication cost of receiving sparsified vectors from the workers for RBGS will be half of that for top-$s$: $\sum_{t \in [T]}(\sum_{k \in [K]}\|\mathcal{C}(\e_{t+\frac{1}{2},k})\|_0$, where $\|\mathcal{C}(\e_{t+\frac{1}{2},k})\|_0/d = \frac{1}{1024}$. Similarly, the communication cost of sending aggregated sparsified vectors to workers will be $K\sum_{t \in [T]}\|\w_{t+1} - \w_t\|_0$. Since the sparsified vectors $\mathcal{C}(\e_{t+\frac{1}{2},k})$ have the same indices across all the workers, the aggregated sparsified vector satisfies $\|\w_{t+1} - \w_t\|_0 /d =\frac{1}{1024}$.
The RCC of both DEF-A and GMC$+$ with the above settings for the compressor can be directly calculated as follows: $$\mbox{RCC}=\frac{1}{2dKT}\sum_{t \in [T]}(K\|\w_{t+1} - \w_t\|_0 + \sum_{k \in [K]}\|\mathcal{C}(\e_{t+\frac{1}{2},k})\|_0)=\frac{1}{1024}\approx 0.098\%.$$

The empirical results of DEF-A and GMC+ with various settings of $\lambda$ in ResNet20/CIFAR100 training are shown in Table~\ref{tab:GMC-lambda}. Both DEF-A and GMC+ almost fail to converge when $\lambda = 0$. 
DEF-A achieves its best performance when $\lambda = 0.3$. In comparison,  GMC+ outperforms DEF-A across different $\lambda$ values and shows a preference for a larger $\lambda$ (e.g., 0.5).
In the following experiments, we set $\lambda$ as 0.3 for DEF-A and 0.5 for GMC+. $\lambda = 0.3$ is also recommended for DEF-A in~\citep{DBLP:conf/icml/XuH22}.

\begin{table*}[!t]\small
  \centering
  \caption{Empirical results with various settings of $\lambda$ in ResNet20/CIFAR100 training under IID data distribution.}\label{tab:GMC-lambda}
  \setlength{\tabcolsep}{2mm}{  
  \begin{tabular}{c|cccccc}
    \hline    
    ~                     &  $\lambda$      & 0.0 &  0.3 & 0.5 & 0.8 & 1.0    \\ \hline
    \multirow{2}*{DEF-A}  & Training Loss   &  4.09  &\textbf{0.56}  &0.57 &  0.68  & 0.76       \\
    ~                     & Test Accuracy   &  6.03\%   &\textbf{66.36\%} &65.28\% & 64.84\%    & 64.45\%       \\ \hline
    \multirow{2}*{GMC+}   & Training Loss   &  3.48   &0.55  &\textbf{0.40} &  0.42   & 0.43         \\ 
    ~                     & Test Accuracy   &  15.81\%  &67.31\% &\textbf{67.94\%}& 67.27\%    & 66.68\%         \\\hline
  \end{tabular}}
  \end{table*}

\begin{table*}[!t]\small
  \centering
  \caption{Test accuracy of DEF-A and GMC+.}\label{tab:DEF-A}
  \setlength{\tabcolsep}{2mm}{  
      \begin{tabular}{c|c|c|ccc}
        \hline    
      Dataset            &Data Distribution          &Model            & DEF-A     & GMC$+$             \\\hline
    \multirow{4}*{CIFAR10}&\multirow{2}*{IID}      & ResNet20~(BN)   & 90.78\%   & \textbf{91.41\%}   \\ \cline{3-5}
    ~                     &~                       & ViT             & 74.21\%   & \textbf{75.78\%}   \\ \cline{2-5}
    ~                     &\multirow{2}*{non-IID}  & ResNet20~(GN)   & 78.17\%   & \textbf{79.25\%}   \\ \cline{3-5}
    ~                     &~                       & ViT             & 65.16\%   & \textbf{66.61\%}   \\ \hline
    \multirow{4}*{CIFAR100}&\multirow{2}*{IID}     & ResNet20~(BN)   & 66.36\%   & \textbf{67.94\%}   \\ \cline{3-5}
    ~                     &~                       & ViT             & 50.85\%   & \textbf{55.50\%}   \\ \cline{2-5}
    ~                     &\multirow{2}*{non-IID}  & ResNet20~(GN)   & 49.39\%   & \textbf{51.22\%}   \\ \cline{3-5}
    ~                     &~                       & ViT             & 46.09\%   & \textbf{50.07\%}   \\ \hline
      \end{tabular}}
  \end{table*}
  \begin{figure*}[!t]
    \centering
    \subfigure[ResNet20, CIFAR10]{
      \begin{minipage}[b]{0.23\textwidth}
        \includegraphics[width=1\linewidth]{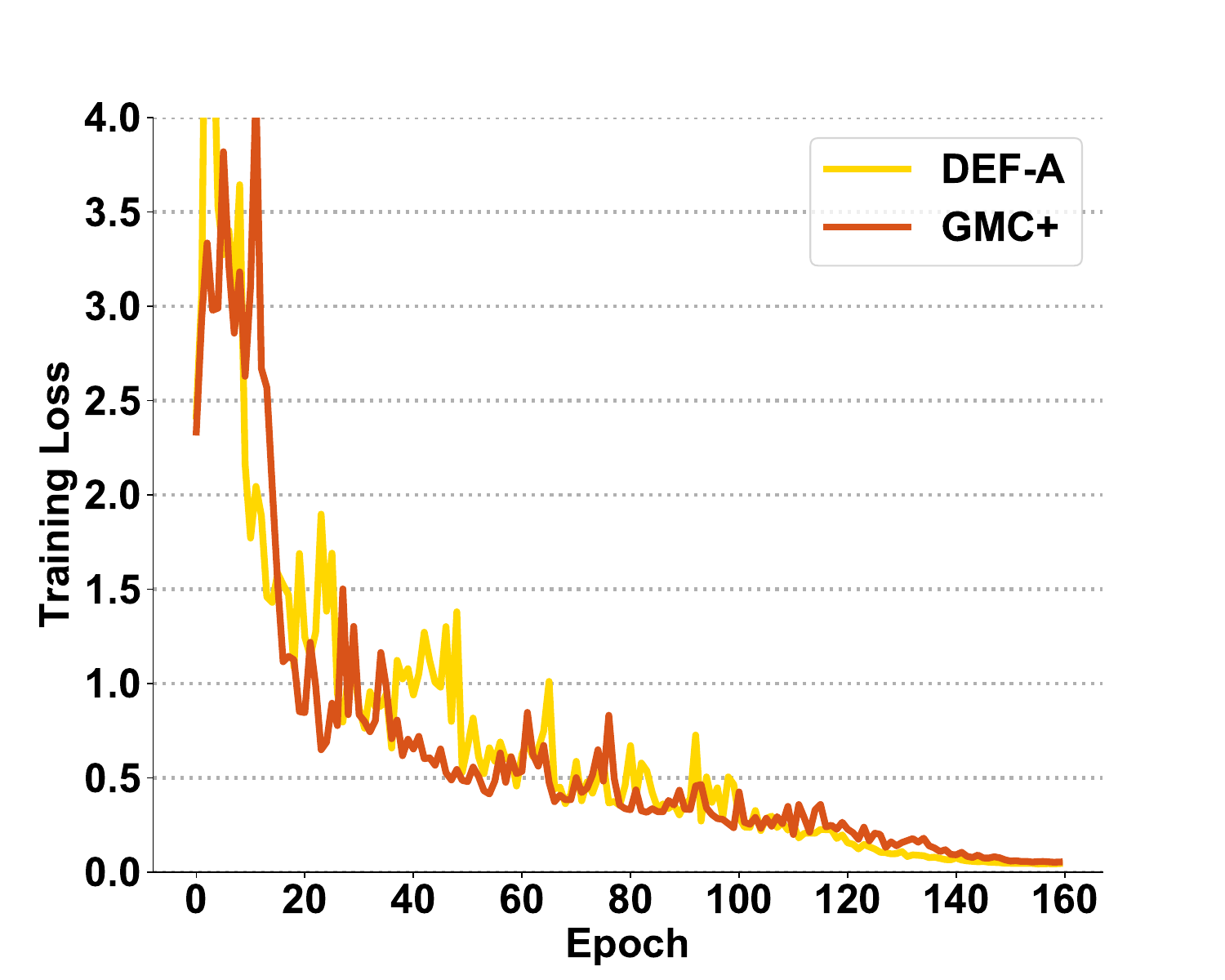}\vspace{-2pt}
        \includegraphics[width=1\linewidth]{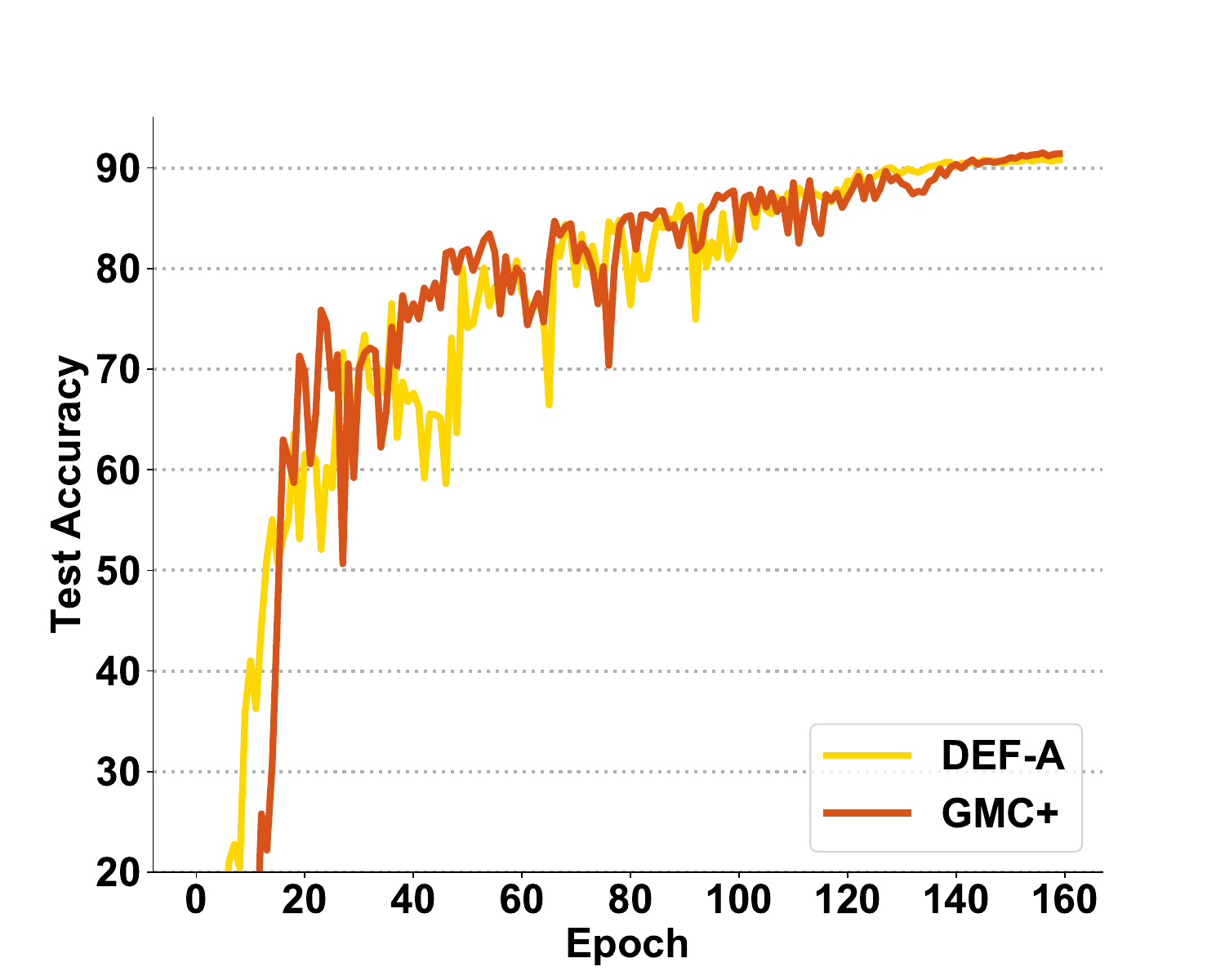}
        \end{minipage}}
    \subfigure[ViT, CIFAR10]{
      \begin{minipage}[b]{0.23\textwidth}
        \includegraphics[width=1\linewidth]{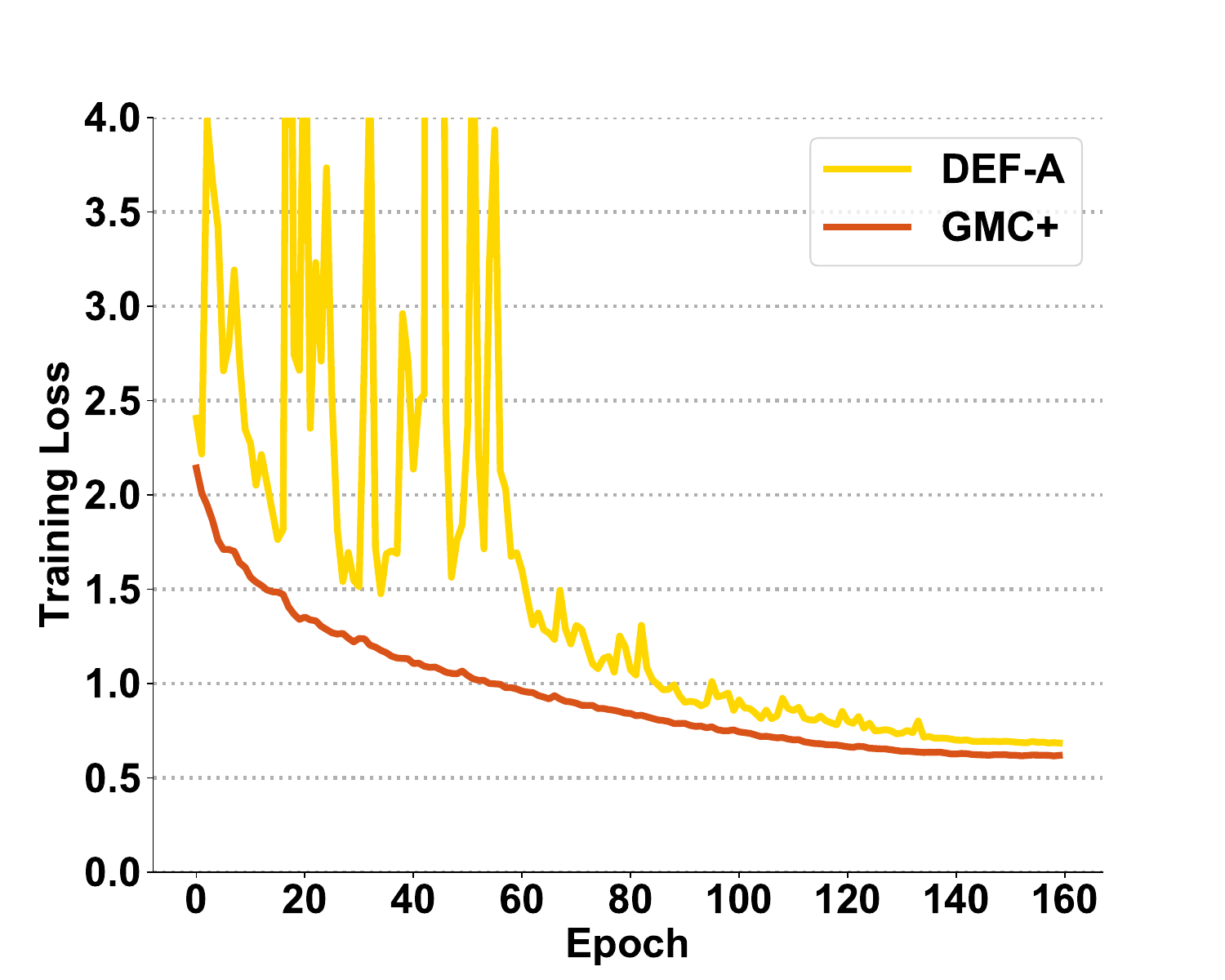}\vspace{-2pt}
        \includegraphics[width=1\linewidth]{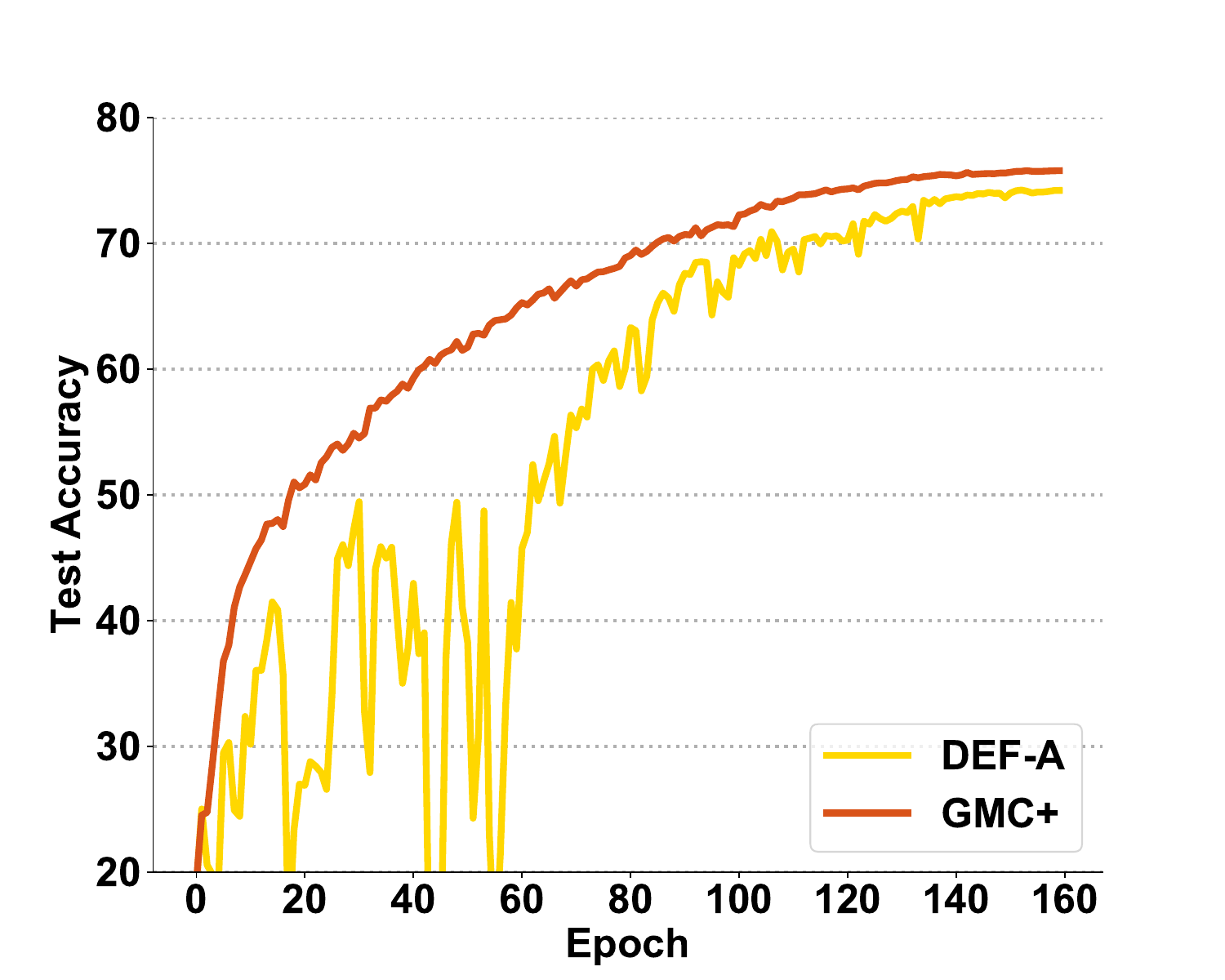}
        \end{minipage}}
    \subfigure[ResNet20, CIFAR100]{
      \begin{minipage}[b]{0.23\textwidth}
        \includegraphics[width=1\linewidth]{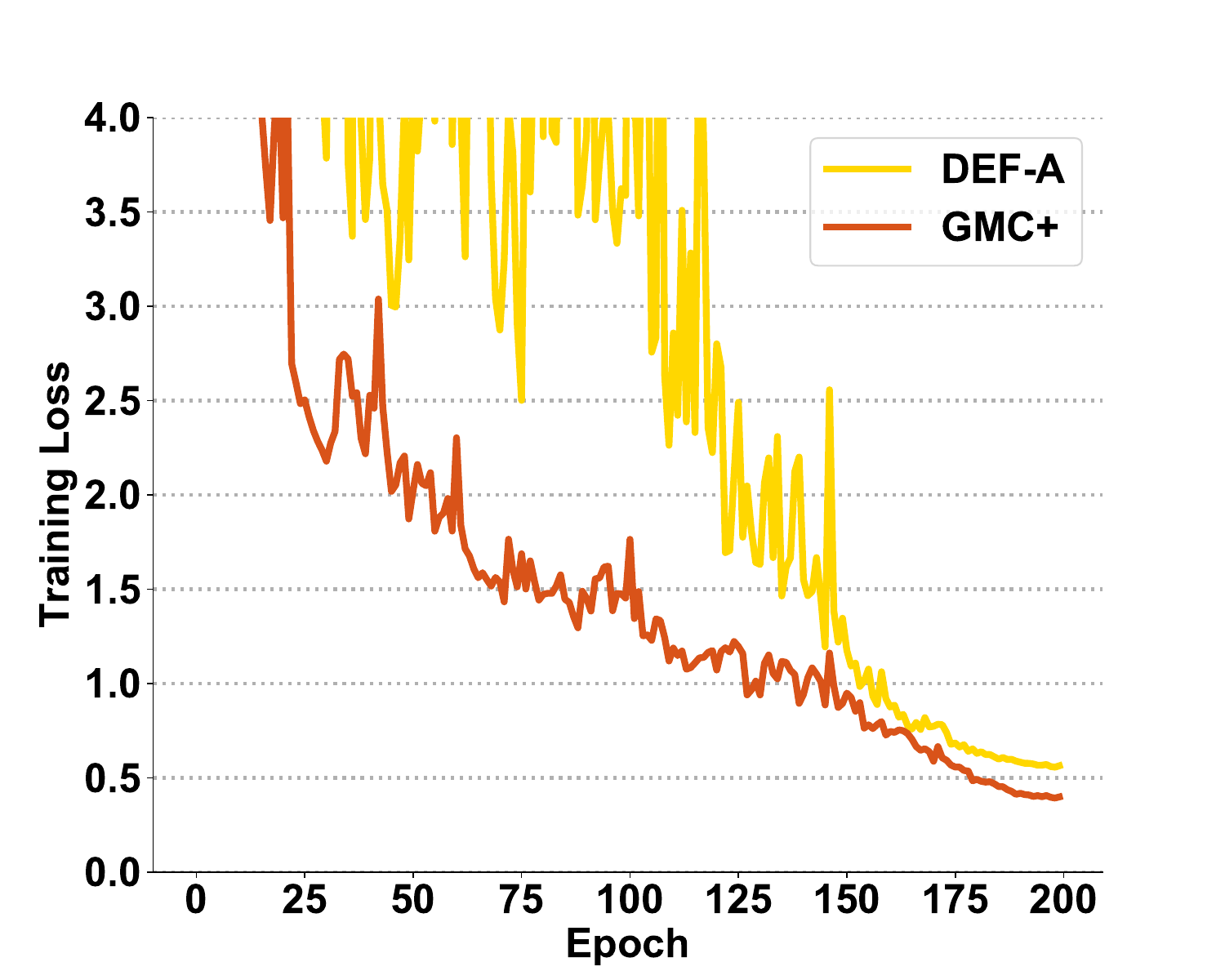}\vspace{-2pt}
        \includegraphics[width=1\linewidth]{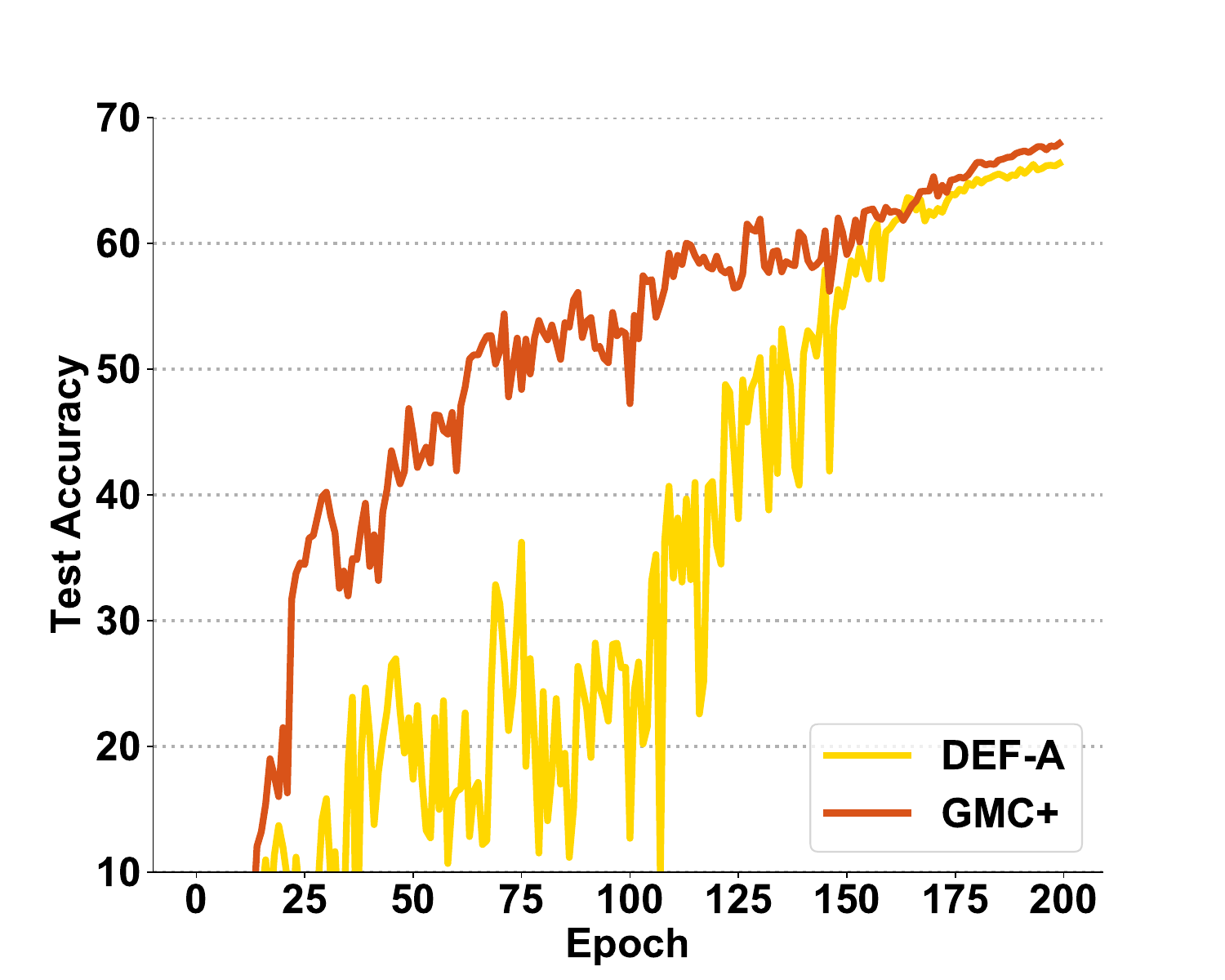}
        \end{minipage}}
    \subfigure[ViT, CIFAR100]{
      \begin{minipage}[b]{0.23\textwidth}
        \includegraphics[width=1\linewidth]{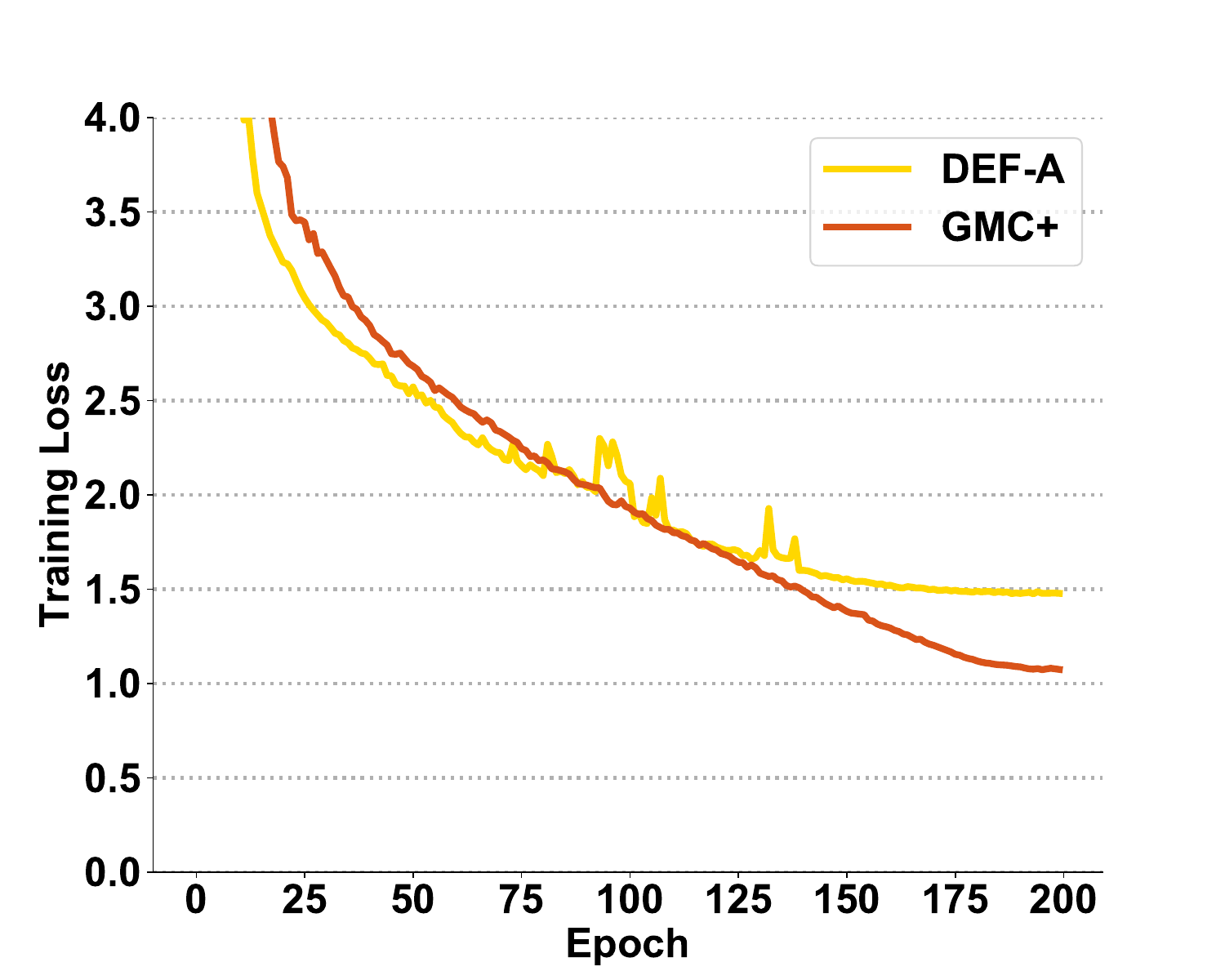}\vspace{-2pt}
        \includegraphics[width=1\linewidth]{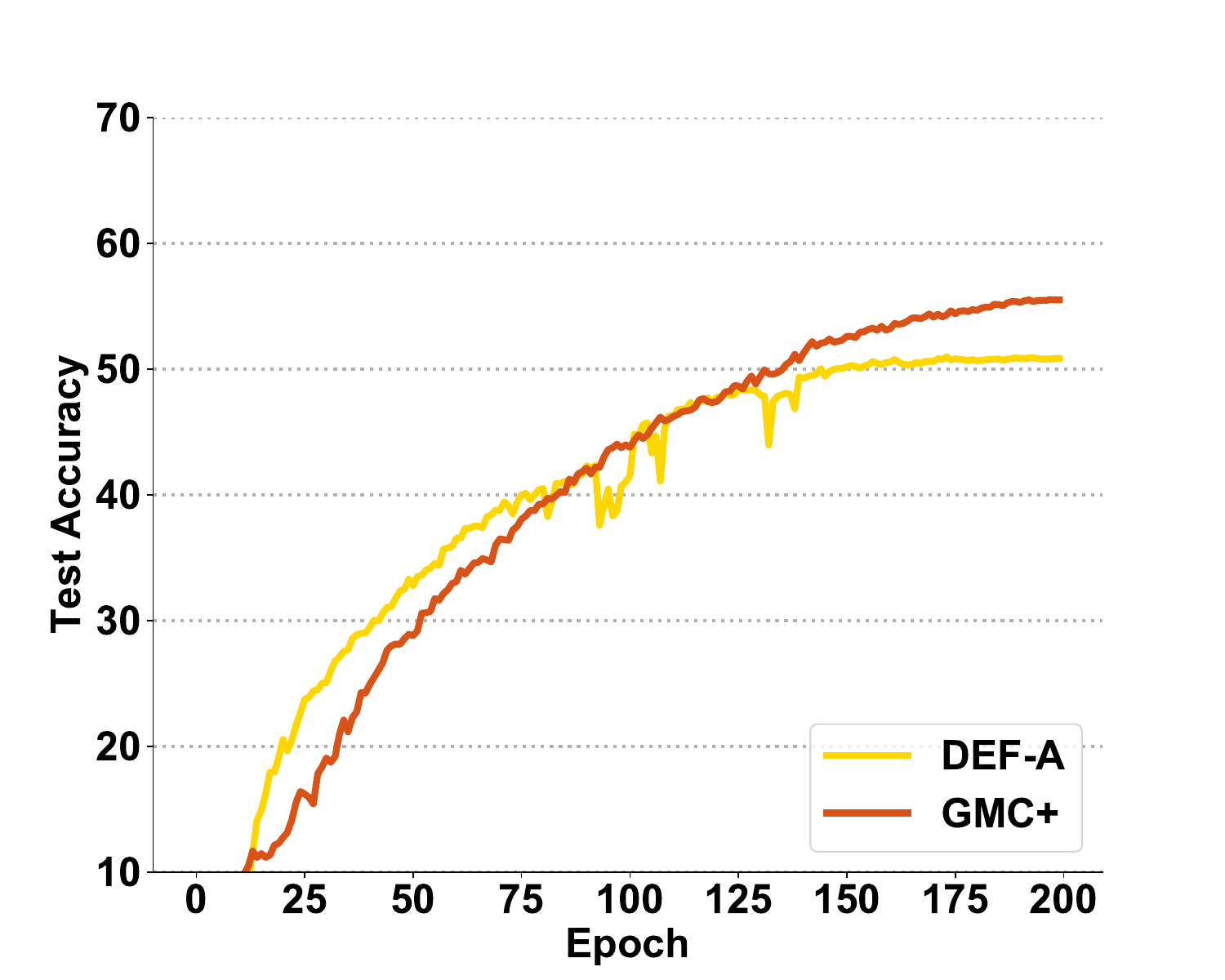}
        \end{minipage}}
  \caption{Training curves of DEF-A and GMC$+$ under IID data distribution.}\label{fig:DEF-IID}
  \end{figure*}   
  
\begin{figure*}[!t]
    \centering
    \subfigure[ResNet20, CIFAR10]{
      \begin{minipage}[b]{0.23\textwidth}
        \includegraphics[width=1\linewidth]{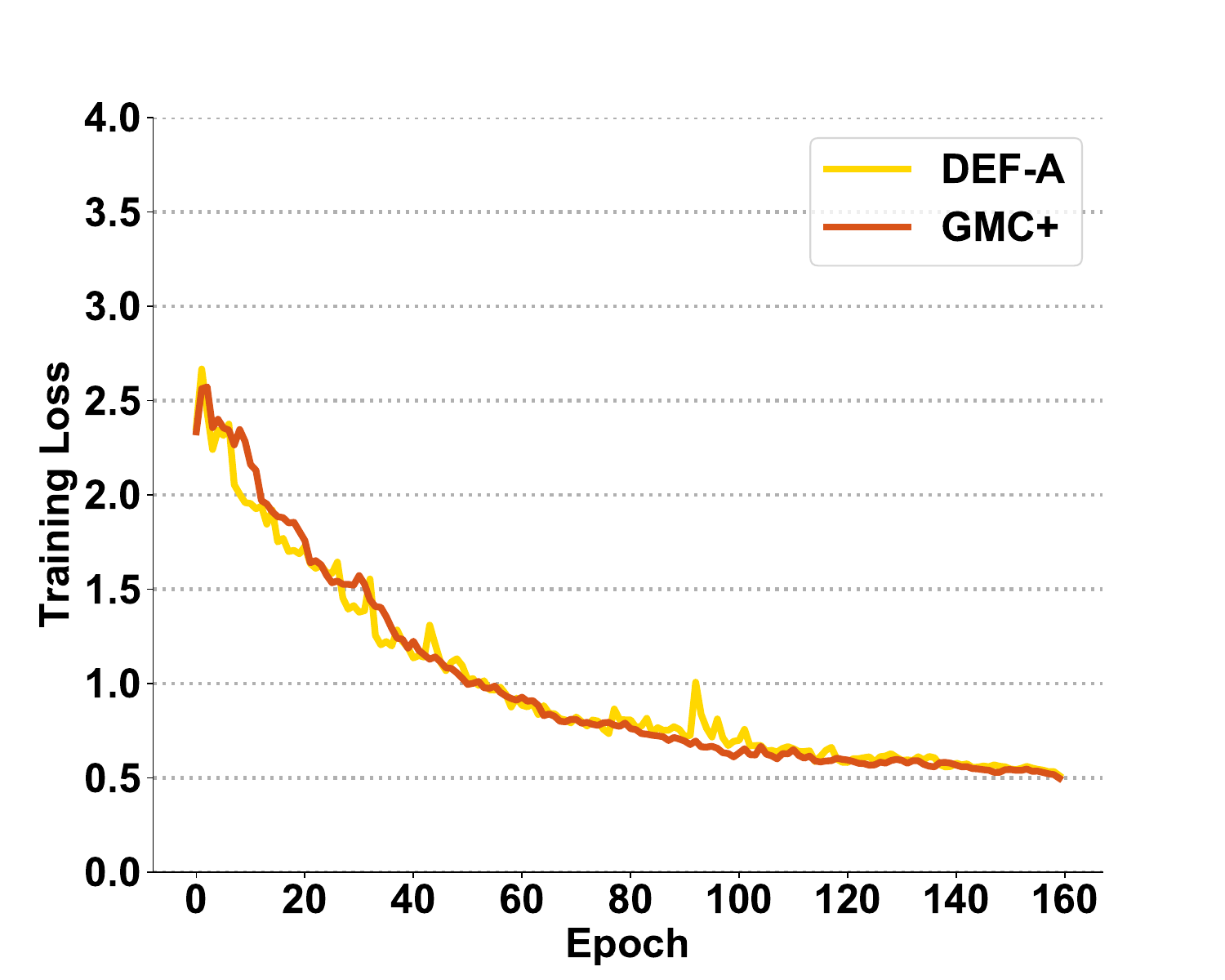}\vspace{-2pt}
        \includegraphics[width=1\linewidth]{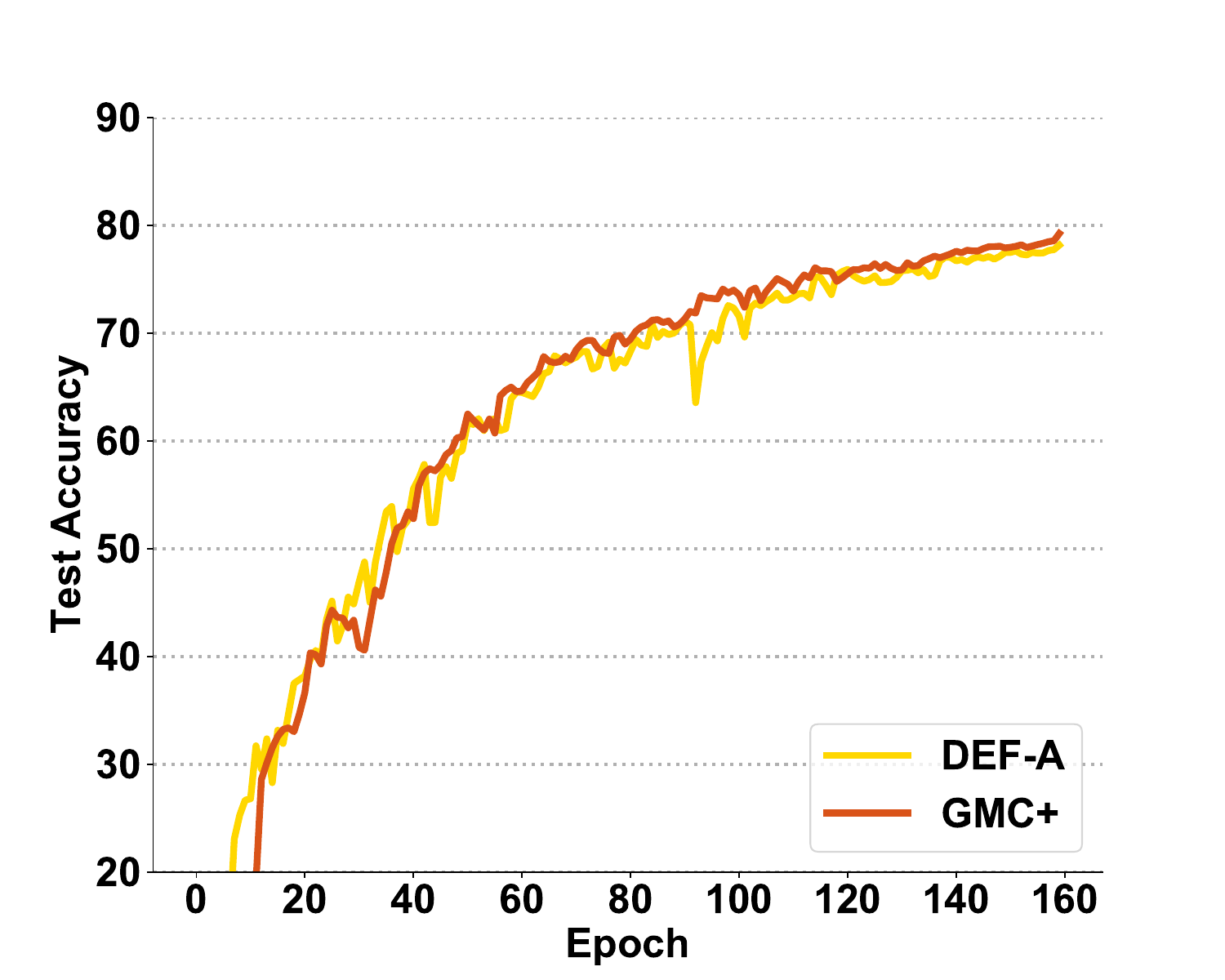}
        \end{minipage}}
    \subfigure[ViT, CIFAR10]{
      \begin{minipage}[b]{0.23\textwidth}
        \includegraphics[width=1\linewidth]{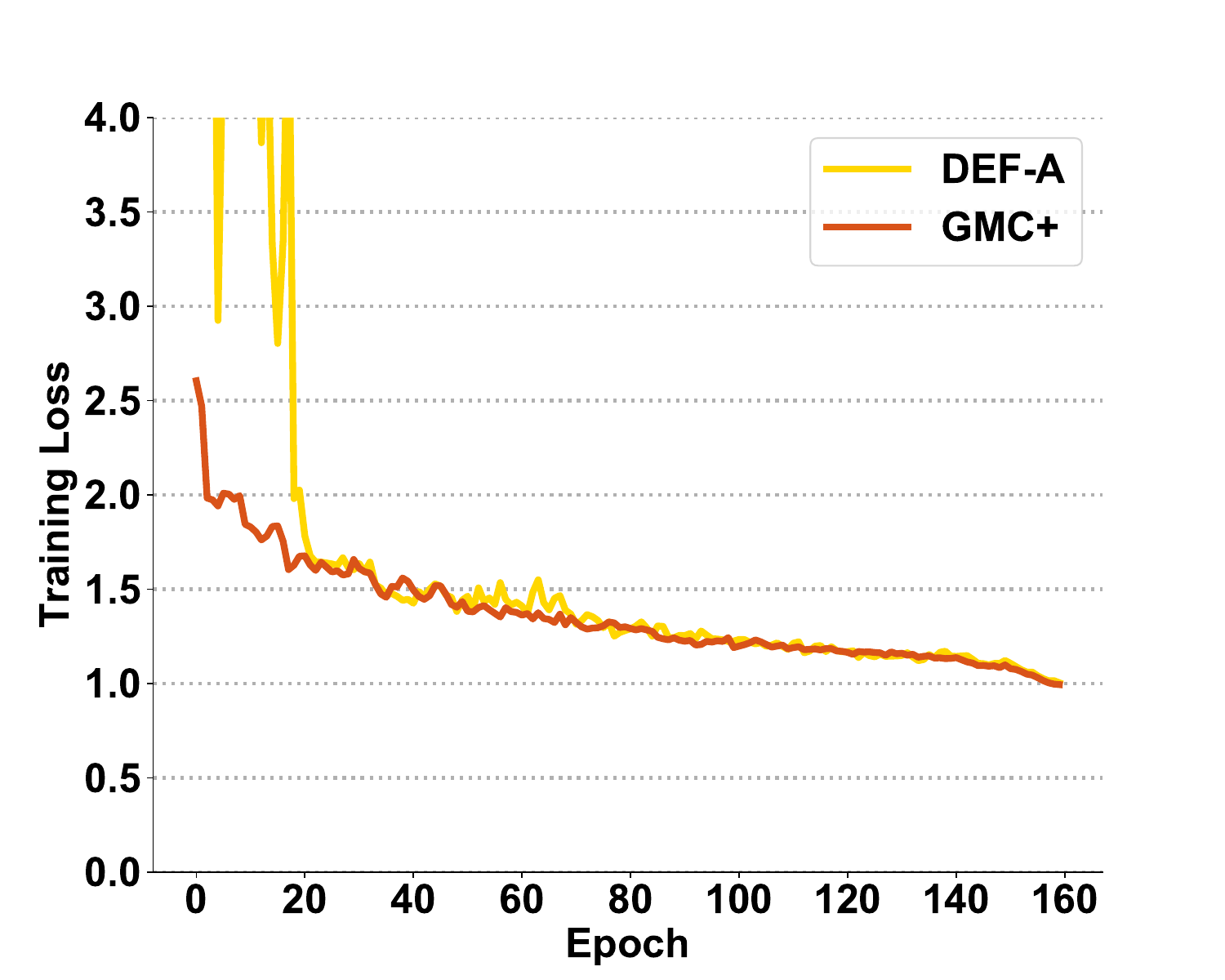}\vspace{-2pt}
        \includegraphics[width=1\linewidth]{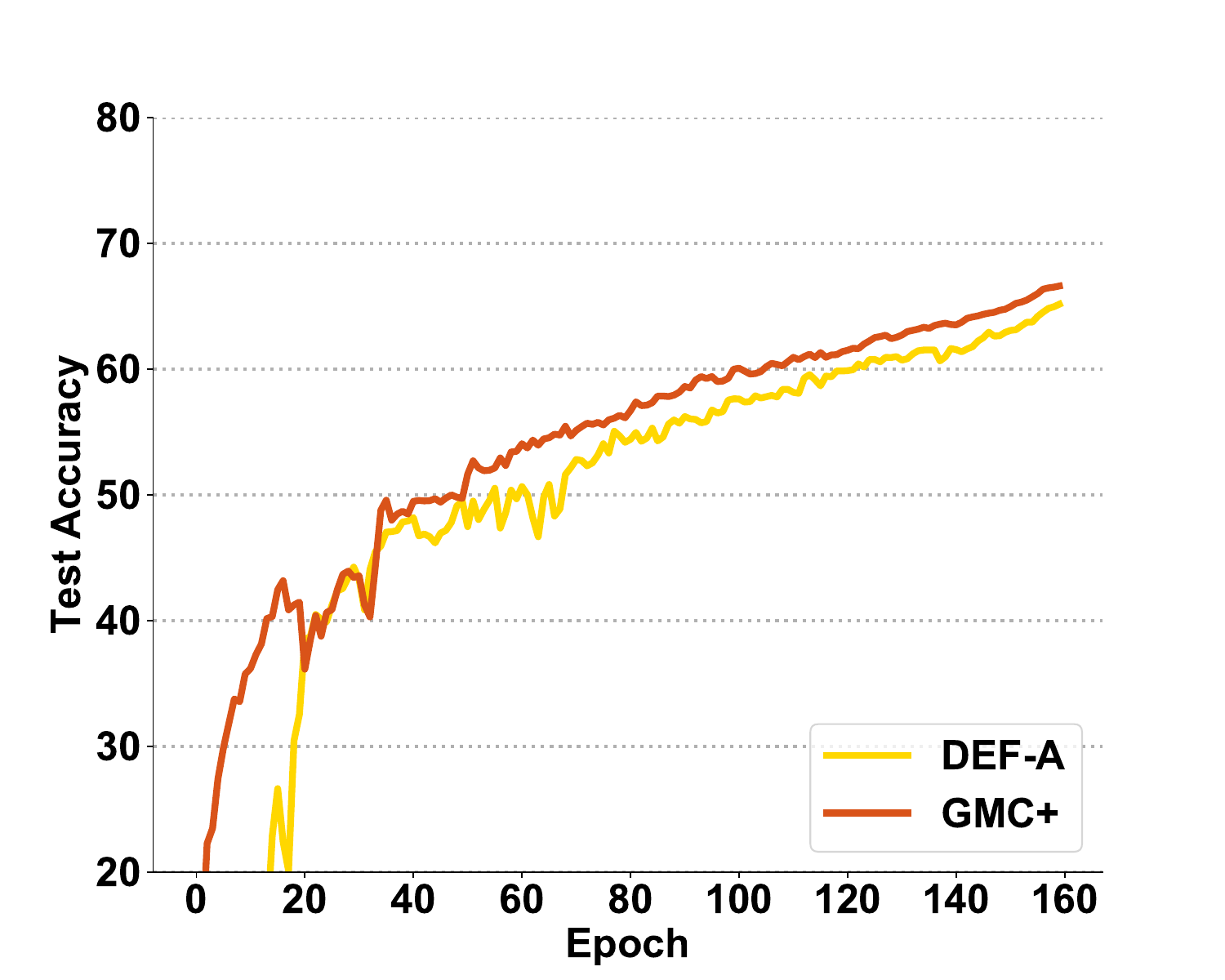}
        \end{minipage}}
    \subfigure[ResNet20, CIFAR100]{
      \begin{minipage}[b]{0.23\textwidth}
        \includegraphics[width=1\linewidth]{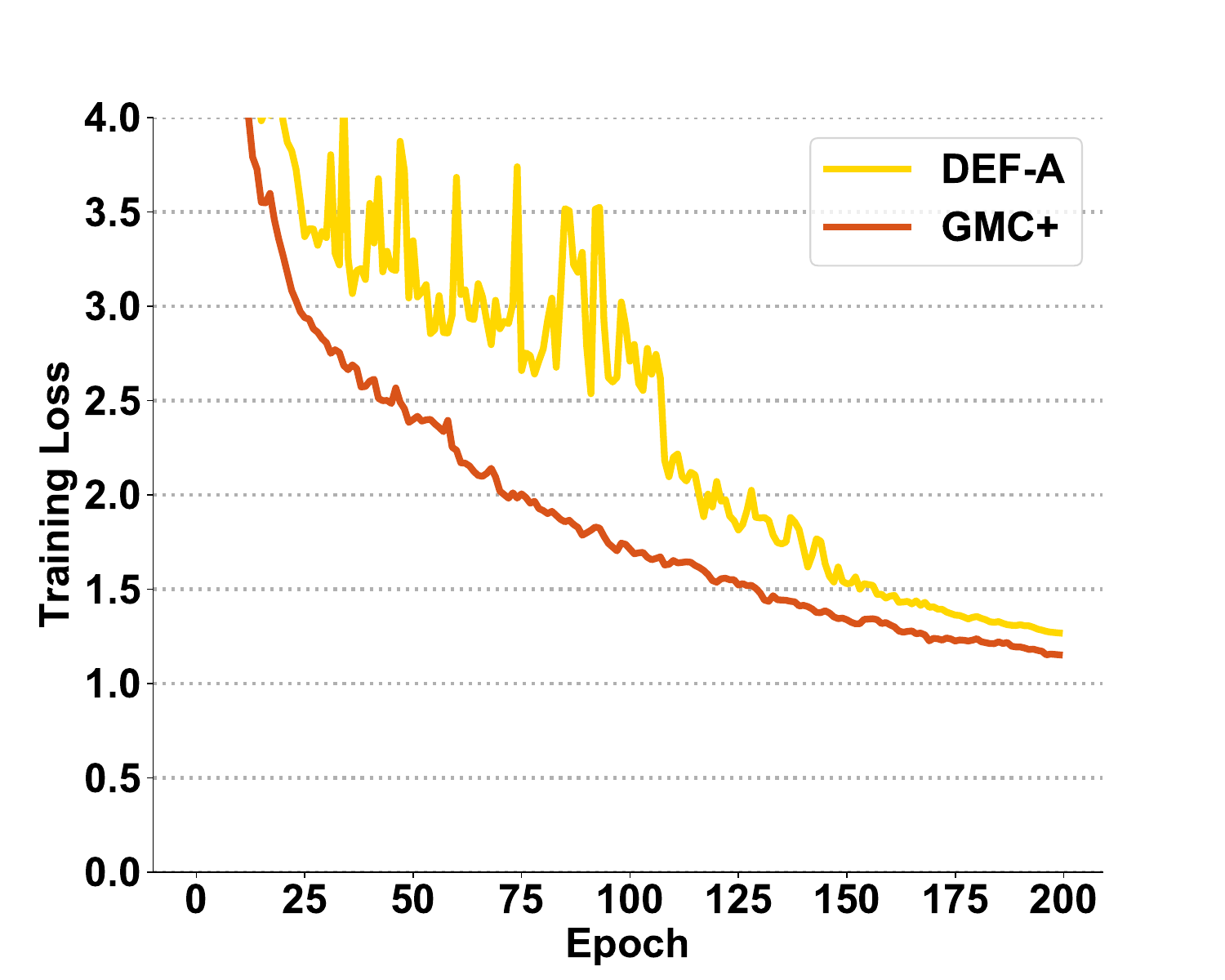}\vspace{-2pt}
        \includegraphics[width=1\linewidth]{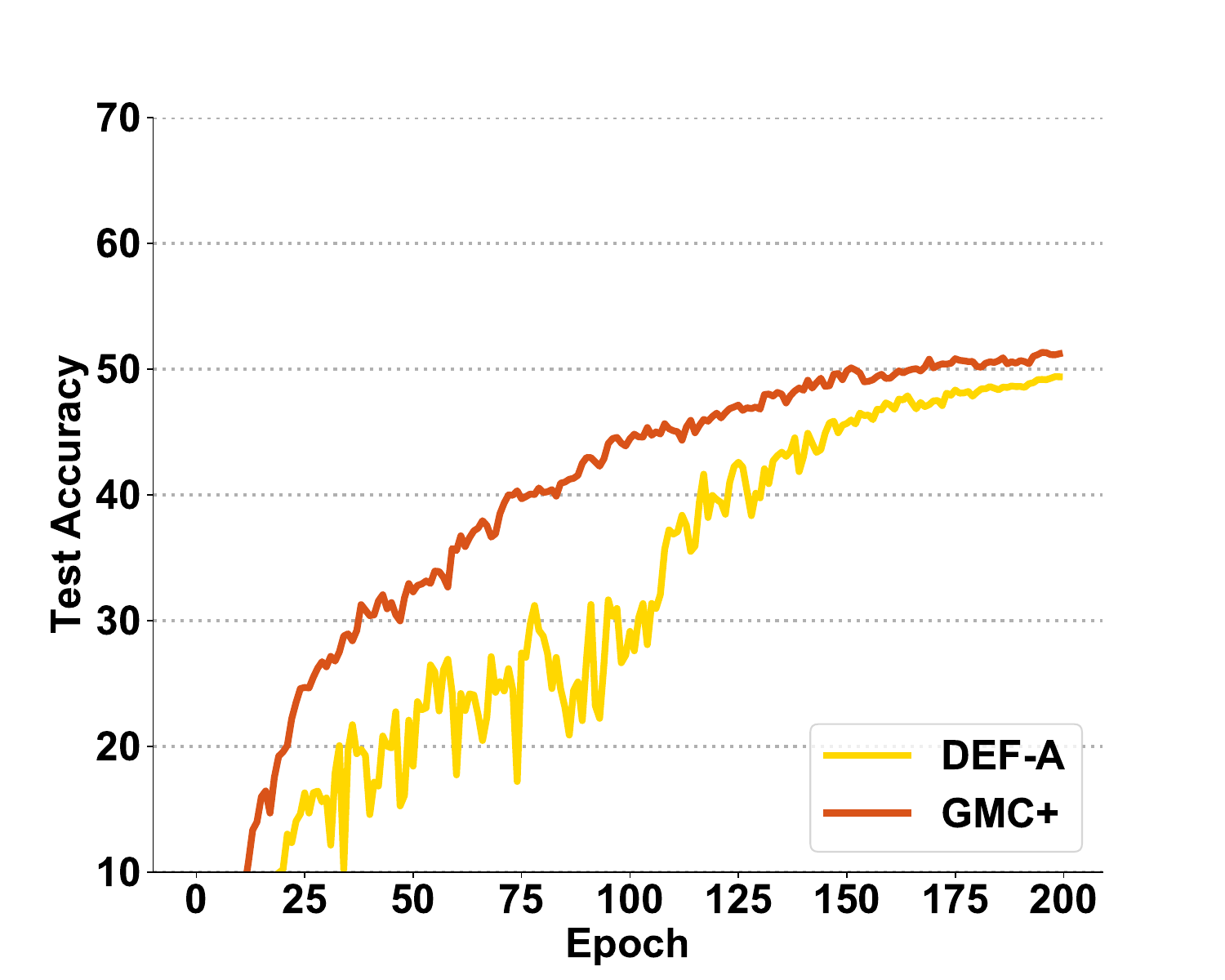}
        \end{minipage}}
    \subfigure[ViT, CIFAR100]{
      \begin{minipage}[b]{0.23\textwidth}
        \includegraphics[width=1\linewidth]{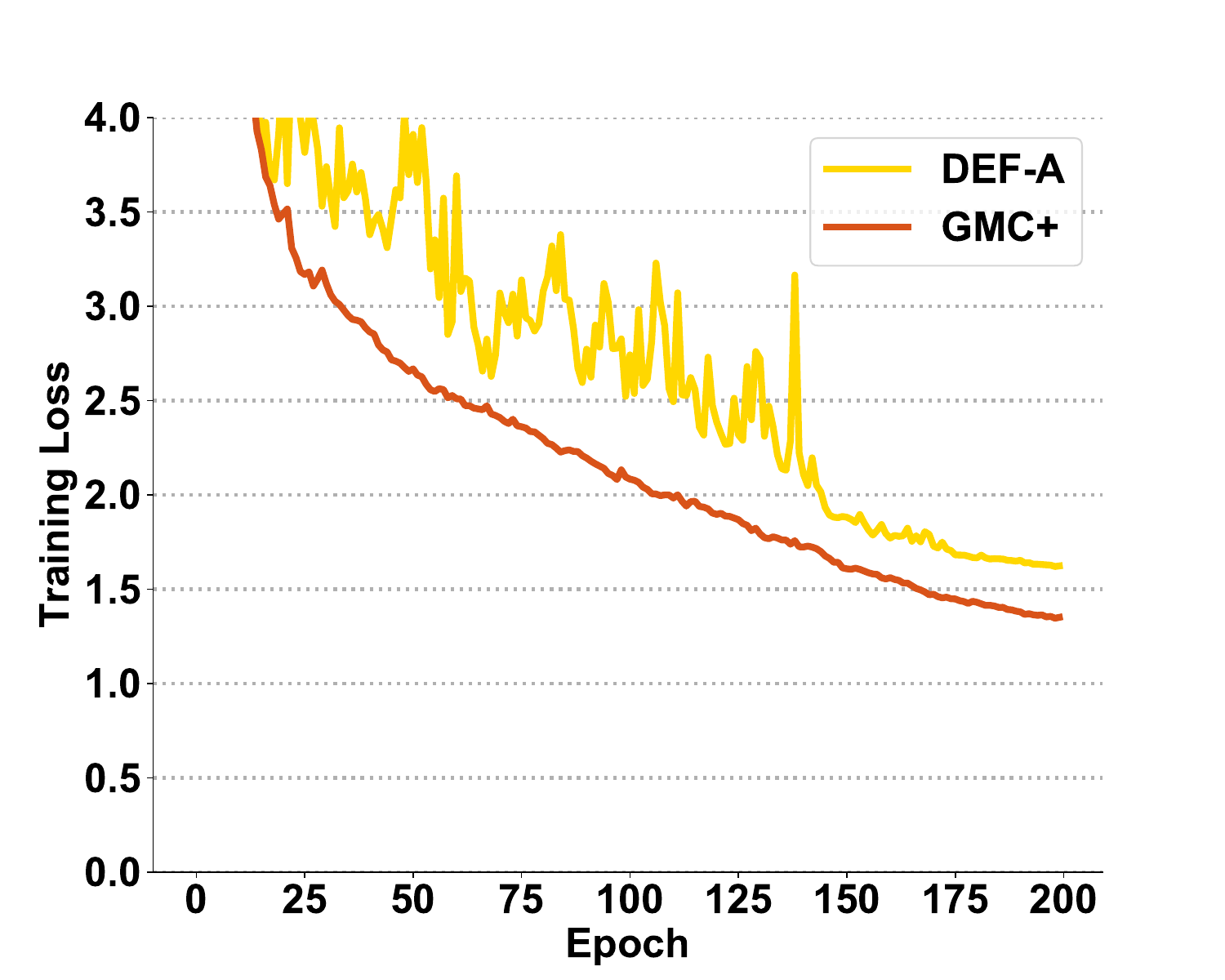}\vspace{-2pt}
        \includegraphics[width=1\linewidth]{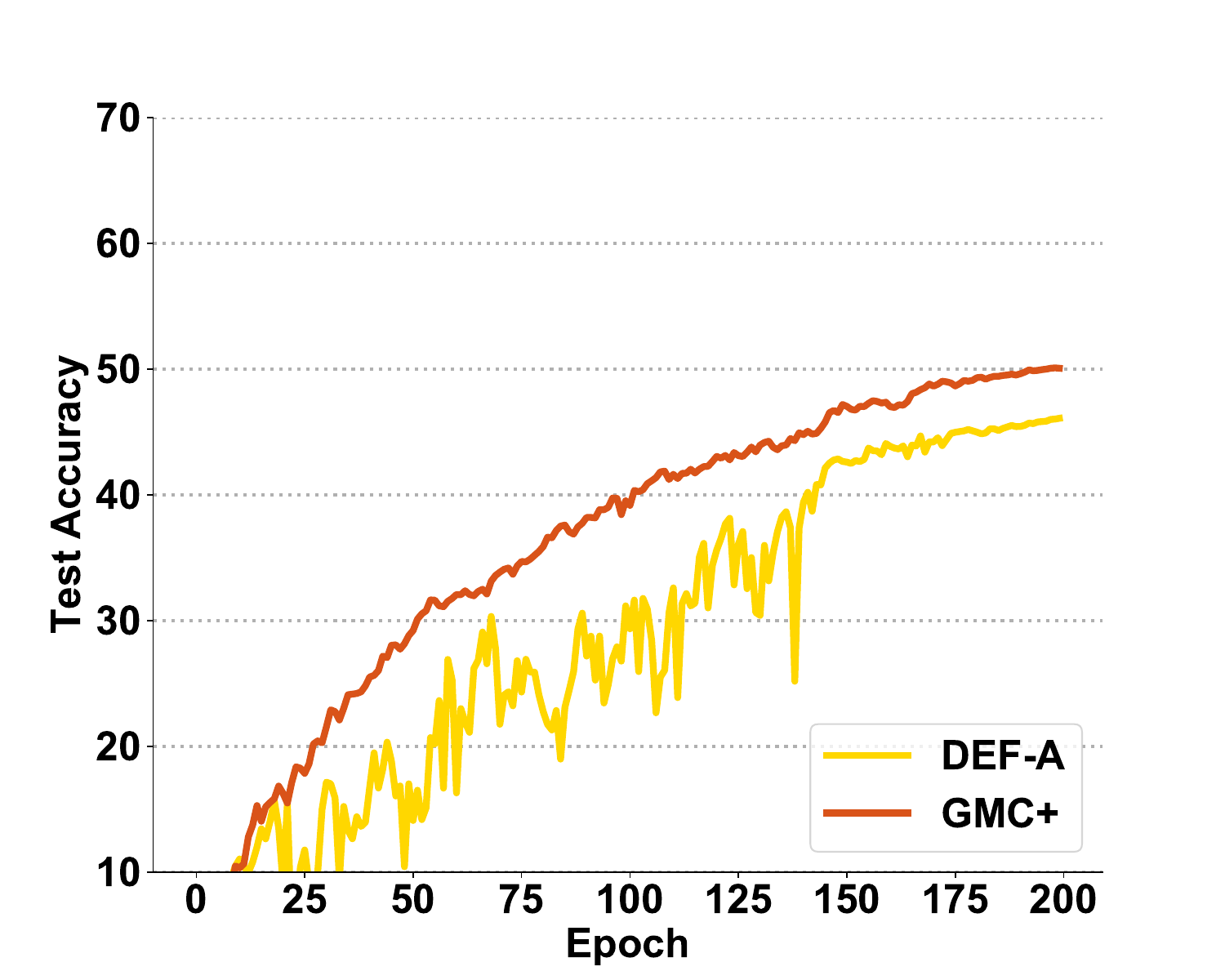}
        \end{minipage}}
  \caption{Training curves of DEF-A and GMC$+$ under non-IID data distribution.}\label{fig:DEF-NONIID}
  \end{figure*}

Table~\ref{tab:DEF-A}, Figure~\ref{fig:DEF-IID} and Figure~\ref{fig:DEF-NONIID} show the performance of DEF-A and GMC$+$. Compared with DEF-A, the global momentum method GMC+ demonstrates better test accuracy and stable convergence behavior for both IID and non-IID cases.

  \section{Conclusion}
  In this paper, we propose a novel method, called \emph{\underline{g}}lobal \emph{\underline{m}}omentum \emph{\underline{c}}ompression~(GMC), for sparse communication in distributed learning. To the best of our knowledge, this is the first work that introduces global momentum for sparse communication in DMSGD. Furthermore, to enhance the convergence performance when using more aggressive sparsification compressors (e.g., RBGS), we extend GMC to GMC+. We prove the convergence of GMC and GMC+ theoretically. Empirical results verify the superiority of global momentum and show that GMC and GMC+ can outperform other baselines to achieve state-of-the-art performance.

\appendix

\section{Other Ways for Combining Momentum and Error Feedback}\label{appendix:global momentum}
It is easy to get one way to combine error feedback and momentum by which  we can put the momentum term $\w_t - \w_{t-1}$ on the server. After receiving sparsified vectors from workers, the server updates the parameter using the momentum. We briefly present it in Algorithm~\ref{alg:trivial1}.

\begin{algorithm}[!thb]
\caption{Error Feedback with Momentum~(momentum on server)}\label{alg:trivial1}
\begin{algorithmic}
\FOR {iteration $t \in [T]$}
\STATE \underline{Workers:}
\FOR {worker $k \in [K]$ parallelly}
\STATE Randomly pick a mini-batch of training data $\IM_{t,k}\subseteq \DM_k$ with $|\IM_{t,k}| = b$ and compute $\nabla f(\w_t;\IM_{t,k}) = \frac{1}{b}\sum_{\xi \in \IM_{t,k}}\nabla f(\w_t;\xi)$;
\STATE $\e_{t+\frac{1}{2},k} = \e_{t,k}+\nabla f(\w_t;\IM_{t,k})$;
\STATE Generate a sparse vector $\mathcal{C}(\e_{t+\frac{1}{2},k})$ and send $\mathcal{C}(\e_{t+\frac{1}{2},k})$ to the server;
\STATE Update the error residual $\e_{t+1,k} = \e_{t+\frac{1}{2},k}-\mathcal{C}(\e_{t+\frac{1}{2},k})$;
\STATE Receive $\w_{t+1} - \w_{t}$ from server;
\STATE Get $\w_{t+1}$ by $\w_{t+1} = \w_t+(\w_{t+1}-\w_t)$;   
\ENDFOR
\STATE \underline{Server:}
\STATE Receive $\mathcal{C}(\e_{t+\frac{1}{2},k}), k \in [K]$ from all the workers;
\STATE $\w_{t+1} = \w_t - \eta \frac{1}{K}\sum_{k \in [K]}\mathcal{C}(\e_{t+\frac{1}{2},k}) + \beta(\w_t - \w_{t-1})$;
\STATE Send $\w_{t+1} - \w_t$ to workers;
\ENDFOR
\end{algorithmic}
\end{algorithm}

Let $\{\w_t\}$, $\{\e_{t,k}\}$ be the sequences produced by Algorithm \ref{alg:trivial1}, then we can get that
\begin{align*}
	\w_{t+1} = & \w_t - \eta \frac{1}{K}\sum_{k \in [K]}\mathcal{C}(\e_{t,k}+\nabla f(\w_t;\IM_{t,k})) + \beta(\w_t - \w_{t-1}), \\
	\e_{t+1,k} = & \e_{t,k}+\nabla f(\w_t;\IM_{t,k}) - \mathcal{C}(\e_{t,k}+\nabla f(\w_t;\IM_{t,k})), \forall k \in [K].
\end{align*}
By eliminating $\mathcal{C}(\cdot)$, we obtain the same equation as that in (\ref{eq:update rule}). Hence, our convergence analysis is also suitable for Algorithm~\ref{alg:trivial1}. The difference from GMC is that the error residual $\e_{t,k}$ in Algorithm~\ref{alg:trivial1} only keeps the compressed error from the stochastic gradients $\nabla f(\w_t;\IM_{t,k})$.  Compared with GMC, the disadvantage is that its error residual does not contain the momentum information which can play the role of correcting the update direction. Hence, GMC gets better performance than Algorithm~\ref{alg:trivial1}. \cite{DBLP:conf/iclr/LinHM0D18} also points out that Algorithm~\ref{alg:trivial1} has some  loss of performance.

Another way is to put the learning rate $\eta$ inside the error residual. We briefly present it in Algorithm~\ref{alg:trivial2}.
\begin{algorithm}[!thb]
\caption{Error Feedback with Momentum~(lr inside error residual)}\label{alg:trivial2}
\begin{algorithmic}
\FOR {iteration $t \in [T]$}
\STATE \underline{Workers:}
\FOR {worker $k \in [K]$ parallelly}
\STATE Randomly pick a mini-batch of training data $\IM_{t,k}\subseteq \DM_k$ with $|\IM_{t,k}| = b$ and compute $\nabla f(\w_t;\IM_{t,k}) = \frac{1}{b}\sum_{\xi \in \IM_{t,k}}\nabla f(\w_t;\xi)$;
\STATE $\e_{t+\frac{1}{2},k} = \e_{t,k}+\eta \nabla f(\w_t;\IM_{t,k})-\beta (\w_{t}-\w_{t-1})$;
\STATE Generate a sparse vector $\mathcal{C}(\e_{t+\frac{1}{2},k})$ and send $\mathcal{C}(\e_{t+\frac{1}{2},k})$ to the server;
\STATE Update the error residual $\e_{t+1,k} = \e_{t+\frac{1}{2},k}-\mathcal{C}(\e_{t+\frac{1}{2},k})$;
\STATE Receive $\w_{t+1} - \w_{t}$ from server;
\STATE Get $\w_{t+1}$ by $\w_{t+1} = \w_t+(\w_{t+1}-\w_t)$;     
\ENDFOR
\STATE \underline{Server:}
\STATE $\w_{t+1} = \w_t - \frac{1}{K}\sum_{k \in [K]}\mathcal{C}(\e_{t+\frac{1}{2},k})$;
\STATE Send $\w_{t+1} - \w_t$ to workers;
\ENDFOR
\end{algorithmic}
\end{algorithm}
If the learning rate $\eta$ is constant during the whole training process, Algorithm~\ref{alg:trivial2} is equivalent to Algorithm~\ref{alg:gmc}. 
However, we often use the learning rate decay strategy for deep model training in practice. We use $\eta_t$ to denote the learning rate in iteration $t$. 
The update rule in Algorithm~\ref{alg:trivial2} can be rewritten as:
\begin{align*}
  \w_{t+1} &= \w_t - \frac{1}{K}\sum_{k \in [K]}\mathcal{C}(\e_{t,k}+\eta_t \nabla f(\w_t;\IM_{t,k})-\beta (\w_{t}-\w_{t-1})), 
\end{align*}
where
\begin{align*}
  \e_{t,k} = \e_{t-\frac{1}{2}, k} - \mathcal{C}(\e_{t-\frac{1}{2}, k}), \e_{t-\frac{1}{2}, k} = \e_{t-1,k}+\eta_{t-1} \nabla f(\w_{t-1};\IM_{t-1,k})-\beta (\w_{t-1}-\w_{t-2}).
\end{align*}
The update rule in GMC can be rewritten as:
\begin{align*}
  \w_{t+1} &= \w_t- \frac{1}{K}\sum_{k \in [K]}\mathcal{C}(\eta_t\e_{t,k}^{GMC}+\eta_t \nabla f(\w_t;\IM_{t,k})-\beta (\w_{t}-\w_{t-1})), 
\end{align*}
 where
 \begin{align*}
  \eta_t \e_{t,k}^{GMC} &= \eta_t \e_{t-\frac{1}{2},k}^{GMC}-\mathcal{C}(\eta_t \e_{t-\frac{1}{2},k}^{GMC}), \\
  \eta_t \e_{t-\frac{1}{2},k}^{GMC} &= \frac{\eta_t}{\eta_{t-1}}[\eta_{t-1}\e_{t-1,k}^{GMC}+ \eta_{t-1}\nabla f(\w_{t-1};\IM_{t-1,k})-\beta (\w_{t-1}-\w_{t-2})].
\end{align*} 

\begin{figure*}[!thb]
  \centering
  \subfigure[IID]{
    \label{fig:appendix_iid}
    \begin{minipage}[b]{0.46\textwidth}
      \includegraphics[width=0.48\linewidth]{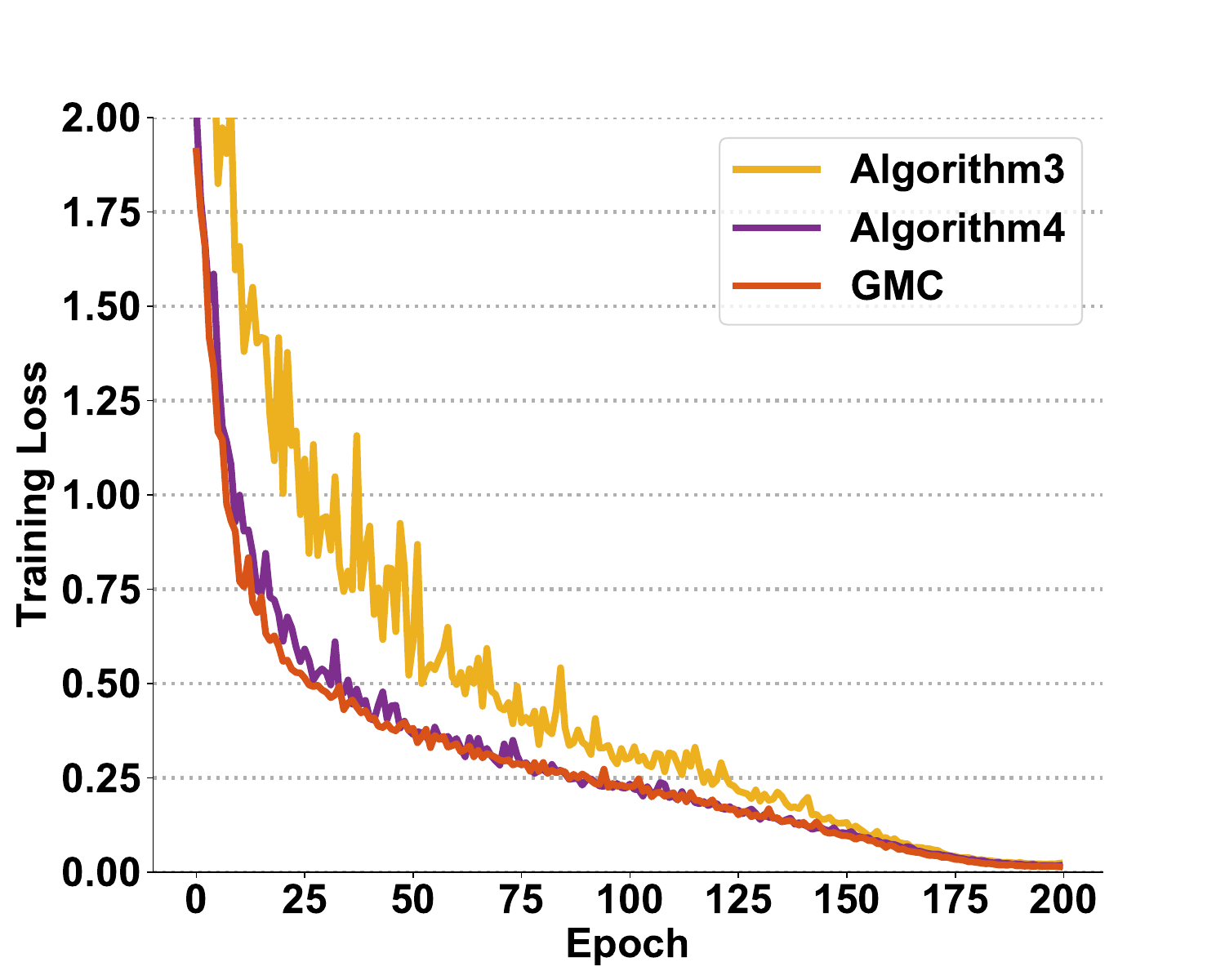}
      \includegraphics[width=0.48\linewidth]{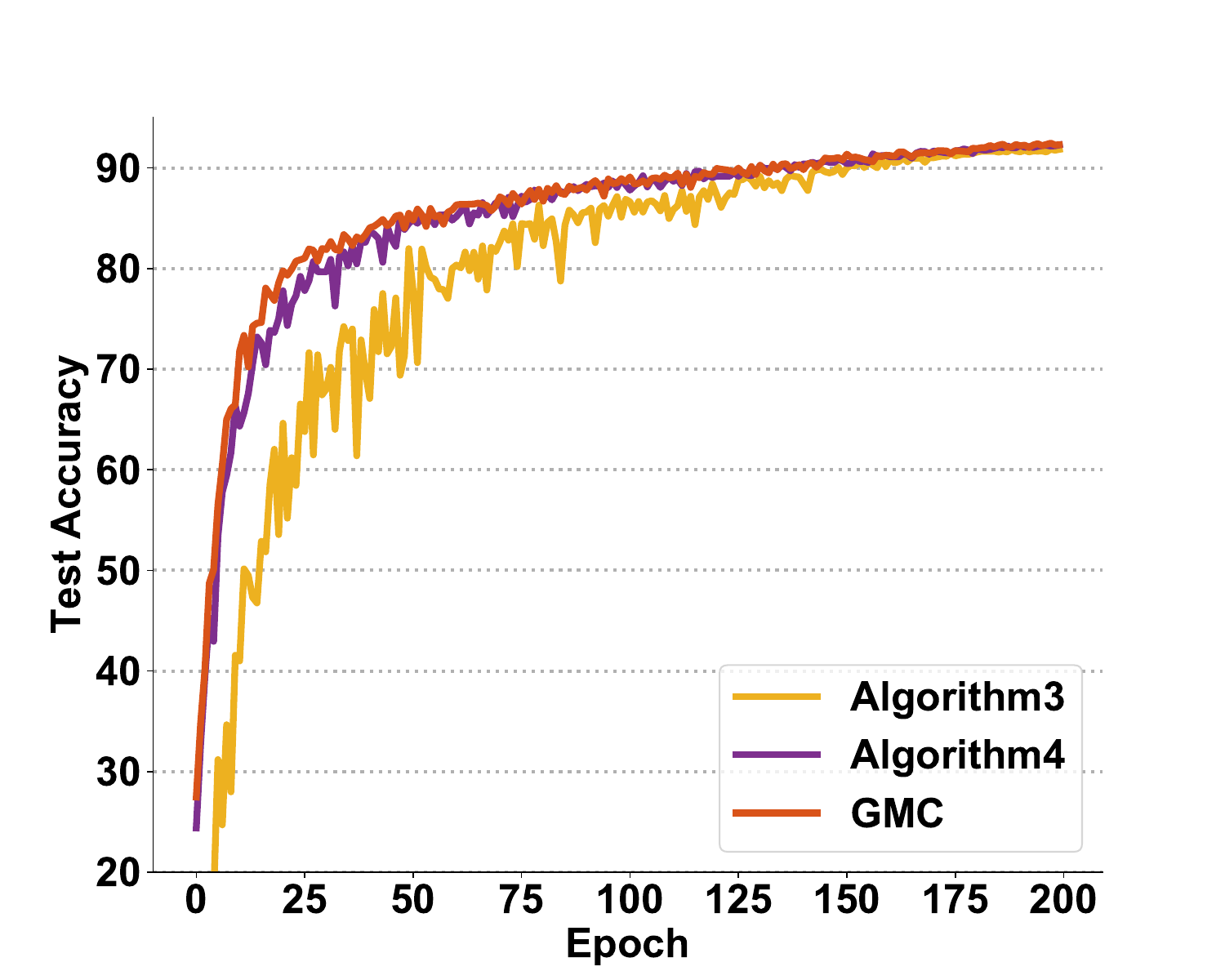}
      \end{minipage}}
  \subfigure[non-IID]{
    \label{fig:appendix_noniid}
    \begin{minipage}[b]{0.46\textwidth}
      \includegraphics[width=0.48\linewidth]{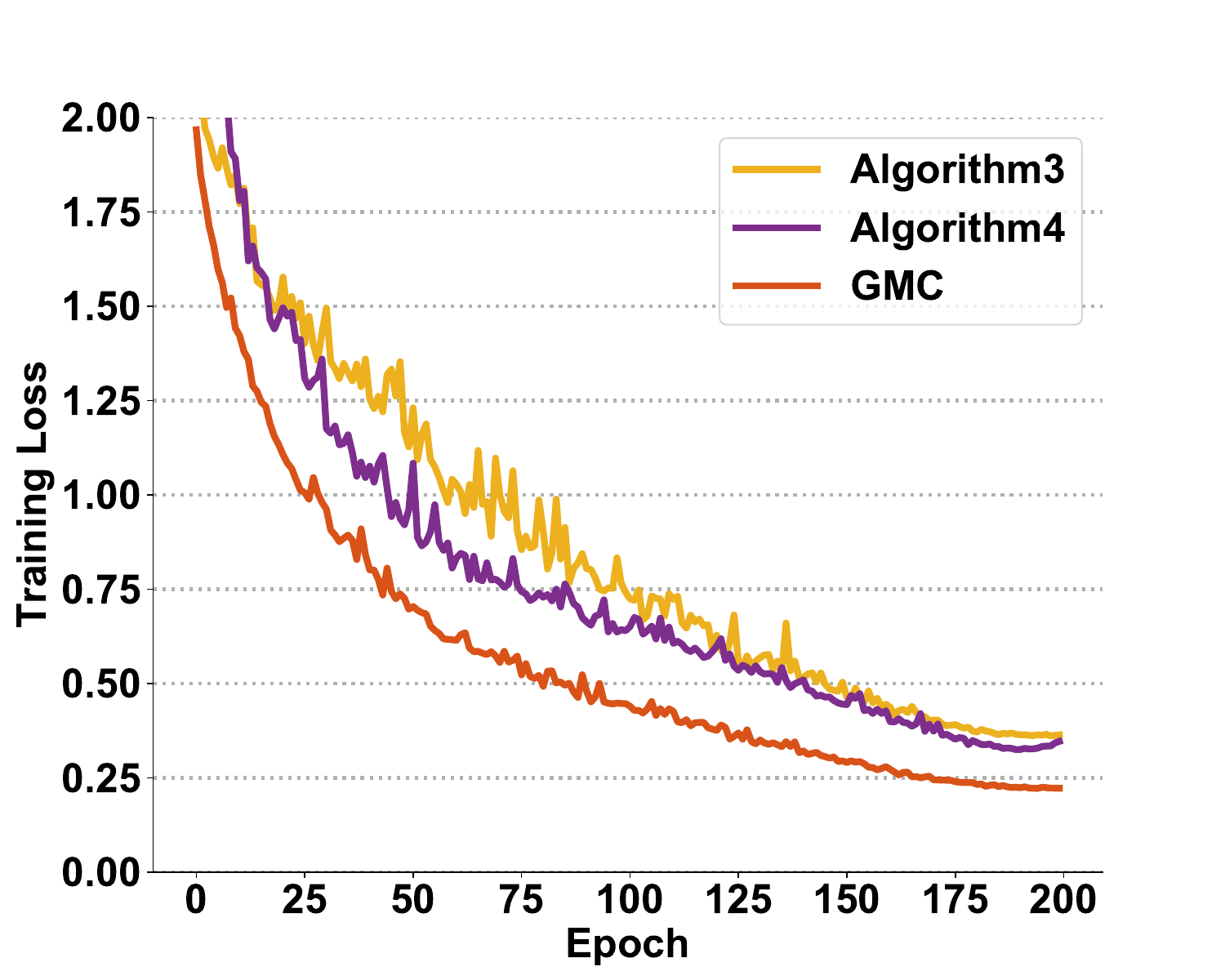}
      \includegraphics[width=0.48\linewidth]{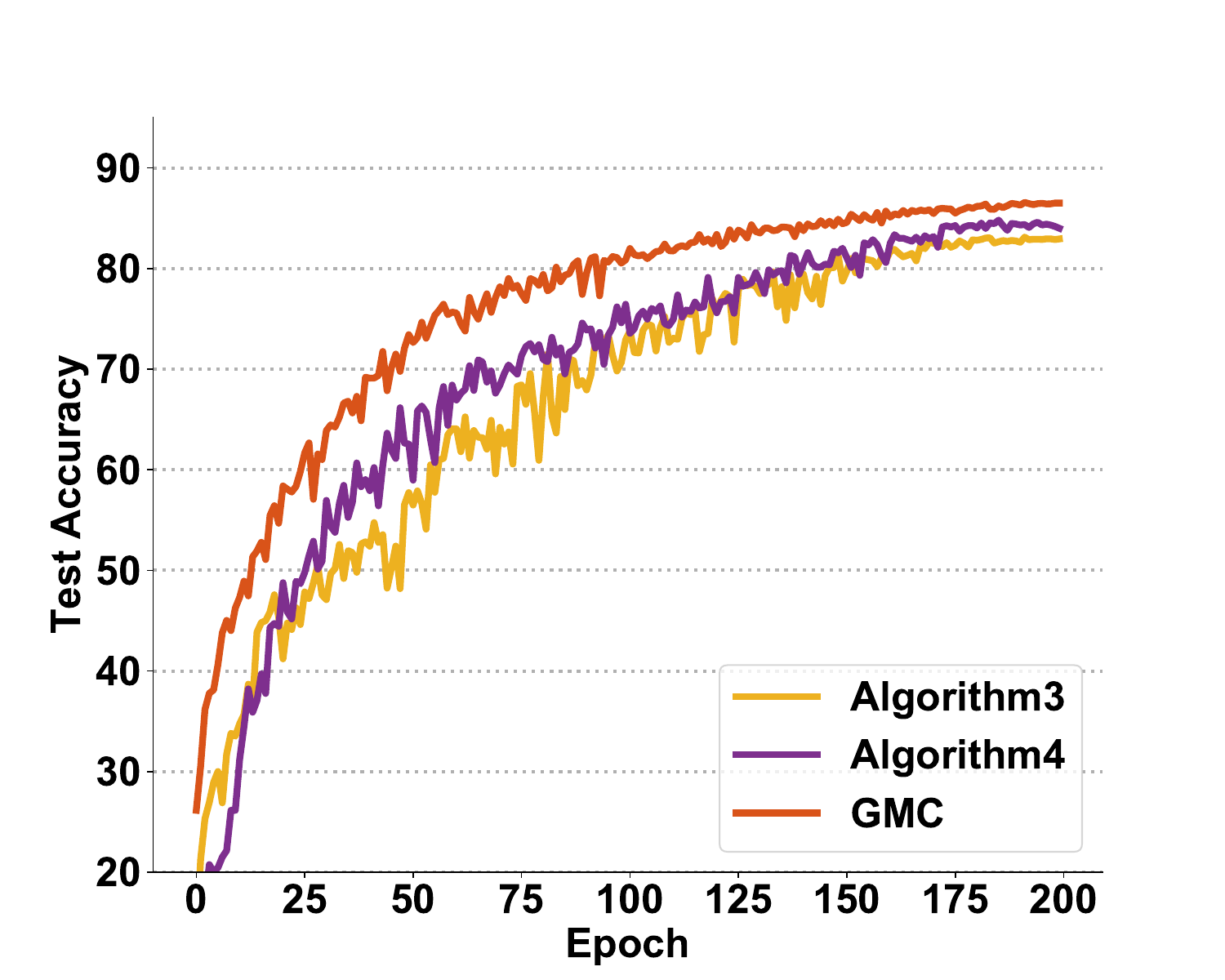}
      \end{minipage}}
\caption{Comparing GMC with Algorithm~\ref{alg:trivial1} and Algorithm~\ref{alg:trivial2}}\label{fig:appendix}
\end{figure*}

We compare $\eta_t \e_{t,k}^{GMC}$ in GMC and $\e_{t,k}$ in Algorithm~\ref{alg:trivial2}. Since $\{\eta_t\}$ is non-increasing, $\frac{\eta_t}{\eta_{t-1}} \leq 1$.
Compared with GMC, the error residual in Algorithm~\ref{alg:trivial2} participates in parameter updates with a larger coefficient, which will deteriorate the convergence.
Figure~\ref{fig:appendix} shows the learning process of ResNet20 on CIFAR10. GMC is better than Algorithm~\ref{alg:trivial1} and Algorithm~\ref{alg:trivial2}, especially under non-IID data distribution. 

\section{Convergence Analysis of GMC} \label{CAGMC}
\subsection{Proof of Lemma~\ref{lemma:w to z}}
  \begin{align*}
    \z_{t+1} = & \w_{t+1} + \frac{\beta}{1-\beta}(\w_{t+1} - \w_{t}) - \frac{\eta}{1-\beta}\bar{\e}_{t+1} \\
             = & \frac{1}{1-\beta}(\w_{t+1} - \beta\w_t) - \frac{\eta}{1-\beta}\bar{\e}_{t+1} \\
             = & \frac{1}{1-\beta}(\w_t - \eta\nabla f(\w_t;\IM_t) + \beta(\w_t - \w_{t-1}) - \eta\bar{\e}_{t} + \eta\bar{\e}_{t+1} - \beta\w_t) - \frac{\eta}{1-\beta}\bar{\e}_{t+1} \\
             = & \frac{1}{1-\beta}(\w_t - \beta \w_{t-1} - \eta\nabla f(\w_t;\IM_t) - \eta\bar{\e}_{t}) \\
             = & \frac{1}{1-\beta}(\w_t - \beta \w_{t-1}) - \frac{\eta}{1-\beta}\bar{\e}_{t} - \frac{\eta}{1-\beta}\nabla f(\w_t;\IM_t)\\
             = & \z_t - \frac{\eta}{1-\beta}\nabla f(\w_t;\IM_t).
  \end{align*}
  \subsection{Proof of Lemma~\ref{lemma:bounded error}}
 \begin{align*}
  \e_{t+\frac{1}{2},k}&= \e_{t,k}+\nabla f(\w_t;\IM_{t,k})-\frac{\beta}{\eta}(\w_t-\w_{t-1}) \\ 
  &= \e_{t,k}+\nabla f(\w_t;\IM_{t,k})+ \frac{\beta}{K}\sum_{k' \in [K]}\mathcal{C}(\e_{t-\frac{1}{2},k'}) \\
  &= \e_{t-\frac{1}{2},k}-\mathcal{C}(\e_{t-\frac{1}{2},k})+\frac{\beta}{K}\mathcal{C}(\e_{t-\frac{1}{2},k})+\nabla f(\w_t;\IM_{t,k})+ \frac{\beta}{K}\sum_{k' \in [K], k' \neq k}\mathcal{C}(\e_{t-\frac{1}{2},k'}) \\
  &= \e_{t-\frac{1}{2},k}-(1-\frac{\beta}{K})\mathcal{C}(\e_{t-\frac{1}{2},k})+\nabla f(\w_t;\IM_{t,k})+ \frac{\beta}{K}\sum_{k' \in [K], k' \neq k}\mathcal{C}(\e_{t-\frac{1}{2},k'}) \\
  &= (1-\frac{\beta}{K})[\e_{t-\frac{1}{2},k}-\mathcal{C}(\e_{t-\frac{1}{2},k})]+\nabla f(\w_t;\IM_{t,k})+\frac{\beta}{K}\e_{t-\frac{1}{2},k}+ \frac{\beta}{K}\sum_{k' \in [K], k' \neq k}\mathcal{C}(\e_{t-\frac{1}{2},k'}). 
\end{align*}
Since a sparsification compressor only selects a few components of the original vector, it's easy to verify that $\mathbb{E}_{\mathcal{C}}\|\mathcal{C}(\w)\|^2 \leq \|\w\|^2, \forall \w \in \mathbb{R}^d$. 
\begin{align*}
  & \mathbb{E}\|\e_{t+\frac{1}{2},k}\|^2 \\
   &= \mathbb{E}\|(1-\frac{\beta}{K})[\e_{t-\frac{1}{2},k}-\mathcal{C}(\e_{t-\frac{1}{2},k})]+\nabla f(\w_t;\IM_{t,k})+\frac{\beta}{K}\e_{t-\frac{1}{2},k}+ \frac{\beta}{K}\sum_{k' \in [K], k' \neq k}\mathcal{C}(\e_{t-\frac{1}{2},k'}) \|^2 \\
   &\leq  (1+a)\mathbb{E}\|(1-\frac{\beta}{K})[\e_{t-\frac{1}{2},k}-\mathcal{C}(\e_{t-\frac{1}{2},k})]+\frac{\beta}{K}\e_{t-\frac{1}{2},k}+ \frac{\beta}{K}\sum_{k' \in [K], k' \neq k}\mathcal{C}(\e_{t-\frac{1}{2},k'})\|^2  \\
   & ~~~~~+(1+\frac{1}{a})\mathbb{E}\|\nabla f(\w_t;\IM_{t,k})\|^2 \\
   &\leq  (1+a)(1+b)(1-\frac{\beta}{K})^2\mathbb{E}\|\e_{t-\frac{1}{2},k}-\mathcal{C}(\e_{t-\frac{1}{2},k})\|^2\\
   &~~~~~+(1+a)(1+\frac{1}{b})\mathbb{E}\|\frac{\beta}{K}\e_{t-\frac{1}{2},k}+\frac{\beta}{K}\sum_{k' \in [K], k' \neq k}\mathcal{C}(\e_{t-\frac{1}{2},k'})\|^2 +(1+\frac{1}{a})\mathbb{E}\|\nabla f(\w_t;\IM_{t,k})\|^2 \\
   &\leq  (1+a)(1+b)(1-\frac{\beta}{K})^2\mathbb{E}\|\e_{t-\frac{1}{2},k}-\mathcal{C}(\e_{t-\frac{1}{2},k})\|^2+(1+a)(1+\frac{1}{b})\beta^2\frac{1}{K}\mathbb{E}\|\e_{t-\frac{1}{2},k}\|^2 \\
   &~~~~~+(1+a)(1+\frac{1}{b})\beta^2 \frac{1}{K}\sum_{k' \in [K], k' \neq k}\mathbb{E}\|\mathcal{C}(\e_{t-\frac{1}{2},k'})\|^2 +(1+\frac{1}{a})\mathbb{E}\|\nabla f(\w_t;\IM_{t,k})\|^2 \\
  &\leq  (1+a)(1+b)(1-\frac{\beta}{K})^2(1-\delta)\mathbb{E}\|\e_{t-\frac{1}{2},k}\|^2+(1+a)(1+\frac{1}{b})\beta^2\frac{1}{K}\sum_{k' \in [K]}\mathbb{E}\|\e_{t-\frac{1}{2},k'}\|^2 \\
  &~~~~~ +(1+\frac{1}{a})G^2. 
    \end{align*}  
  Summing up the above equation from $k=0$ to $K-1$, we can get
    \begin{align*}
      \frac{1}{K}\sum_{k \in [K]}\mathbb{E}\|\e_{t+\frac{1}{2},k}\|^2 & \leq  [(1+a)(1+b)(1-\frac{\beta}{K})^2(1-\delta)+(1+a)(1+\frac{1}{b})\beta^2]\frac{1}{K}\sum_{k \in [K]}\mathbb{E}\|\e_{t-\frac{1}{2},k}\|^2\\
      &~~~~~ +(1+\frac{1}{a})G^2. 
      \end{align*}  
    Let $a = \frac{\delta}{4}$, $b = \frac{\delta}{2}$, we get
    \begin{align*}
      &\frac{1}{K}\sum_{k \in [K]}\mathbb{E}\|\e_{t+\frac{1}{2},k}\|^2 \\
      & \leq  [(1+\frac{\delta}{4})(1+\frac{\delta}{2})(1-\delta)(1-\frac{\beta}{K})^2+(1+\frac{\delta}{4})(1+\frac{2}{\delta})\beta^2]\frac{1}{K}\sum_{k \in [K]}\mathbb{E}\|\e_{t-\frac{1}{2},k}\|^2 +(1+\frac{4}{\delta})G^2 \\
      & \leq  [(1-\frac{\delta}{4})(1-\frac{\beta}{K})^2+(1+\frac{\delta}{4})(1+\frac{2}{\delta})\beta^2]\frac{1}{K}\sum_{k \in [K]}\mathbb{E}\|\e_{t-\frac{1}{2},k}\|^2 +(1+\frac{4}{\delta})G^2. 
    \end{align*}   
    Assuming $\beta \leq \frac{\delta}{4\sqrt{2+\delta}}$, we have
    \begin{align*}
      (1-\frac{\delta}{4})(1-\frac{\beta}{K})^2+(1+\frac{\delta}{4})(1+\frac{2}{\delta})\beta^2 \leq (1-\frac{\delta}{4})+2(1+\frac{2}{\delta})\beta^2 \leq 1- \frac{\delta}{8} < 1.
      \end{align*} 
    Then we get 
    \begin{align*}
      \frac{1}{K}\sum_{k \in [K]}\mathbb{E}\|\e_{t+\frac{1}{2},k}\|^2 &\leq  [(1-\frac{\delta}{4})(1-\frac{\beta}{K})^2+(1+\frac{\delta}{4})(1+\frac{2}{\delta})\beta^2]\frac{1}{K}\sum_{k \in [K]}\mathbb{E}\|\e_{t-\frac{1}{2},k}\|^2 +(1+\frac{4}{\delta})G^2 \\
      &\leq  \frac{(1+\frac{4}{\delta})G^2}{1-[(1-\frac{\delta}{4})(1-\frac{\beta}{K})^2+(1+\frac{\delta}{4})(1+\frac{2}{\delta})\beta^2]}. 
    \end{align*} 
    \begin{align*}
      \frac{1}{K}\sum_{k \in [K]}\mathbb{E}\|\e_{t+1,k}\|^2 &= \frac{1}{K}\sum_{k \in [K]}\mathbb{E}\|\e_{t+\frac{1}{2},k}-\mathcal{C}(\e_{t+\frac{1}{2},k})\|^2 \leq  (1-\delta)\frac{1}{K}\sum_{k \in [K]}\mathbb{E}\|\e_{t+\frac{1}{2},k}\|^2 \\
    &\leq \frac{(1-\delta)(1+\frac{4}{\delta})G^2}{1-[(1-\frac{\delta}{4})(1-\frac{\beta}{K})^2+(1+\frac{\delta}{4})(1+\frac{2}{\delta})\beta^2]}.
    \end{align*}

\subsection{Proof of Lemma~\ref{lemma:bounded gap}}
\begin{align*}
  \mathbb{E}\|\w_{t}-\w_{t-1}\|^2 &= \mathbb{E}\|\beta(\w_{t-1}-\w_{t-2})+\eta \bar{\e}_{t}-\eta \bar{\e}_{t-1}-\eta \nabla f(\w_{t-1};\IM_{t-1})\|^2\\
  &\leq  (1+a)\beta^2 \mathbb{E}\|\w_{t-1}-\w_{t-2}\|^2+(1+\frac{1}{a})\eta^2\mathbb{E}\|\bar{\e}_{t}- \bar{\e}_{t-1}- \nabla f(\w_{t-1};\IM_{t-1})\|^2\\
  &\leq  (1+a)\beta^2 \mathbb{E}\|\w_{t-1}-\w_{t-2}\|^2+3(1+\frac{1}{a})(2 E^2+G^2)\eta^2.
\end{align*}
If $\beta \in (0,1)$, let $a=\frac{1}{\beta}-1$, then we have
\begin{align*}
  \mathbb{E}\|\w_{t}-\w_{t-1}\|^2 \leq \beta \mathbb{E}\|\w_{t-1}-\w_{t-2}\|^2+\frac{3 (2E^2+G^2)}{1-\beta}\eta^2 \leq \frac{3 (2E^2+G^2)}{(1-\beta)^2}\eta^2.
\end{align*}    
If $\beta = 0$, it's obvious that $\mathbb{E}\|\w_{t}-\w_{t-1}\|^2$ can be bounded by $\frac{3 (2E^2+G^2)}{(1-\beta)^2}\eta^2$.

\begin{align*}
  &\mathbb{E}\|\z_t - \w_t\|^2 \\
  &= \mathbb{E}\|\frac{\beta}{1-\beta}(\w_{t} - \w_{t-1}) - \frac{\eta}{1-\beta}\bar{\e}_{t}\|^2 \leq  2\frac{\beta^2}{(1-\beta)^2}\mathbb{E}\|\w_{t} - \w_{t-1}\|^2 +\frac{2\eta^2}{(1-\beta)^2}\mathbb{E}\|\bar{\e}_{t}\|^2 \\
      &\leq  2\frac{\beta^2}{(1-\beta)^2}\mathbb{E}\|\w_{t} - \w_{t-1}\|^2 +\frac{2\eta^2}{(1-\beta)^2}E^2 \leq  [\frac{6 (2E^2+G^2)\beta^2}{(1-\beta)^4} +\frac{2E^2}{(1-\beta)^2}]\eta^2. 
    \end{align*} 

\subsection{Proof of Theorem~\ref{theorem:gmc}}
\begin{align*}
  F(\z_{t+1}) &\leq  F(\z_t)-\frac{\eta}{1-\beta}\nabla F(\z_t)^{T}\nabla f(\w_t;\IM_t)+ \frac{L\eta^2}{2(1-\beta)^2}\|\nabla f(\w_t;\IM_t)\|^2 \\
  \mathbb{E} F(\z_{t+1}) &\leq  F(\z_t)-\frac{\eta}{1-\beta}\nabla F(\z_t)^{T}\mathbb{E}[\nabla f(\w_t;\IM_t)]+ \frac{L\eta^2}{2(1-\beta)^2} \mathbb{E}\|\nabla f(\w_t;\IM_t)\|^2 \\
  &=  F(\z_t)-\frac{\eta}{1-\beta}\nabla F(\z_t)^{T}\nabla F(\w_t)+ \frac{L\eta^2}{2(1-\beta)^2} \mathbb{E}\|\nabla f(\w_t;\IM_t)\|^2. 
\end{align*}

\begin{align*}
-\frac{\eta}{1-\beta}\nabla F(\z_t)^{T}\nabla F(\w_t)&=-\frac{\eta}{1-\beta}(\nabla F(\z_t)-\nabla F(\w_t))^{T}\nabla F(\w_t)-\frac{\eta}{1-\beta}\|\nabla F(\w_t)\|^2\\
    & \leq \frac{\eta}{2(1-\beta)}\|\nabla F(\z_t)-\nabla F(\w_t)\|^2 -\frac{\eta}{2(1-\beta)}\|\nabla F(\w_t)\|^2\\
    & \leq \frac{\eta L^2}{2(1-\beta)}\|\z_t-\w_t\|^2 -\frac{\eta}{2(1-\beta)}\|\nabla F(\w_t)\|^2\\
    & \leq \frac{C_1 L^2}{2(1-\beta)}\eta^3 -\frac{\eta}{2(1-\beta)}\|\nabla F(\w_t)\|^2.
  \end{align*}
\begin{align*}
  \mathbb{E}\|\nabla f(\w_{t};\IM_t)\|^2 &= \mathbb{E}\|\frac{1}{K}\sum_{k \in [K]}[\nabla f(\w_t;\IM_{t,k})-\nabla F_k(\w_t)]+\frac{1}{K}\sum_{k \in [K]}\nabla F_k(\w_t)]\|^2 \\
  &= \mathbb{E}\|\frac{1}{K}\sum_{k \in [K]}(\nabla f(\w_t;\IM_{t,k})-\nabla F_k(\w_t))\|^2+\|\nabla F(\w_t)\|^2 \\
  & \leq \frac{\sigma^2}{Kb}+\|\nabla F(\w_t)\|^2.  
\end{align*}
\begin{align*}
  \mathbb{E} F(\z_{t+1}) &\leq F(\z_t)-\frac{\eta}{1-\beta}\nabla F(\z_t)^{T}\nabla F(\w_t)+ \frac{L\eta^2}{2(1-\beta)^2} \mathbb{E}\|\nabla f(\w_t;\IM_t)\|^2 \\      
    & \leq F(\z_t) + \frac{C_1 L^2}{2(1-\beta)}\eta^3 + \frac{L\sigma^2}{2(1-\beta)^2b}\frac{\eta^2}{K} -(\frac{\eta}{2(1-\beta)}-\frac{L\eta^2}{2(1-\beta)^2})\|\nabla F(\w_t)\|^2.
\end{align*}
Since we assume $\eta \leq \frac{1-\beta}{2L}$, we have $-(\frac{\eta}{2(1-\beta)}-\frac{L\eta^2}{2(1-\beta)^2}) \leq -\frac{\eta}{4(1-\beta)}$.
\begin{align*}
  \mathbb{E} F(\z_{t+1}) & \leq \mathbb{E}F(\z_t) + \frac{C_1 L^2}{2(1-\beta)}\eta^3 + + \frac{L\sigma^2}{2(1-\beta)^2b}\frac{\eta^2}{K} -(\frac{\eta}{2(1-\beta)}-\frac{L\eta^2}{2(1-\beta)^2})\mathbb{E}\|\nabla F(\w_t)\|^2\\
  & \leq \mathbb{E}F(\z_t) + \frac{C_1 L^2}{2(1-\beta)}\eta^3 + \frac{L\sigma^2}{2(1-\beta)^2b}\frac{\eta^2}{K} -\frac{\eta}{4(1-\beta)}\mathbb{E}\|\nabla F(\w_t)\|^2. 
\end{align*}    
\begin{align*}
    \mathbb{E}\|\nabla F(\w_t)\|^2  &\leq \frac{4(1-\beta)(\mathbb{E} F(\z_t)- \mathbb{E} F(\z_{t+1}))}        {\eta}+ \frac{2L\sigma^2}{(1-\beta)b}\frac{\eta}{K}+ 2C_1 L^2\eta^2.  
\end{align*}
Summing up the above equation from $t=0$ to $T-1$, we have
\begin{align*}
  \frac{1}{T}\sum_{t \in [T]}\mathbb{E}\|\nabla F(\w_{t})\|^2\leq  \frac{4(1-\beta)(F(\z_{0})-F^*)}{T\eta} + \frac{2L\sigma^2}{(1-\beta)b}\frac{\eta}{K}+ 2C_1 L^2\eta^2.  
\end{align*}  

\section{Convergence Analysis of GMC+} \label{CAGMC+}

\subsection{Proof of Lemma~\ref{lemma:bounded error of def}}
The derivation details are similar to that in Lemma~\ref{lemma:bounded error}.
 \begin{align*}
  \e_{t+\frac{1}{2},k}&= \e_{t,k}+\nabla f(\w_{t,k};\IM_{t,k})-\frac{\beta}{\eta}(\w_t-\w_{t-1}) \\ 
                      &= \e_{t-\frac{1}{2},k}-\mathcal{C}(\e_{t-\frac{1}{2},k})+\nabla f(\w_{t,k};\IM_{t,k})+ \frac{\beta}{K}\sum_{k' \in [K]}\mathcal{C}(\e_{t-\frac{1}{2},k'}) \\
                      &= (1-\frac{\beta}{K})[\e_{t-\frac{1}{2},k}-\mathcal{C}(\e_{t-\frac{1}{2},k})]+\nabla f(\w_{t,k};\IM_{t,k})+\frac{\beta}{K}\e_{t-\frac{1}{2},k}+ \frac{\beta}{K}\sum_{k' \in [K], k' \neq k}\mathcal{C}(\e_{t-\frac{1}{2},k'}). 
\end{align*}

\begin{align*}
  & \mathbb{E}\|\e_{t+\frac{1}{2},k}\|^2 \\
   &= \mathbb{E}\|(1-\frac{\beta}{K})[\e_{t-\frac{1}{2},k}-\mathcal{C}(\e_{t-\frac{1}{2},k})]+\nabla f(\w_{t,k};\IM_{t,k})+\frac{\beta}{K}\e_{t-\frac{1}{2},k}+ \frac{\beta}{K}\sum_{k' \in [K], k' \neq k}\mathcal{C}(\e_{t-\frac{1}{2},k'}) \|^2 \\
  &\leq  (1+a)(1+b)(1-\frac{\beta}{K})^2(1-\delta)\mathbb{E}\|\e_{t-\frac{1}{2},k}\|^2+(1+a)(1+\frac{1}{b})\frac{\beta^2}{K}\sum_{k' \in [K]}\mathbb{E}\|\e_{t-\frac{1}{2},k'}\|^2 +(1+\frac{1}{a})G^2. 
    \end{align*}  
  Summing up the above equation from $k=0$ to $K-1$, we can get
    \begin{align*}
      \frac{1}{K}\sum_{k \in [K]}\mathbb{E}\|\e_{t+\frac{1}{2},k}\|^2 & \leq  [(1+a)(1+b)(1-\frac{\beta}{K})^2(1-\delta)+(1+a)(1+\frac{1}{b})\beta^2]\frac{1}{K}\sum_{k \in [K]}\mathbb{E}\|\e_{t-\frac{1}{2},k}\|^2\\
      &~~~~~ +(1+\frac{1}{a})G^2. 
      \end{align*}  
    Let $a = \frac{\delta}{4}$, $b = \frac{\delta}{2}$, we get
    \begin{align*}
      \frac{1}{K}\sum_{k \in [K]}\mathbb{E}\|\e_{t+\frac{1}{2},k}\|^2  \leq  [(1-\frac{\delta}{4})(1-\frac{\beta}{K})^2+(1+\frac{\delta}{4})(1+\frac{2}{\delta})\beta^2]\frac{1}{K}\sum_{k \in [K]}\mathbb{E}\|\e_{t-\frac{1}{2},k}\|^2 +(1+\frac{4}{\delta})G^2. 
    \end{align*}   
    Let $\beta \leq \frac{\delta}{4\sqrt{2+\delta}}$, then we have
    \begin{align*}
      (1-\frac{\delta}{4})(1-\frac{\beta}{K})^2+(1+\frac{\delta}{4})(1+\frac{2}{\delta})\beta^2 \leq (1-\frac{\delta}{4})+2(1+\frac{2}{\delta})\beta^2 \leq 1- \frac{\delta}{8} < 1.
      \end{align*} 
    Then we get 
    \begin{align*}
      \frac{1}{K}\sum_{k \in [K]}\mathbb{E}\|\e_{t+1,k}\|^2 &= \frac{1}{K}\sum_{k \in [K]}\mathbb{E}\|\e_{t+\frac{1}{2},k}-\mathcal{C}(\e_{t+\frac{1}{2},k})\|^2 \leq  (1-\delta)\frac{1}{K}\sum_{k \in [K]}\mathbb{E}\|\e_{t+\frac{1}{2},k}\|^2 \\
    &\leq \frac{(1-\delta)(1+\frac{4}{\delta})G^2}{1-[(1-\frac{\delta}{4})(1-\frac{\beta}{K})^2+(1+\frac{\delta}{4})(1+\frac{2}{\delta})\beta^2]}.
    \end{align*}    

\subsection{Proof of Lemma~\ref{lemma:w to barz}}
  \begin{align*}
    \bar{\z}_{t+1} &= \bar{\w}_{t+1}+\frac{\beta}{1-\beta}(\bar{\w}_{t+1} - \bar{\w}_t) - \frac{1}{1-\beta}[(1-\lambda) \eta \bar{\e}_{t+1}+ \beta\lambda \eta \bar{\e}_t] \\
               &= \frac{1}{1-\beta}\bar{\w}_{t+1} -\frac{\beta}{1-\beta}\bar{\w}_t - \frac{1}{1-\beta}[(1-\lambda) \eta \bar{\e}_{t+1}+ \beta\lambda \eta \bar{\e}_t] \\
               &= \frac{1}{1-\beta} (\bar{\w}_t-\beta\bar{\w}_{t-1})- \frac{1}{1-\beta}[(1-\lambda)\eta \bar{\e}_t+\beta\lambda \eta \bar{\e}_{t-1}]-\frac{\eta}{1-\beta}\frac{1}{K}\sum_{k \in [K]}\nabla f(\w_{t,k};\IM_{t,k})\\
               &= \bar{\z}_t-\frac{\eta}{1-\beta}\frac{1}{K}\sum_{k \in [K]}\nabla f(\w_{t,k};\IM_{t,k}).
              \end{align*}

\subsection{Proof of Lemma~\ref{lemma:bounded bargap}}
    \begin{align*}
    &\mathbb{E}\|\bar{\w}_{t+1}-\bar{\w}_{t}\|^2 \\
    &= \mathbb{E}\|\beta (\bar{\w}_{t}-\bar{\w}_{t-1})+ \eta (1-\lambda)(\bar{\e}_{t+1}-\bar{\e}_{t}) +\beta\lambda \eta (\bar{\e}_{t}-\bar{\e}_{t-1})-\frac{\eta}{K}\sum_{k \in [K]}\nabla f(\w_{t,k};\IM_{t,k})\|^2\\
    &\leq  (1+a)\beta^2 \mathbb{E}\|\bar{\w}_{t}-\bar{\w}_{t-1}\|^2\\
    &~~~~+(1+\frac{1}{a})\eta^2\mathbb{E}\|(1-\lambda)(\bar{\e}_{t+1}-\bar{\e}_{t})+\beta\lambda (\bar{\e}_{t}-\bar{\e}_{t-1})-\frac{1}{K}\sum_{k \in [K]}\nabla f(\w_{t,k};\IM_{t,k})\|^2\\
    &\leq  (1+a)\beta^2 \mathbb{E}\|\bar{\w}_{t}-\bar{\w}_{t-1}\|^2+5(1+\frac{1}{a})\eta^2[2(1-\lambda)^2 E^2+2\beta^2\lambda^2 E^2+G^2].
  \end{align*}
  If $\beta \in (0,1)$, let $a=\frac{1}{\beta}-1$, then we have
  \begin{align*}
    \mathbb{E}\|\bar{\w}_{t}-\bar{\w}_{t-1}\|^2 &\leq  \beta \mathbb{E}\|\bar{\w}_{t-1}-\bar{\w}_{t-2}\|^2+\frac{10((1-\lambda)^2+\beta^2\lambda^2)E^2+5G^2}{1-\beta}\eta^2  \\
    &\leq  \frac{10((1-\lambda)^2+\beta^2\lambda^2)E^2+5G^2}{(1-\beta)^2}\eta^2. 
  \end{align*}    
  If $\beta = 0$, it's obvious that $\mathbb{E}\|\w_{t}-\w_{t-1}\|^2$ can be bounded by $\frac{10((1-\lambda)^2+\beta^2\lambda^2)E^2+5G^2}{(1-\beta)^2}\eta^2$.

  \begin{align*}
    \mathbb{E}\|\bar{\z}_t - \bar{\w}_t\|^2 &= \mathbb{E}\|\frac{\beta}{1-\beta}(\bar{\w}_{t} - \bar{\w}_{t-1}) - \frac{1}{1-\beta}[(1-\lambda) \eta \bar{\e}_{t}+ \beta\lambda \eta \bar{\e}_{t-1}]\|^2 \\
        &\leq  2\frac{\beta^2}{(1-\beta)^2}\mathbb{E}\|\bar{\w}_{t} - \bar{\w}_{t-1}\|^2 +\frac{2\eta^2}{(1-\beta)^2}\mathbb{E}\|(1-\lambda) \bar{\e}_{t}+ \beta\lambda \bar{\e}_{t-1}\|^2 \\
        &\leq 2\frac{\beta^2}{(1-\beta)^2}\mathbb{E}\|\bar{\w}_{t} - \bar{\w}_{t-1}\|^2 +\frac{4\eta^2}{(1-\beta)^2}[(1-\lambda)^2\mathbb{E}\|\bar{\e}_{t}\|^2+\beta^2\lambda^2\mathbb{E}\|\bar{\e}_{t-1}\|^2] \\
        &\leq 2\frac{\beta^2}{(1-\beta)^2}\mathbb{E}\|\bar{\w}_{t} - \bar{\w}_{t-1}\|^2 +\frac{4\eta^2E^2((1-\lambda)^2+\beta^2\lambda^2)}{(1-\beta)^2}\\
        &\leq 2\frac{\beta^2}{(1-\beta)^2}\frac{10((1-\lambda)^2+\beta^2\lambda^2)E^2+5G^2}{(1-\beta)^2}\eta^2 +\frac{4\eta^2E^2((1-\lambda)^2+\beta^2\lambda^2)}{(1-\beta)^2}\\
        &\leq [\frac{20\beta^2((1-\lambda)^2+\beta^2\lambda^2)E^2+10\beta^2G^2}{(1-\beta)^4} +\frac{4E^2((1-\lambda)^2+\beta^2\lambda^2)}{(1-\beta)^2}]\eta^2.
      \end{align*}

\subsection{Proof of Theorem~\ref{theorem:defgmc}}

 \begin{align*}
  \mathbb{E} F(\bar{\z}_{t+1}) &\leq  F(\bar{\z}_{t})-\frac{\eta}{1-\beta}\nabla F(\bar{\z}_{t})^{T}(\frac{1}{K}\sum_{k \in [K]}\nabla F_k(\w_{t,k}))+ \frac{L\eta^2}{2(1-\beta)^2} \mathbb{E}\|\frac{1}{K}\sum_{k \in [K]}\nabla f(\w_{t,k};\IM_{t,k})\|^2. 
\end{align*}
\begin{align*}
  &-\frac{\eta}{1-\beta}\nabla F(\bar{\z}_{t})^{T}(\frac{1}{K}\sum_{k \in [K]}\nabla F_k(\w_{t,k}))  \\
  &=-\frac{\eta}{1-\beta}(\nabla F(\bar{\z}_{t})-\nabla F(\w_t))^{T}(\frac{1}{K}\sum_{k \in [K]}\nabla F_k(\w_{t,k}))-\frac{\eta}{1-\beta}(\nabla F(\w_t))^{T}(\frac{1}{K}\sum_{k \in [K]}\nabla F_k(\w_{t,k})) \\
  &\leq -\frac{\eta}{1-\beta}[2(\nabla F(\bar{\z}_t)-\nabla F(\w_t))]^{T}[\frac{1}{2K}\sum_{k \in [K]}\nabla F_k(\w_{t,k})]-\frac{\eta}{1-\beta}(\nabla F(\w_t))^{T}(\frac{1}{K}\sum_{k \in [K]}\nabla F_k(\w_{t,k})) \\
  &\leq \frac{2\eta}{1-\beta}\|\nabla F(\bar{\z}_t)-\nabla F(\w_t)\|^2+\frac{\eta}{8(1-\beta)}\|\frac{1}{K}\sum_{k \in [K]}\nabla F_k(\w_{t,k})\|^2 \\
  &~~~~- \frac{\eta}{2(1-\beta)} [\|\nabla F(\w_t)\|^2 + \|\frac{1}{K}\sum_{k \in [K]}\nabla F_k(\w_{t,k})\|^2-\|\nabla F(\w_t)-\frac{1}{K}\sum_{k \in [K]}\nabla F_k(\w_{t,k})\|^2]   \\
   &\leq \frac{2\eta L^2}{1-\beta}\|\bar{\z}_t-\w_t\|^2 - \frac{\eta}{2(1-\beta)} \|\nabla F(\w_t)\|^2+\frac{\eta}{2(1-\beta)}\|\nabla F(\w_t)-\frac{1}{K}\sum_{k \in [K]}\nabla F_k(\w_{t,k})\|^2\\
   &~~~~-\frac{3\eta}{8(1-\beta)} \|\frac{1}{K}\sum_{k \in [K]}\nabla F_k(\w_{t,k})\|^2.  
  \end{align*}

\begin{align*}
\mathbb{E}\|\frac{1}{K}\sum_{k \in [K]}\nabla f(\w_{t,k};\IM_{t,k})\|^2 &= \mathbb{E}\|\frac{1}{K}\sum_{k \in [K]}(\nabla f(\w_{t,k};\IM_{t,k})-\nabla F_k(\w_{t,k})+\nabla F_k(\w_{t,k}))\|^2 \\
  &= \mathbb{E}\|\frac{1}{K}\sum_{k \in [K]}(\nabla f(\w_{t,k};\IM_{t,k})-\nabla F_k(\w_{t,k}))\|^2+\|\frac{1}{K}\sum_{k \in [K]}\nabla F_k(\w_{t,k})\|^2 \\
  &= \frac{1}{K^2}\sum_{k \in [K]}\mathbb{E}\|\nabla f(\w_{t,k};\IM_{t,k})-\nabla F_k(\w_{t,k})\|^2+\|\frac{1}{K}\sum_{k \in [K]}\nabla F_k(\w_{t,k})\|^2 \\
  & \leq \frac{\sigma^2}{Kb}+\|\frac{1}{K}\sum_{k \in [K]}\nabla F_k(\w_{t,k})\|^2. 
\end{align*}

\begin{align*}
  \mathbb{E} F(\bar{\z}_{t+1}) &\leq  F(\bar{\z}_{t})-\frac{\eta}{1-\beta}\nabla F(\bar{\z}_{t})^{T}(\frac{1}{K}\sum_{k \in [K]}\nabla F_k(\w_{t,k}))+ \frac{L\eta^2}{2(1-\beta)^2} \mathbb{E}\|\frac{1}{K}\sum_{k \in [K]}\nabla f(\w_{t,k};\IM_{t,k})\|^2 \\      
  \leq &  F(\bar{\z}_{t})+\frac{2\eta L^2}{1-\beta}\|\bar{\z}_t-\w_t\|^2+[-\frac{3\eta}{8(1-\beta)}+ \frac{L\eta^2}{2(1-\beta)^2}]\|\frac{1}{K}\sum_{k \in [K]}\nabla F_k(\w_{t,k})\|^2 \\
  &~~~~- \frac{\eta}{2(1-\beta)} \|\nabla F(\w_t)\|^2+\frac{\eta}{2(1-\beta)}\|\nabla F(\w_t)-\frac{1}{K}\sum_{k \in [K]}\nabla F_k(\w_{t,k})\|^2+\frac{L\sigma^2}{2(1-\beta)^2b}\frac{\eta^2}{K}.  
\end{align*}
Since we assume $\eta \leq \frac{3(1-\beta)}{4L}$, we can obtain that $-\frac{3\eta}{8(1-\beta)}+ \frac{L\eta^2}{2(1-\beta)^2} \leq 0$.
\begin{align*}
  \mathbb{E} F(\bar{\z}_{t+1}) &\leq   F(\bar{\z}_{t})+\frac{2\eta L^2}{1-\beta}\|\bar{\z}_t-\w_t\|^2+\frac{\eta}{2(1-\beta)}\|\nabla F(\w_t)-\frac{1}{K}\sum_{k \in [K]}\nabla F_k(\w_{t,k})\|^2\\
  &~~~~+\frac{L\sigma^2}{2(1-\beta)^2b}\frac{\eta^2}{K} - \frac{\eta}{2(1-\beta)} \|\nabla F(\w_t)\|^2.  
\end{align*}

\begin{align*}
  \mathbb{E}\|\nabla F(\w_t)-\frac{1}{K}\sum_{k \in [K]}\nabla F_k(\w_{t,k})\|^2  &= \mathbb{E}\|\frac{1}{K}\sum_{k \in [K]}[\nabla F_k(\w_t)-\nabla F_k(\w_{t,k})]\|^2\\
  &\leq  \frac{1}{K}\sum_{k \in [K]}\mathbb{E}\|\nabla F_k(\w_t)-\nabla F_k(\w_{t,k})\|^2 \\
  &\leq  \frac{L^2}{K}\sum_{k \in [K]}\mathbb{E}\|\w_t-\w_{t,k}\|^2  \leq  \frac{L^2 }{K}\sum_{k \in [K]}\mathbb{E}\|\lambda\eta\e_{t,k}\|^2 \\
  &\leq  \frac{L^2 \lambda^2 \eta^2}{K}\sum_{k \in [K]}\mathbb{E}\|\e_{t,k}\|^2 \leq L^2 \lambda^2E^2\eta^2.
\end{align*}
\begin{align*}
  \|\bar{\z}_t-\w_t\|^2 & \leq 2\|\bar{\z}_t-\bar{\w}_t\|^2 + 2\|\bar{\w}_t-\w_t\|^2 \leq 2(C_2 +\lambda^2E^2)\eta^2.
\end{align*}
\begin{align*}
  \mathbb{E} F(\bar{\z}_{t+1}) &\leq \mathbb{E}F(\bar{\z}_{t})+\frac{2\eta L^2}{1-\beta}\|\bar{\z}_t-\w_t\|^2+\frac{\eta}{2(1-\beta)}\mathbb{E}\|\nabla F(\w_t)-\frac{1}{K}\sum_{k \in [K]}\nabla F_k(\w_{t,k})\|^2\\
  &~~~~+\frac{L\sigma^2}{2b(1-\beta)^2}\frac{\eta^2}{K} - \frac{\eta}{2(1-\beta)} \mathbb{E}\|\nabla F(\w_t)\|^2  \\
  &\leq  \mathbb{E} F(\bar{\z}_{t})+\frac{(4C_2 +5\lambda^2E^2)L^2\eta^3}{1-\beta}+\frac{L\sigma^2}{2(1-\beta)^2b}\frac{\eta^2}{K} - \frac{\eta}{2(1-\beta)} \mathbb{E}\|\nabla F(\w_t)\|^2.  
\end{align*}
\begin{align*}
 \mathbb{E}\|\nabla F(\w_t)\|^2 \leq  \frac{2(1-\beta)[\mathbb{E} F(\bar{\z}_{t})-\mathbb{E} F(\bar{\z}_{t+1})]}{\eta}+\frac{L\sigma^2}{(1-\beta)b}\frac{\eta}{K} +(8C_2 +10\lambda^2E^2)L^2\eta^2.  
\end{align*}
Summing up the above equation from $t=0$ to $T-1$, we have
\begin{align*}
  \frac{1}{T}\sum_{t \in [T]}\mathbb{E}\|\nabla F(\w_t)\|^2 \leq  \frac{2(1-\beta) (F(\bar{\z}_{0})-F^{*})}{T\eta}+\frac{L\sigma^2}{(1-\beta)b}\frac{\eta}{K} +(8C_2 +10\lambda^2E^2)L^2\eta^2.  
 \end{align*} 

% \vskip 0.2in
\bibliography{sample}

\begin{thebibliography}{37}
\providecommand{\natexlab}[1]{#1}
\providecommand{\url}[1]{\texttt{#1}}
\expandafter\ifx\csname urlstyle\endcsname\relax
  \providecommand{\doi}[1]{doi: #1}\else
  \providecommand{\doi}{doi: \begingroup \urlstyle{rm}\Url}\fi

\bibitem[Aji and Heafield(2017)]{DBLP:conf/emnlp/AjiH17}
Alham~Fikri Aji and Kenneth Heafield.
\newblock Sparse communication for distributed gradient descent.
\newblock In \emph{Proceedings of the Conference on Empirical Methods in
  Natural Language Processing}, pages 440--445, 2017.

\bibitem[Alistarh et~al.(2017)Alistarh, Grubic, Li, Tomioka, and
  Vojnovic]{DBLP:conf/nips/AlistarhG0TV17}
Dan Alistarh, Demjan Grubic, Jerry Li, Ryota Tomioka, and Milan Vojnovic.
\newblock {QSGD}: Communication-efficient {SGD} via gradient quantization and
  encoding.
\newblock In \emph{Advances in Neural Information Processing Systems}, pages
  1707--1718, 2017.

\bibitem[Alistarh et~al.(2018)Alistarh, Hoefler, Johansson, Konstantinov,
  Khirirat, and Renggli]{DBLP:conf/nips/AlistarhH0KKR18}
Dan Alistarh, Torsten Hoefler, Mikael Johansson, Nikola Konstantinov, Sarit
  Khirirat, and C{\'{e}}dric Renggli.
\newblock The convergence of sparsified gradient methods.
\newblock In \emph{Advances in Neural Information Processing Systems}, pages
  5977--5987, 2018.

\bibitem[Basu et~al.(2019)Basu, Data, Karakus, and
  Diggavi]{DBLP:conf/nips/0001DKD19}
Debraj Basu, Deepesh Data, Can Karakus, and Suhas~N. Diggavi.
\newblock Qsparse-local-sgd: Distributed {SGD} with quantization,
  sparsification and local computations.
\newblock In \emph{Advances in Neural Information Processing Systems}, pages
  14668--14679, 2019.

\bibitem[Bottou(2010)]{DBLP:conf/compstat/Bottou10}
L{\'{e}}on Bottou.
\newblock Large-scale machine learning with stochastic gradient descent.
\newblock In \emph{Proceedings of the International Conference on Computational
  Statistics}, pages 177--186, 2010.

\bibitem[Brown et~al.(2020)Brown, Mann, Ryder, Subbiah, Kaplan, Dhariwal,
  Neelakantan, Shyam, Sastry, Askell, et~al.]{brown2020language}
Tom Brown, Benjamin Mann, Nick Ryder, Melanie Subbiah, Jared~D Kaplan, Prafulla
  Dhariwal, Arvind Neelakantan, Pranav Shyam, Girish Sastry, Amanda Askell,
  et~al.
\newblock Language models are few-shot learners.
\newblock In \emph{Advances in Neural Information Processing Systems}, pages
  1877--1901, 2020.

\bibitem[Devlin et~al.(2019)Devlin, Chang, Lee, and
  Toutanova]{Devlin2019BERTPO}
Jacob Devlin, Ming-Wei Chang, Kenton Lee, and Kristina Toutanova.
\newblock Bert: Pre-training of deep bidirectional transformers for language
  understanding.
\newblock In \emph{Proceedings of the Conference of the North American Chapter
  of the Association for Computational Linguistics: Human Language
  Technologies}, pages 4171--4186, 2019.

\bibitem[He et~al.(2016)He, Zhang, Ren, and Sun]{DBLP:conf/cvpr/HeZRS16}
Kaiming He, Xiangyu Zhang, Shaoqing Ren, and Jian Sun.
\newblock Deep residual learning for image recognition.
\newblock In \emph{Proceedings of the IEEE/CVF Conference on Computer Vision
  and Pattern Recognition}, pages 770--778, 2016.

\bibitem[Hsieh et~al.(2020)Hsieh, Phanishayee, Mutlu, and
  Gibbons]{DBLP:conf/icml/HsiehPMG20}
Kevin Hsieh, Amar Phanishayee, Onur Mutlu, and Phillip~B. Gibbons.
\newblock The non-iid data quagmire of decentralized machine learning.
\newblock In \emph{Proceedings of the International Conference on Machine
  Learning}, pages 4387--4398, 2020.

\bibitem[Hsu et~al.(2019)Hsu, Qi, and Brown]{DBLP:journals/corr/abs-1909-06335}
Tzu{-}Ming~Harry Hsu, Hang Qi, and Matthew Brown.
\newblock Measuring the effects of non-identical data distribution for
  federated visual classification.
\newblock \emph{CoRR}, abs/1909.06335, 2019.

\bibitem[Jiang and Agrawal(2018)]{DBLP:conf/nips/JiangA18}
Peng Jiang and Gagan Agrawal.
\newblock A linear speedup analysis of distributed deep learning with sparse
  and quantized communication.
\newblock In \emph{Advances in Neural Information Processing Systems}, pages
  2530--2541, 2018.

\bibitem[Johnson and Zhang(2013)]{DBLP:conf/nips/Johnson013}
Rie Johnson and Tong Zhang.
\newblock Accelerating stochastic gradient descent using predictive variance
  reduction.
\newblock In \emph{Advances in Neural Information Processing Systems}, pages
  315--323, 2013.

\bibitem[Karimireddy et~al.(2019)Karimireddy, Rebjock, Stich, and
  Jaggi]{DBLP:conf/icml/KarimireddyRSJ19}
Sai~Praneeth Karimireddy, Quentin Rebjock, Sebastian~U. Stich, and Martin
  Jaggi.
\newblock Error feedback fixes signsgd and other gradient compression schemes.
\newblock In \emph{Proceedings of International Conference on Machine
  Learning}, pages 3252--3261, 2019.

\bibitem[Kingma and Ba(2015)]{DBLP:journals/corr/KingmaB14}
Diederik~P. Kingma and Jimmy Ba.
\newblock Adam: {A} method for stochastic optimization.
\newblock In \emph{Proceedings of International Conference on Learning
  Representations}, 2015.

\bibitem[Koloskova et~al.(2020)Koloskova, Lin, Stich, and
  Jaggi]{koloskova2019decentralized}
Anastasia Koloskova, Tao Lin, Sebastian~U Stich, and Martin Jaggi.
\newblock Decentralized deep learning with arbitrary communication compression.
\newblock In \emph{Proceedings of International Conference on Learning
  Representations}, 2020.

\bibitem[Krizhevsky et~al.(2012)Krizhevsky, Sutskever, and
  Hinton]{DBLP:conf/nips/KrizhevskySH12}
Alex Krizhevsky, Ilya Sutskever, and Geoffrey~E. Hinton.
\newblock Imagenet classification with deep convolutional neural networks.
\newblock In \emph{Advances in Neural Information Processing Systems}, pages
  1106--1114, 2012.

\bibitem[Lan(2012)]{DBLP:journals/mp/Lan12}
Guanghui Lan.
\newblock An optimal method for stochastic composite optimization.
\newblock \emph{Mathematical Programming}, 133\penalty0 (1-2):\penalty0
  365--397, 2012.

\bibitem[Lee et~al.(2021)Lee, Lee, and Song]{DBLP:journals/corr/abs-2112-13492}
Seung~Hoon Lee, Seunghyun Lee, and Byung~Cheol Song.
\newblock Vision transformer for small-size datasets.
\newblock \emph{CoRR}, abs/2112.13492, 2021.

\bibitem[Li et~al.(2014)Li, Andersen, Park, Smola, Ahmed, Josifovski, Long,
  Shekita, and Su]{DBLP:conf/osdi/LiAPSAJLSS14}
Mu~Li, David~G. Andersen, Jun~Woo Park, Alexander~J. Smola, Amr Ahmed, Vanja
  Josifovski, James Long, Eugene~J. Shekita, and Bor{-}Yiing Su.
\newblock Scaling distributed machine learning with the parameter server.
\newblock In \emph{Proceedings of the 11th Symposium on Operating Systems
  Design and Implementation}, pages 583--598, 2014.

\bibitem[Lin et~al.(2021)Lin, Karimireddy, Stich, and
  Jaggi]{DBLP:conf/icml/0004KSJ21}
Tao Lin, Sai~Praneeth Karimireddy, Sebastian~U. Stich, and Martin Jaggi.
\newblock Quasi-global momentum: Accelerating decentralized deep learning on
  heterogeneous data.
\newblock In \emph{Proceedings of International Conference on Machine
  Learning}, pages 6654--6665, 2021.

\bibitem[Lin et~al.(2018)Lin, Han, Mao, Wang, and
  Dally]{DBLP:conf/iclr/LinHM0D18}
Yujun Lin, Song Han, Huizi Mao, Yu~Wang, and Bill Dally.
\newblock Deep gradient compression: Reducing the communication bandwidth for
  distributed training.
\newblock In \emph{Proceedings of International Conference on Learning
  Representations}, 2018.

\bibitem[Loshchilov and Hutter(2017)]{DBLP:conf/iclr/LoshchilovH17}
Ilya Loshchilov and Frank Hutter.
\newblock {SGDR}: Stochastic gradient descent with warm restarts.
\newblock In \emph{Proceedings of International Conference on Learning
  Representations}, 2017.

\bibitem[Polyak(1964)]{article}
Boris Polyak.
\newblock Some methods of speeding up the convergence of iteration methods.
\newblock \emph{Ussr Computational Mathematics and Mathematical Physics},
  4:\penalty0 1--17, 12 1964.

\bibitem[Robbins and Monro(1951)]{Robbins&Monro:1951}
Herbert Robbins and Sutton Monro.
\newblock A stochastic approximation method.
\newblock \emph{Annals of Mathematical Statistics}, 22:\penalty0 400--407,
  1951.

\bibitem[Stich et~al.(2018)Stich, Cordonnier, and
  Jaggi]{DBLP:conf/nips/StichCJ18}
Sebastian~U. Stich, Jean{-}Baptiste Cordonnier, and Martin Jaggi.
\newblock Sparsified {SGD} with memory.
\newblock In \emph{Advances in Neural Information Processing Systems}, pages
  4452--4463, 2018.

\bibitem[Sutskever et~al.(2013)Sutskever, Martens, Dahl, and
  Hinton]{DBLP:conf/icml/SutskeverMDH13}
Ilya Sutskever, James Martens, George~E. Dahl, and Geoffrey~E. Hinton.
\newblock On the importance of initialization and momentum in deep learning.
\newblock In \emph{Proceedings of International Conference on Machine
  Learning}, pages 1139--1147, 2013.

\bibitem[Tang et~al.(2019)Tang, Yu, Lian, Zhang, and
  Liu]{DBLP:conf/icml/TangYLZL19}
Hanlin Tang, Chen Yu, Xiangru Lian, Tong Zhang, and Ji~Liu.
\newblock Doublesqueeze: Parallel stochastic gradient descent with double-pass
  error-compensated compression.
\newblock In \emph{Proceedings of International Conference on Machine
  Learning}, pages 6155--6165, 2019.

\bibitem[Touvron et~al.(2023)Touvron, Lavril, Izacard, Martinet, Lachaux,
  Lacroix, Rozi{\`{e}}re, Goyal, Hambro, Azhar, Rodriguez, Joulin, Grave, and
  Lample]{DBLP:journals/corr/abs-2302-13971}
Hugo Touvron, Thibaut Lavril, Gautier Izacard, Xavier Martinet, Marie{-}Anne
  Lachaux, Timoth{\'{e}}e Lacroix, Baptiste Rozi{\`{e}}re, Naman Goyal, Eric
  Hambro, Faisal Azhar, Aur{\'{e}}lien Rodriguez, Armand Joulin, Edouard Grave,
  and Guillaume Lample.
\newblock Llama: Open and efficient foundation language models.
\newblock \emph{CoRR}, abs/2302.13971, 2023.

\bibitem[Tseng(1998)]{DBLP:journals/siamjo/Tseng98}
Paul Tseng.
\newblock An incremental gradient(-projection) method with momentum term and
  adaptive stepsize rule.
\newblock \emph{{SIAM} Journal on Optimization}, 8\penalty0 (2):\penalty0
  506--531, 1998.

\bibitem[Vogels et~al.(2019)Vogels, Karimireddy, and
  Jaggi]{DBLP:conf/nips/VogelsKJ19}
Thijs Vogels, Sai~Praneeth Karimireddy, and Martin Jaggi.
\newblock Powersgd: Practical low-rank gradient compression for distributed
  optimization.
\newblock In \emph{Advances in Neural Information Processing Systems}, 2019.

\bibitem[Wen et~al.(2017)Wen, Xu, Yan, Wu, Wang, Chen, and
  Li]{DBLP:conf/nips/WenXYWWCL17}
Wei Wen, Cong Xu, Feng Yan, Chunpeng Wu, Yandan Wang, Yiran Chen, and Hai Li.
\newblock Terngrad: Ternary gradients to reduce communication in distributed
  deep learning.
\newblock In \emph{Advances in Neural Information Processing Systems}, pages
  1508--1518, 2017.

\bibitem[Wu and He(2018)]{wu2018group}
Yuxin Wu and Kaiming He.
\newblock Group normalization.
\newblock In \emph{Proceedings of the European Conference on Computer Vision},
  pages 3--19, 2018.

\bibitem[Xie et~al.(2020)Xie, Zheng, Koyejo, Gupta, Li, and
  Lin]{DBLP:conf/nips/XieZKGLL20}
Cong Xie, Shuai Zheng, Oluwasanmi Koyejo, Indranil Gupta, Mu~Li, and Haibin
  Lin.
\newblock {CSER}: Communication-efficient {SGD} with error reset.
\newblock In \emph{Advances in Neural Information Processing Systems}, pages
  12593--12603, 2020.

\bibitem[Xu and Huang(2022)]{DBLP:conf/icml/XuH22}
An~Xu and Heng Huang.
\newblock Detached error feedback for distributed {SGD} with random
  sparsification.
\newblock In \emph{Proceedings of International Conference on Machine
  Learning}, pages 24550--24575, 2022.

\bibitem[Zhao et~al.(2018)Zhao, Zhang, Li, and Li]{NEURIPS2018_f4334c13}
Shen-Yi Zhao, Gong-Duo Zhang, Ming-Wei Li, and Wu-Jun Li.
\newblock Proximal scope for distributed sparse learning.
\newblock In \emph{Advances in Neural Information Processing Systems}, 2018.

\bibitem[Zhao et~al.(2020)Zhao, Xie, and Li]{DBLP:journals/corr/abs-2007-13985}
Shen{-}Yi Zhao, Yin{-}Peng Xie, and Wu{-}Jun Li.
\newblock Stochastic normalized gradient descent with momentum for large batch
  training.
\newblock \emph{CoRR}, abs/2007.13985, 2020.

\bibitem[Zhao et~al.(2021)Zhao, Xie, and Li]{DBLP:journals/chinaf/ZhaoXL21}
Shen{-}Yi Zhao, Yin{-}Peng Xie, and Wu{-}Jun Li.
\newblock On the convergence and improvement of stochastic normalized gradient
  descent.
\newblock \emph{Sci. China Inf. Sci.}, 64\penalty0 (3), 2021.

\end{thebibliography}

\end{document}